\pdfoutput=1
\documentclass[11pt]{article}

\usepackage[final]{acl}
\usepackage{times}
\usepackage{latexsym}
\usepackage{hyperref}
\usepackage[T1]{fontenc}
\usepackage[utf8]{inputenc}
\usepackage{microtype}
\usepackage{inconsolata}
\usepackage{booktabs}
\usepackage{multicol}
\usepackage{multirow}
\usepackage{graphicx}
\usepackage{capt-of}
\usepackage{scalerel}
\usepackage{tablefootnote}
\usepackage{caption}
\usepackage{subcaption}
\usepackage{float}
\usepackage{titlesec}
\usepackage{enumitem}
\usepackage{fancyvrb}
\usepackage{adjustbox}
\usepackage{fancyvrb}
\usepackage{varwidth}
\usepackage{fontawesome5}
\usepackage{placeins}
\usepackage[noabbrev,capitalize,nameinlink]{cleveref}


\definecolor{CommentGrey}{HTML}{949494}
\definecolor{DeepForest}{HTML}{227805}
\definecolor{DeepBlood}{HTML}{780505}
\definecolor{Twitch}{HTML}{8c34eb}
\definecolor{Saffron}{HTML}{eb5c34}
\definecolor{Navy}{HTML}{2c59a5}
\newcommand{\diffup}[1]{\small{\textbf{\color{DeepForest}\texttt{+#1}}}}
\newcommand{\diffdo}[1]{\small{\textbf{\color{DeepBlood}\texttt{-#1}}}}

\titlespacing*{\section}
{0pt}{1.2ex plus .2ex minus .2ex}{1.2ex plus .2ex}
\titlespacing*{\subsection}
{0pt}{1.2ex plus .2ex minus .2ex}{1.2ex plus .2ex}

\title{\scalerel*{\includegraphics{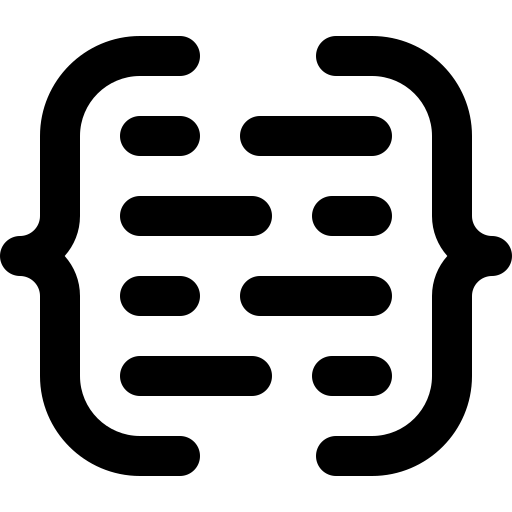}}{\texttt{I}} \texttt{IRCoder}: Intermediate Representations Make
\\Language Models Robust Multilingual Code Generators}

\author{
Indraneil Paul\textsuperscript{1},
Goran Glavas\textsuperscript{2},
\and
Iryna Gurevych\textsuperscript{1}
\\
\textsuperscript{1} Ubiquitous Knowledge Processing Lab (UKP Lab)\\ 
Department of Computer Science and Hessian Center for AI (hessian.AI)\\
Technische Universität Darmstadt\\
\textsuperscript{2} CAIDAS, University of Würzburg \\
\vspace{-1.3em} \\
\texttt{\small \href{www.ukp.tu-darmstadt.de}{www.ukp.tu-darmstadt.de}} \\
\vspace{-0.8em} \\
\scalerel*{\includegraphics{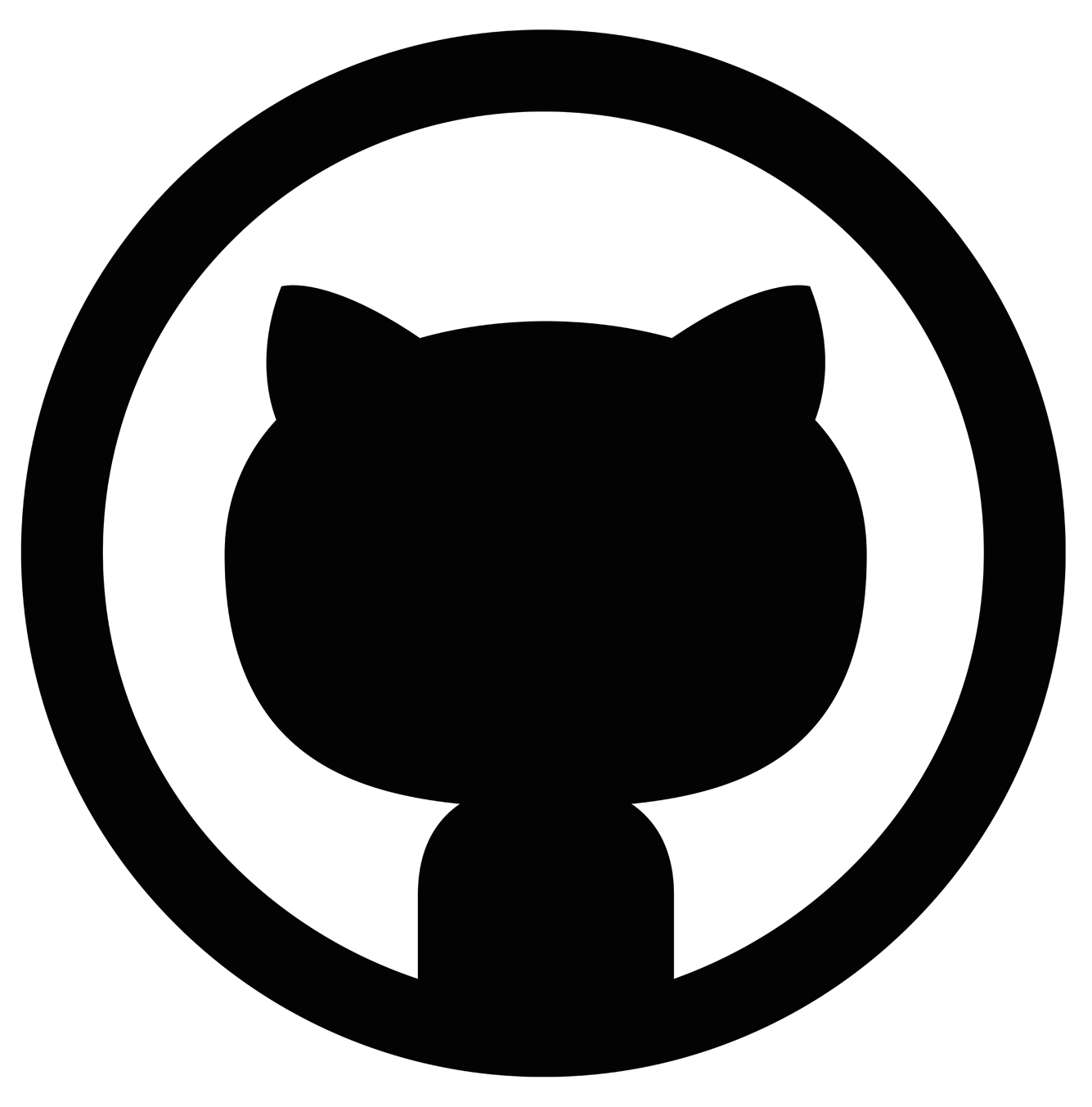}}{\texttt{C}} \texttt{\href{https://github.com/UKPLab/arxiv2024-IRCoder}{Code}} \hspace{0.5em} \scalerel*{\includegraphics{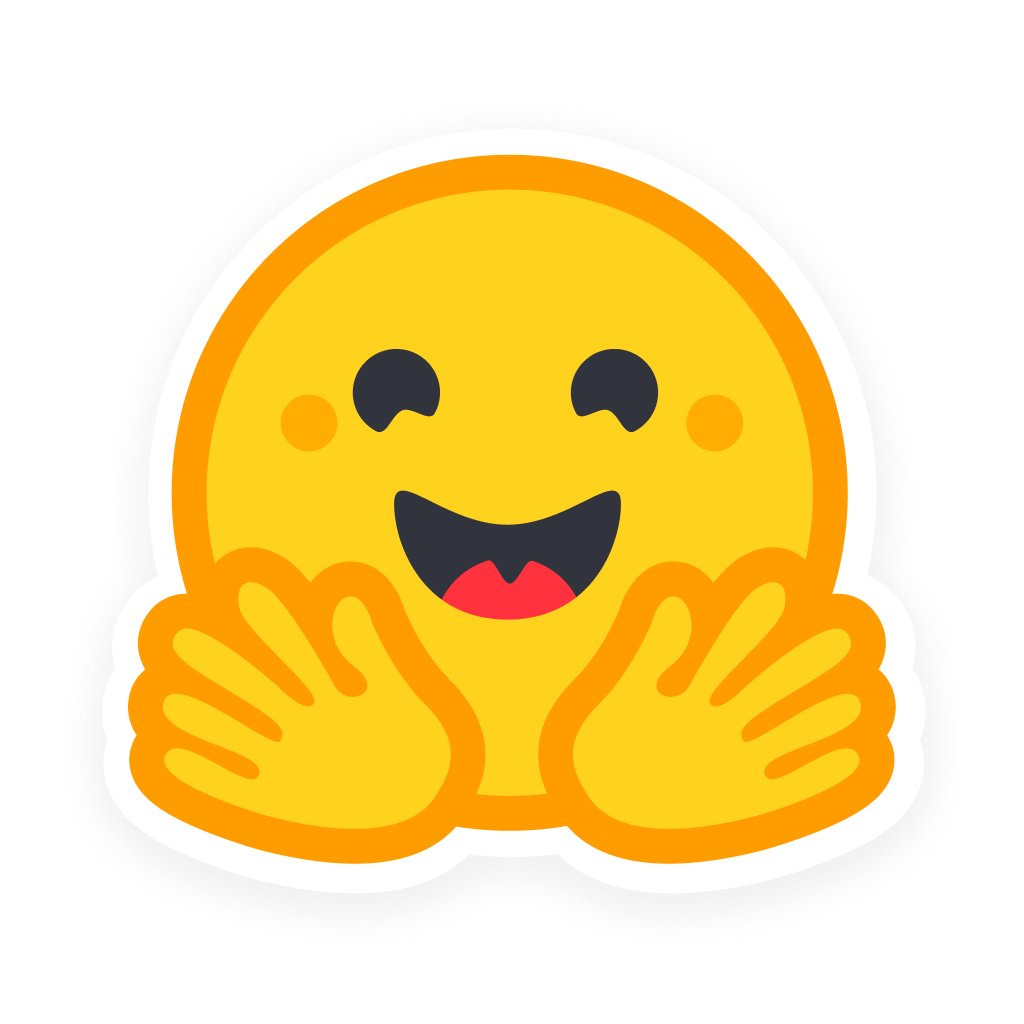}}{\texttt{D}} \texttt{\href{https://huggingface.co/datasets/UKPLab/SLTrans}{Dataset}} \\
}

\begin{document}

\maketitle
\begin{abstract}

Code understanding and generation have fast become some of the most popular applications of language models (LMs). Nonetheless, research on multilingual aspects of Code-LMs (i.e., LMs for code generation) such as cross-lingual transfer between different programming languages, language-specific data augmentation, and post-hoc LM adaptation, alongside exploitation of data sources other than the original textual content, has been much sparser than for their natural language counterparts. In particular, most mainstream Code-LMs have been pre-trained on source code files alone. 
In this work, we investigate the prospect of leveraging readily available compiler \textit{intermediate representations} (IR)---shared across programming languages---to improve the multilingual capabilities of Code-LMs and facilitate cross-lingual transfer. 

To this end, we first compile \textbf{\texttt{SLTrans}}, a parallel dataset consisting of nearly 4M self-contained source code files coupled with their respective intermediate representations. 
Next, starting from various base Code-LMs (ranging in size from 1.1B to 7.3B parameters), we carry out continued causal language modelling training on \texttt{SLTrans}, forcing the Code-LMs to (1) learn the IR language and (2) align the IR constructs with respective constructs of various programming languages. 
Our resulting models, dubbed \textbf{\texttt{IRCoder}}, display sizeable and consistent gains across a wide variety of code generation tasks and metrics, including prompt robustness, multilingual code completion, code understanding, and instruction following.
\end{abstract}

\section{Introduction}
\label{sec:intro}

Language models for code generation (Code-LMs) are some of the most promising tools for enhancing the productivity of software developers. They have proliferated into automating several parts of the traditional software development lifecycle including code infilling, comment generation, refactoring, and build error prediction~\cite{didact2024google, dunay2024multiline}, inter alia. Despite a strong demand for such capabilities across all programming languages, the benchmarking of Code-LMs has largely been dominated by the most resourced languages. For instance, popular benchmarks such as HumanEval~\cite{chen2021evaluating}, MBPP~\cite{austin2021program} and APPS~\cite{ hendrycks2021measuring} all test Code-LMs' competence only in \texttt
{Python}: this can result in misleading conclusions on the global utility of Code-LMs.
More recent transpilation-oriented benchmarks like Multipl-E~\cite{cassano2022multipl} and BabelCode~\cite{pmlr-v202-orlanski23a}---that test competence on several languages---have laid bare the gaps in Code-LMs' performance across different programming languages. For instance, the state-of-the-art DeepSeekCoder's~\cite{guo2024deepseek} code completion performance in \texttt{Bash}, the most popular shell scripting language, lags its \texttt{Python pass@1} performance by 30\%+ points.

\begin{figure}
     \centering
     \begin{subfigure}[b]{0.49\linewidth}
         \centering
         \includegraphics[width=\linewidth, height=3.9cm]{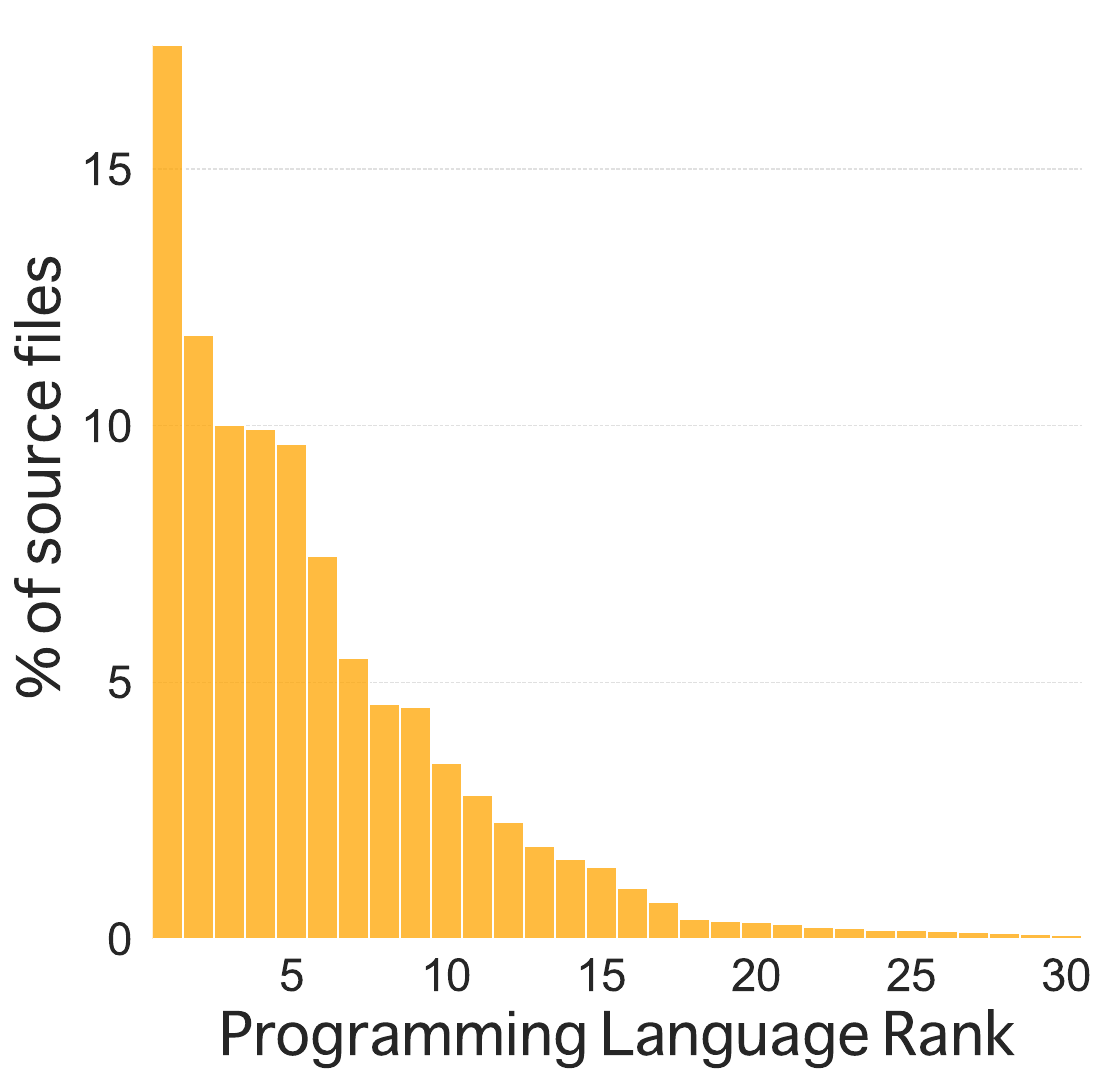}
     \end{subfigure}
     \hfill
     \begin{subfigure}[b]{0.49\linewidth}
         \centering
         \includegraphics[width=\linewidth, height=3.9cm]{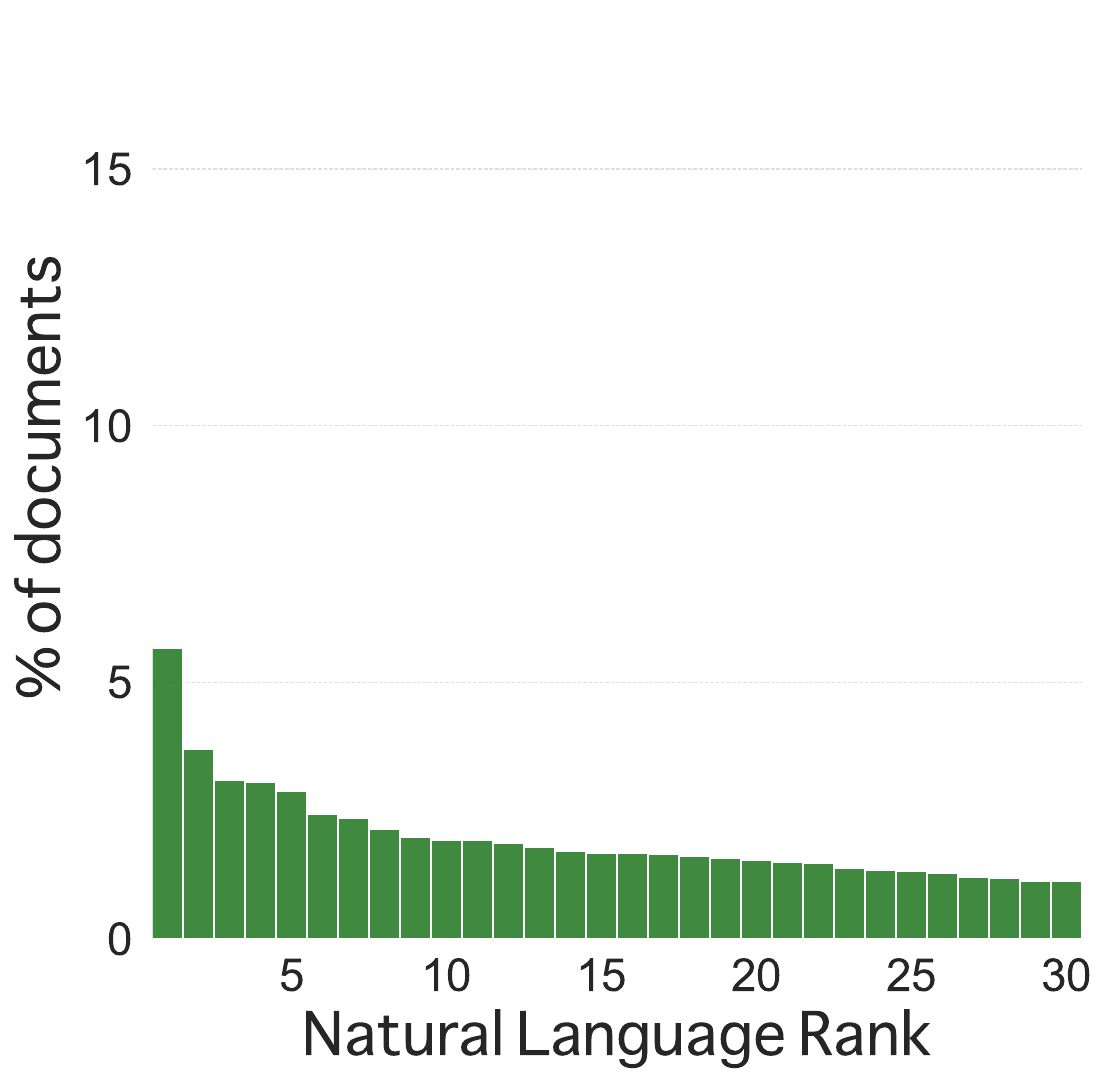}
     \end{subfigure}
     \vspace{-1.em}
        \caption{Comparison of the distribution of the top 30 programming languages on GitHub (left) against the top 30 natural languages in the mC4 corpus (right).}
        \label{fig:pl_nl_dist}
\end{figure}

The problem is exacerbated by the fact that the distribution of programming languages in code corpora is far more skewed than the distribution of natural languages in standard multilingual text corpora. 
As an example, Ukrainian, considered to be a moderate-to-low-resource natural language~\cite{tracey-etal-2019-corpus}, comprises a higher proportion of the massively multilingual mC4 corpus \cite{xue-etal-2021-mt5} than \texttt{Rust} (the 13th most popular programming language) does of GitHub\footnote{\href{https://madnight.github.io/githut}{GitHub Language Stats}}. This relative scarcity in code corpora, however, belies \texttt{Rust's} criticality to digital systems: Rust is one of only two languages approved for use in Linux kernel development\footnote{\href{https://lore.kernel.org/lkml/202210010816.1317F2C@keescook/}{Linux Kernel Lore: Rust introduction for v6.1-rc1}}. \cref{fig:pl_nl_dist} illustrates this problem by comparing relative distributions of natural and programming languages in respective multilingual corpora.  
Moreover, unlike the spread of digital content over natural languages, global code distribution over programming languages changes rapidly, reflecting sudden gains or drops in the popularity of individual programming languages. 
Such changes mean that consequential programming languages at some point in time, may not have been represented in the pre-training corpora of the Code-LMs. One prominent example is \texttt{HCL}, the fastest growing programming language according to GitHub\footnote{\href{https://octoverse.github.com/2022/top-programming-languages}{Octoverse Report 2022}}, used for configuring production infrastructure deployments, which did not make it into common pre-training corpora for Code-LMs~\cite{li2023starcoder}.

\vspace{-.5em}
\paragraph{The Case for Intermediate Code Representation.} 
The aforementioned limitations and properties of multilingual code corpora, i.e., skewed and rapidly changing distribution over programming languages, warrant a departure from the conventional approach of pre-training on ever-larger file-level source-code corpora. Indeed, recent evidence points to tangible downstream gains from the adoption of smaller but curated or synthesized data~\cite{gunasekar2023textbooks} as well as from grounding code generation using metadata from language toolchains\footnote{Toolchain is a set of tools required to create functional software, e.g., \textit{compiler}, \textit{linker}, \textit{libraries}, or \textit{debugger}.}~\cite{chen-etal-2023-pass, gong2024ast}. The latter, in principle, allows one to tap into more than half a century of research on programming languages and compilers and utilize views of the source code that often contain additional or more explicitly laid out information. This makes intuitive sense: skewed and fast-evolving distribution of programming languages implies that truly robust multilingual models cannot be obtained from heterogeneous source code alone; instead, some type of \textit{code interlingua} should be leveraged to facilitate cross-lingual transfer from high- to low(er)-resource languages.

In this work, we propose \textit{compiler intermediate representations} (IR) to be this interlingua for grounding source code understanding across heterogeneous languages, allowing for the creation of all-around stronger models. The IR are the artifacts of transformations performed by the compiler in three sequential phases: frontend, middle-end, and backend transformations. In popular cross-language compiler frameworks, the frontend IR contains language-specific constructs, whereas the backend IR contains the target platform-specific execution constructs. The middle-end IR, however, is agnostic to the source programming language and target execution platform and thus represents, we argue, an ideal shared representation for positive knowledge transfer in multilingual Code-LMs, offering both (1) a way to better semantically align constructs from heterogeneous languages and (2) an alternative (and possibly more informative) view of the source code. 
\vspace{-.5em}
\paragraph{Contributions and Research Questions.} 
Our work makes the following contributions:

    \vspace{0.15em} \textbf{1)} We create \texttt{SLTrans}, a parallel dataset consisting of nearly 4M pairs of  self-contained source code and corresponding IR;
    
    \vspace{0.15em} \textbf{2)} We conduct a systematic investigation of the benefits of grounding Code-LMs in IR, demonstrating sizeable and consistent empirical gains across a broad range of tasks and programming languages;
    
    \vspace{0.15em} \textbf{3)} We create and publicly release a suite of base and instruction-tuned Code-LMs dubbed \texttt{IRCoder}, the result of continued pre-training of state-of-the-art Code-LMs, ranging in size from 1.1B to 7.3B, on a mixture of parallel data from \texttt{SLTrans} and monolingual data of source languages.

\vspace{0.2em}\noindent We test the effectiveness of grounding on IR, in creating all-around stronger Code-LMs by structuring our inquiry into the following research questions: 

    \vspace{0.15em}\noindent \textbf{RQ1:} Does training with explicit grounding via parallel source code-IR corpora provide benefits over continued pre-training on (unpaired) source code or IR alone? 
    
    \vspace{0.15em}\noindent \textbf{RQ2:} Does grounding on IR improve robustness to prompt perturbations common in human inputs?
    
    \vspace{0.15em}\noindent \textbf{RQ3:} Does training on parallel source-IR data improve multilingual performance on code completion and understanding, with IR driving the positive knowledge transfer? 
    
    \vspace{0.15em}\noindent \textbf{RQ4:} What effect does pre-training on IR have on multilingual instruction following?

\section{Related Work}
\vspace{-0.3em}
\label{sec:related}
We provide a concise overview of the three most pertinent lines of work: (1) high-quality pre-training data curation, (2) grounding with toolchain metadata, and (3) alignment across and cross-lingual transfer between languages.

\paragraph{Curated Data For Multilingual Generalization.} 
Curating high-quality and domain-specific data with instructional value leads to more sample-efficient LM pretraining: Phi-1~\cite{gunasekar2023textbooks}, for example, trained on as few as 7B tokens, performs on a par with models trained on hundreds of times more uncurated data. Curation alone, as \citet{cassano2023knowledge} show, does not suffice for multilingual Code-LMs to generalize to underrepresented and unseen programming languages. Instead, the authors  
resort to using a bare-bones testcase transpiler to translate synthetic testcases to the target language, validating the quality of the synthetic target language data generated this way. This finding is in line with results from~\cite{roziere2022leveraging}, where the benefits of such verification have been demonstrated for code translation.

The compiler IR that we leverage in this work is the result of several sequentially executed transformations that---inter alia---eliminate dead code, unroll loops, combine expressions, and inline subroutines and thus offer significant instructional value without the need for generating unit tests, as the transformations are guaranteed to preserve the correctness of the source code.

\vspace{1.4mm}
\noindent\textbf{Grounding in Toolchain Metadata.}
There exists an extensive body of work that leverages the structure of the code as well as information originating from artifacts of various stages of compilation to ground code generation. Starting with compiler frontend artifacts, attempts have been made to leverage Abstract Syntax Trees (ASTs) for grounding source code understanding by linearizing them and encoding with LSTMs~\cite{jiang2022ast}, GNNs~\cite{zhang2022learning}, CNNs~\cite{mou2016convolutional}, Transformers~\cite{guo-etal-2022-unixcoder}, or some combination thereof~\cite{sun2020treegen}. Other modes of reliance on ASTs include (i) using them as a search prior for graph-based decoding~\cite{brockschmidt2018generative}, (ii) predicting (heuristically selected) paths from the tree as an auxiliary pre-training objective~\cite{tipirneni2024structcoder} and, (iii) leveraging them for data augmentation: heuristic generatation of meaning-preserving transformations, leveraged for contrastive learning~\cite{jain-etal-2021-contrastive, quiring2019misleading, bahrami2021augmentedcode}. Other compiler frontend artifacts such as Data Flow Graphs (DFGs)~\cite{brauckmann2020compiler} and Control Flow Graphs (CFGs)~\cite{nair2020funcgnn} have also been employed in grounding program understanding. Finally, there is work~\cite{shojaee2023executionbased, le2022coderl} that derives the reward that guides program generation via reinforcement learning (RL) from AST, CFG, and DFG matches between the generated and reference code. 

On the opposite end, compiler backend outputs have also been employed to ground Code-LMs, with compilation feedback in text form being favored by several recent efforts~\cite{jiang2023selfevolve, chen2023teaching, gou2023critic} to guide refining of tentative program generations. Concurrent work~\cite{liu2023rltf} has proposed to create an RL reward to guide generation based on the kind and severity of compilation error outputs. 

Finally, several existing efforts also leverage the IR produced by the compiler middle-end during its optimization passes, with LLVM being the most frequent choice of IR choice. IRGen~\cite{li2022unleashing} performs an exploratory study into using the IR itself as a meaning-preserving augmentation to perform contrastive learning on \texttt{C} source code. MulCS~\cite{ma2023mulcs} reports improvements to multilingual code snippet search when the GNN encoder utilizes a custom semantic graph derived from the IR. In the work most closely related to ours, \citet{szafraniec2023code} address code translation between four languages, pre-training the translation model using a wide variety of objectives, including source code to IR translation. Their effort, however, is limited to code translation (i.e., they do not consider any other task) and parallel source-to-IR data only at the function level (i.e., short context). In this work, in contrast, we investigate the general utility (i.e., for a wide range of downstream tasks) of pre-training multilingual Code-LMs using parallel source-to-IR data, scaling additionally up the data collection effort to (i) 12 programming languages and, importantly, (ii) self-contained file-level programs, which, intuitively, allows for grounding of many more source-code concepts (e.g., those instantiated with longer code spans) in IR. Importantly, we demonstrate that standard LM training on parallel source-to-IR data alone improves the robustness and multilingual ability of Code-LMs, without any architectural interventions and training auxiliary objectives.

\vspace{1.4mm}
\noindent\textbf{Cross-lingual Transfer and Alignment.} 
Most mainstream code generation models~\cite{li2023starcoder, guo2024deepseek, nijkamp2023codegen, roziere2023code, chai-etal-2023-ernie}, due to being pre-trained on GitHub code, are multilingual by default. Hence, they are subject to the \textit{curse of multilinguality} i.e. the degradation of model performance on high resource languages when the number of training languages or the proportion of low resource data in the pre-training corpus of a multilingual model is scaled up. This is usually caused by negative interference between unrelated languages to which the model can only allocate a fixed capacity~\cite{lauscher-etal-2020-zero, wu-dredze-2020-languages} and is a well-documented phenomenon in natural language models~\cite{conneau-etal-2020-unsupervised, arivazhagan2019massively}. Attempts to circumvent it without scaling up the model to impractical sizes have resorted to sparsity~\cite{ansell-etal-2022-composable,lee-hwang-2023-multilingual}, modularity~\cite{pfeiffer-etal-2022-lifting} and model merging~\cite{blevins2024breaking}. While the presence of similar phenomena has been verified in multilingual Code-LMs~\cite{pmlr-v202-orlanski23a, athiwaratkun2023multilingual}, research into cross-lingual transfer and alignment across programming languages has been rather sparse, exploring a limited set of tasks and languages~\cite{chen2022transferability} or introducing task specific architectural interventions~\cite{yuan2022circle, pian2023metatptrans} which are hard to scale. 

\newcite{wang-etal-2020-negative} indicate that separation of model parameters into 
language-agnostic and language-specific subsets can result in language-specific parameters being the cause of negative interference. This, we believe, presents an opportunity to minimize such interference by means of a shared intermediate representation rather than language-specific parameters. While it is unclear what such representation would be in the case of natural languages, intermediate compiler representations make an obvious choice for programming languages. 
Grounding Code-LM pretraining on IR data, we believe, should also and improve generalization (including to languages unseen in pretraining) and consequently facilitate cross-lingual transfer in downstream tasks, akin to 
cross-lingual transfer between non-English languages by models trained on English-centric bi-texts~\cite{gao-etal-2023-learning-multilingual,artetxe-schwenk-2019-massively}. Our experiments on a large array of tasks and programming languages show that this indeed is the case. 

\section{\texttt{SLTrans}: A \underline{\texttt{S}}ource Code to \underline{\texttt{L}}LVM IR \underline{\texttt{Trans}}lation Pairs Dataset}
\label{sec:sltrans}

In order to test the hypotheses we posit in \cref{sec:intro}, we seek to acquire parallel source-IR data for a mixture of low-, medium-, and high-resource programming languages. 

\begin{figure*}[t]
\centering
\includegraphics[width=\textwidth, height=6.5cm]{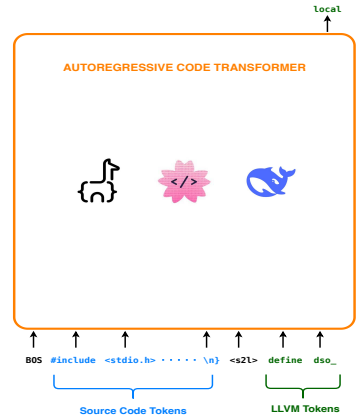}   
\label{fig:50k-view}
\vspace{-2.2em}
\caption{A high-level overview of our parallel data sourcing and training objective. Each source file is compiled via a corresponding LLVM frontend to obtain human-readable IR. The source and IR are then concatenated and the model is required to auto-regressively predict the tokens of one with the other in context thus aligning the constructs in the respective programming language with their analogues in LLVM.}
\end{figure*}

\paragraph{Intermediate Code Representation.} We utilize LLVM~\citep{lattner2004llvm} as the intermediate representation of our choice because it possesses many of the qualities we deem beneficial for our objectives. LLVM is the most prevalent IR in existing code corpora~\cite{kocetkov2023the} and one of the few frameworks that maintain a well-developed human-readable IR standard\footnote{\url{https://llvm.org/docs/LangRef.html}} rendering its syntax and semantics learnable via language modelling. Additionally, LLVM is adopted as the target IR of many compiler frontends across several programming languages\footnote{\url{https://llvm.org/ProjectsWithLLVM/}} mainly due to the ease with which its tooling infrastructure enables upstart languages to attain general availability. 
\vspace{-0.3em}
\begin{table}[h]
    \centering
    \renewcommand{\arraystretch}{1.12}
    \scalebox{0.66}{
    \begin{tabular}{llrrr}
        \toprule
        
        \multirow{3}{*}{\texttt{Language}} & \multirow{3}{*}{\texttt{Frontend}} & \multirow{3}{*}{\begin{tabular}{c}
                   \texttt{Avg. Len.}\\
                   \texttt{Multiplier}\end{tabular}} & \multicolumn{2}{c}{\texttt{No. Samples}}\\
        \cmidrule{4-5}
        
        &  &  & \texttt{Opt-Level} & \texttt{Opt-Level}\\
        
        &  &  & \multicolumn{1}{c}{\texttt{-Oz}} & \multicolumn{1}{c}{\texttt{-O3}}\\
        
        \midrule
        
        \scalerel*{\includegraphics{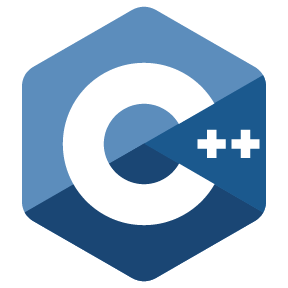}}{B} \texttt{C++} & \scalerel*{\includegraphics{Graphics/github.png}}{B} \texttt{\href{https://github.com/llvm/llvm-project/tree/main/clang}{clang}} & \texttt{5.08x} & \texttt{2,956,611} & \texttt{2,897,477}\\
        
        \scalerel*{\includegraphics{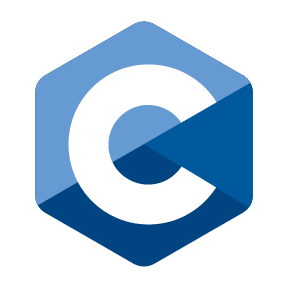}}{B} \texttt{C} &  \scalerel*{\includegraphics{Graphics/github.png}}{B} \texttt{\href{https://github.com/llvm/llvm-project/tree/main/clang}{clang}} & \texttt{3.26x} & \texttt{419,227} & \texttt{411,332}\\ 
        
        \scalerel*{\includegraphics{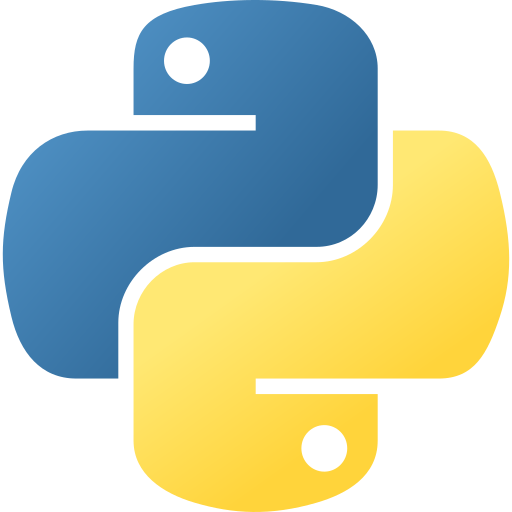}}{B} \texttt{Python}\tablefootnote{We source \texttt{Python} data via a \texttt{Codon}, which implements a statically typed subset of the \texttt{Python} language specification}  &  \multirow{2}{*}{\scalerel*{\includegraphics{Graphics/github.png}}{B} \texttt{\href{https://github.com/exaloop/codon}{codon}}} & \multirow{2}{*}{\texttt{11.43x}} & \multirow{2}{*}{\texttt{291,011}} & \multirow{2}{*}{\texttt{284,676}}\vspace{-4pt}\\ 
        
        [\scalerel*{\includegraphics{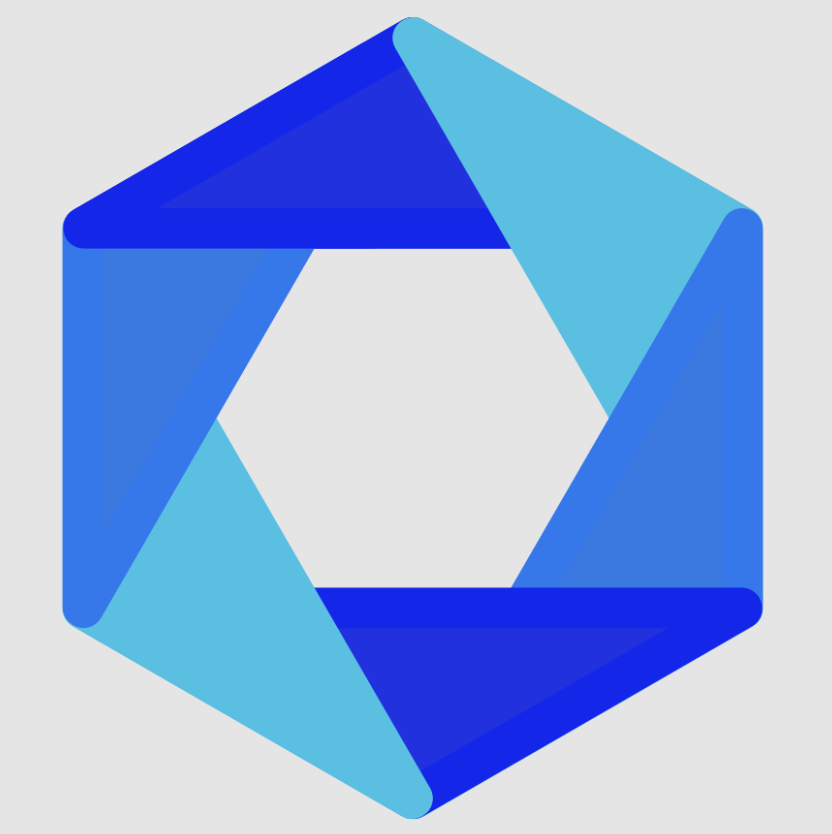}}{B} \texttt{Codon}] &   &  & \\

        \scalerel*{\includegraphics{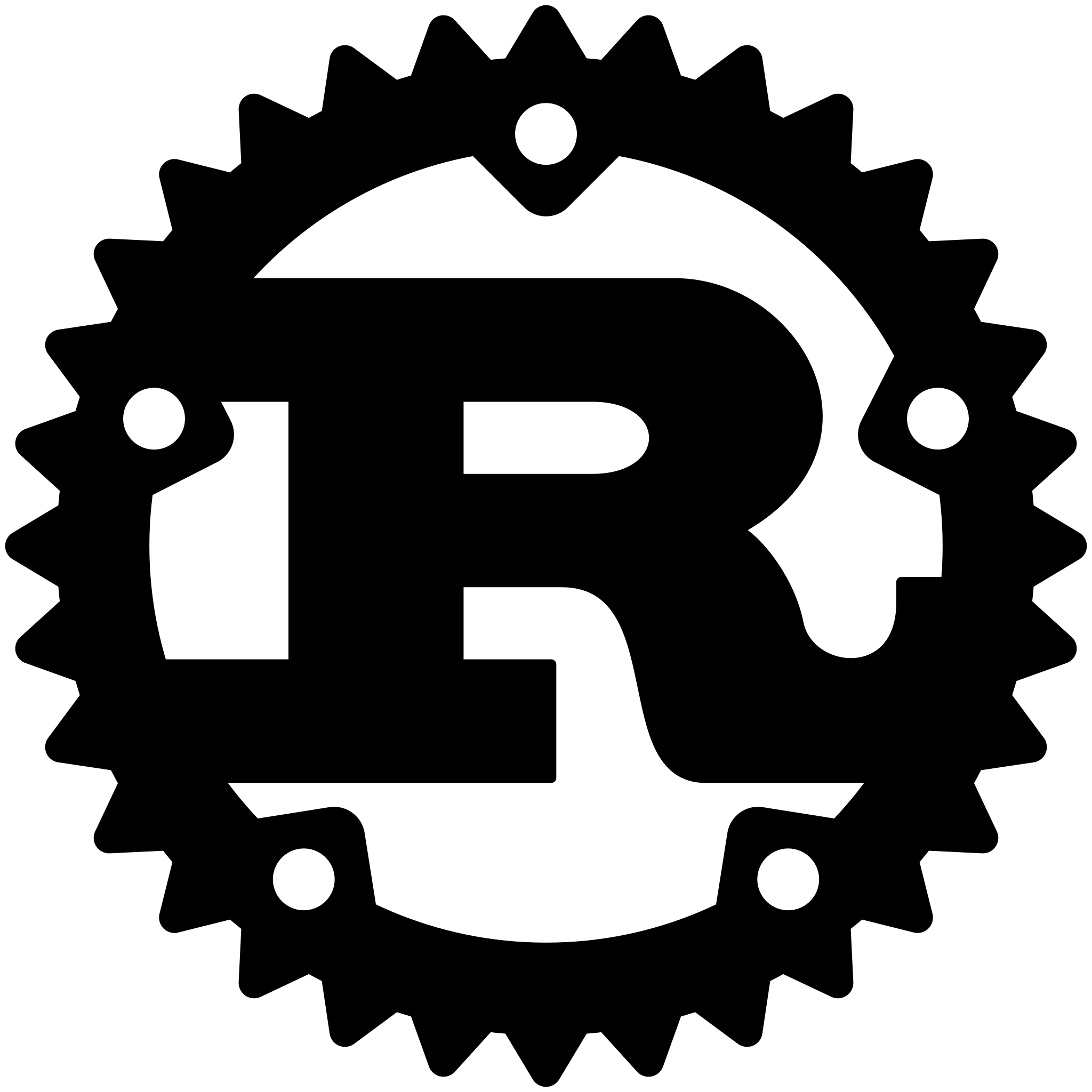}}{B} \texttt{Rust} & \scalerel*{\includegraphics{Graphics/github.png}}{B} \texttt{\href{https://github.com/rust-lang/rust/tree/master/compiler}{rustc}} & \texttt{21.36x} & \texttt{82,667} & \texttt{74,689}\\
        
        \scalerel*{\includegraphics{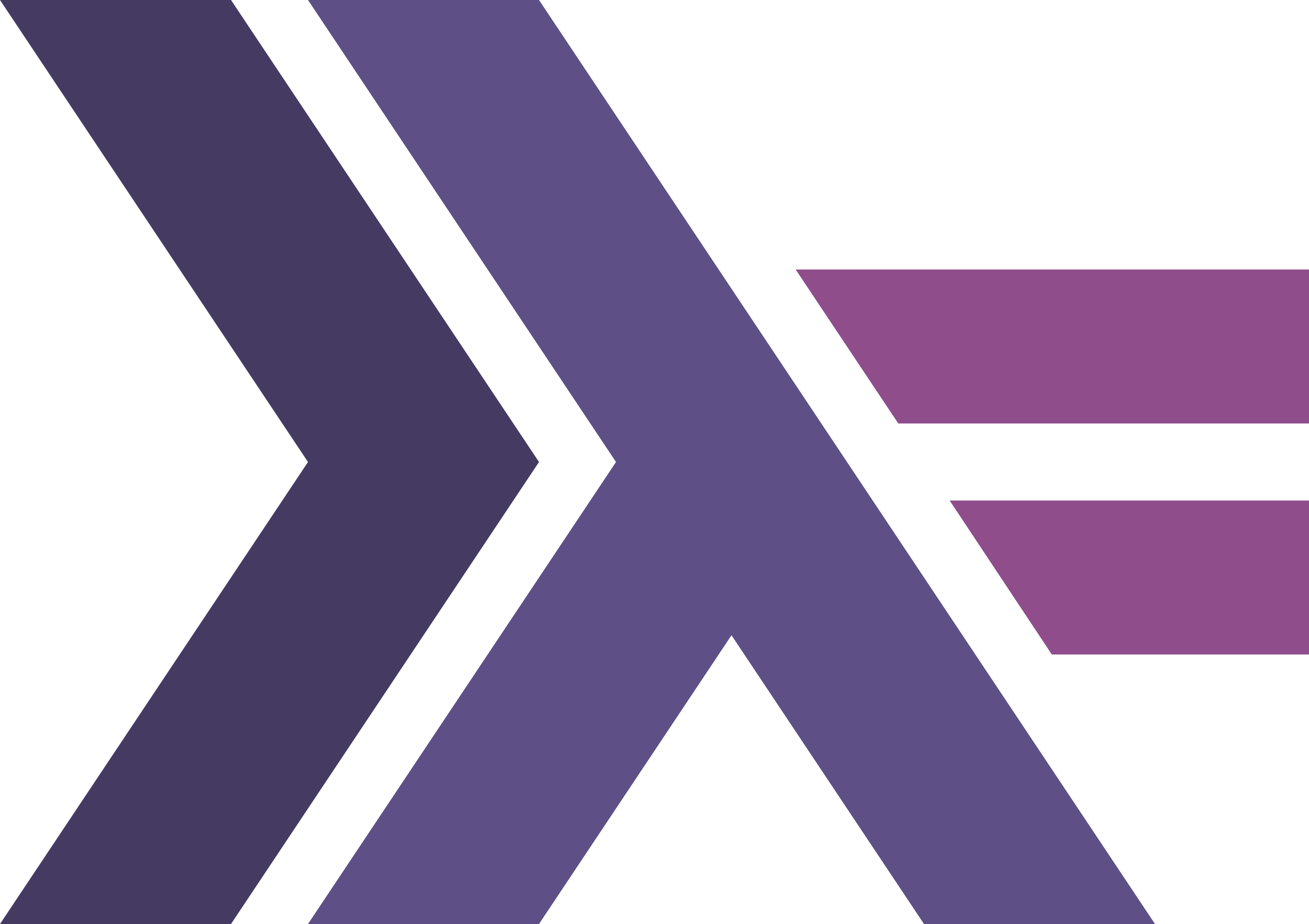}}{B} \texttt{Haskell} & \scalerel*{\includegraphics{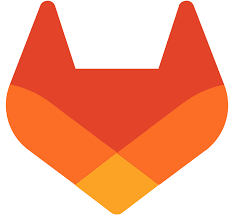}}{B} \texttt{\href{https://gitlab.haskell.org/ghc/ghc}{ghc}}  & \texttt{16.58x} & \texttt{61,483} & \texttt{59,378}\\
        
        \scalerel*{\includegraphics{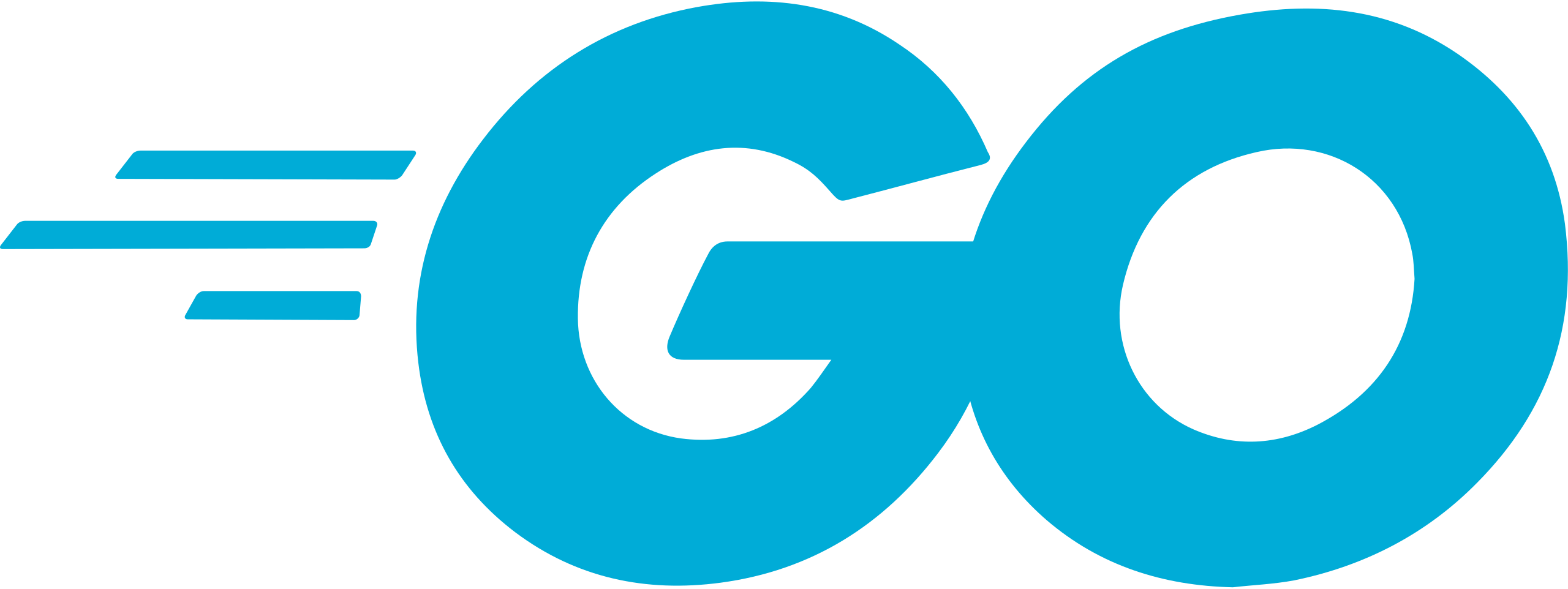}}{B} \texttt{Go} &  \scalerel*{\includegraphics{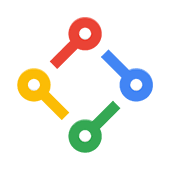}}{B} \texttt{\href{https://go.googlesource.com/gollvm/}{gollvm}} & \texttt{13.87x} & \texttt{55,578} & \texttt{42,241}\\
        
        \scalerel*{\includegraphics{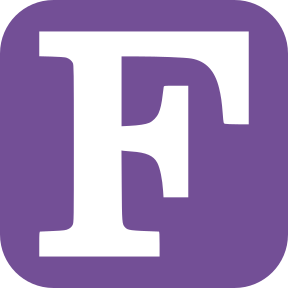}}{B} \texttt{Fortran} & \scalerel*{\includegraphics{Graphics/github.png}}{B} \texttt{\href{https://github.com/llvm/llvm-project/tree/main/flang}{flang}} & \texttt{4.59} & \texttt{35,288} & \texttt{31,299}\\
        
        \scalerel*{\includegraphics{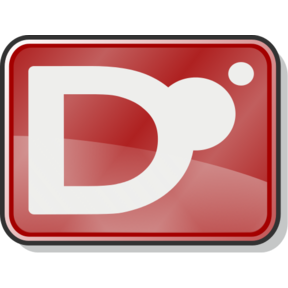}}{B} \texttt{D} & \scalerel*{\includegraphics{Graphics/github.png}}{B} \texttt{\href{https://github.com/ldc-developers/ldc}{ldc}} & \texttt{26.11x} & \texttt{18,111} & \texttt{6,125}\\
                
        \scalerel*{\includegraphics{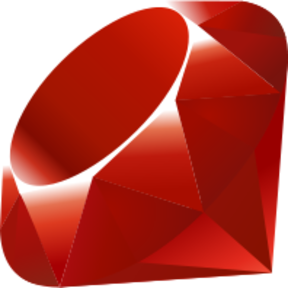}}{B} \texttt{Ruby}\tablefootnote{We source \texttt{Ruby} samples via \texttt{Crystal} --- a statically typed and compiled derivative of the language}  &  \multirow{2}{*}{\scalerel*{\includegraphics{Graphics/github.png}}{B} \texttt{\href{https://github.com/crystal-lang/crystal}{crystal}}} & \multirow{2}{*}{\texttt{6.78x}} & \multirow{2}{*}{\texttt{13,949}} & \multirow{2}{*}{\texttt{5,787}}\vspace{-4pt}\\ 
        
        [\scalerel*{\includegraphics{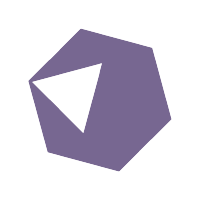}}{B} \texttt{Crystal}] &   &  & \\
        
        \scalerel*{\includegraphics{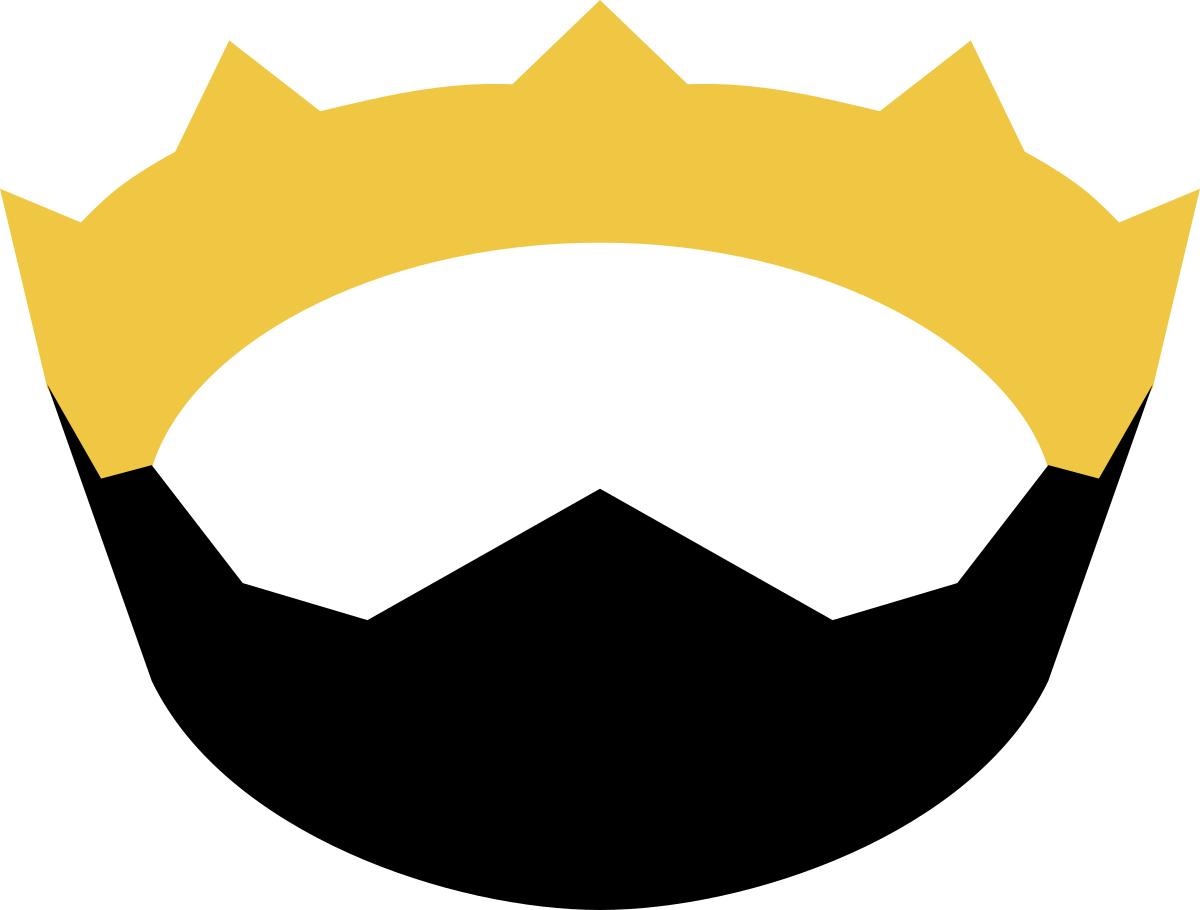}}{B} \texttt{Nim} & \scalerel*{\includegraphics{Graphics/github.png}}{B} \texttt{\href{https://github.com/arnetheduck/nlvm}{nlvm}} & \texttt{18.84x} &  \texttt{2,865} & \texttt{-}\\
        
        \scalerel*{\includegraphics{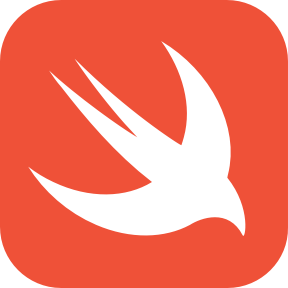}}{B} \texttt{Swift} & \scalerel*{\includegraphics{Graphics/github.png}}{B} \texttt{\href{https://github.com/apple/swift/tree/main/SwiftCompilerSources/Sources}{swiftc}} & \texttt{8.79x} & \texttt{2,179} & \texttt{1,354}\\
        
        \scalerel*{\includegraphics{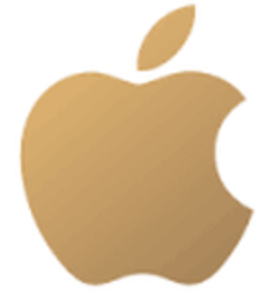}}{B} \texttt{Obj-C} & \scalerel*{\includegraphics{Graphics/github.png}}{B} \texttt{\href{https://github.com/llvm/llvm-project/tree/main/clang}{clang}} & \texttt{3.88x} & \texttt{403} & \texttt{261}\\
        
        \midrule
        & \texttt{Total:} &  & \texttt{3,939,372} & \texttt{3,814,619}\\
        
        \bottomrule
    \end{tabular}
    }
    \vspace{-.6em}
    \caption{Breakdown of \texttt{SLTrans}  across programming languages (with respective compiler frontends).}
        \label{table:slpairs}
        
\end{table}

\vspace{-.9em}
\noindent\textbf{\texttt{SLTrans} Creation.} Extracting LLVM IR from free-form source code in GitHub demands compilable and complete code units, the collection of which comes with several challenges. 
The proportion of compilable code units in free-form code is abysmally low due to the need for tracking dependencies. Many languages such as \texttt{C} and \texttt{C++} do not have mature package management systems, which makes following dependency paths across repository boundaries virtually impossible. 
The problem is exacerbated by the difficulty of reliably following within-repository file-level dependencies due to aggressive de-duplication of source files during curation of language modelling code corpora for performance reasons~\cite{allamanis2019adverse, lee-etal-2022-deduplicating}: this mangles the repository structure\footnote{A particularly common scenario is a popular repository with several forks being torn apart as the de-duplication pipeline non-deterministically selects disjoint subsets of files from different forks}. 
Additionally, there are also obstacles to obtaining \textit{complete} compilation units. Languages such as \texttt{Rust}, \texttt{Go} or \texttt{Swift} simply cannot be compiled at the file level (unless the files are self-contained) as their respective LLVM frontends operate on module or package-level compilation units. As of this writing, multi-file repository-level code language modeling is unsupported by most mainstream code models~\cite{roziere2023code, li2023starcoder, nijkamp2023codegen}. 
As a result, prior attempts at extracting parallel source-IR data have been stymied by the need to implement language-specific mechanisms to track dependency code for successful compilation~\citep{grossman2023compile} and thus mostly restricted to function/snippet level code~\citep{szafraniec2023code}, which is very limiting in terms of coverage of language constructs~\cite{li2022unleashing}. 

We sidestep the above issues by sourcing self-contained compilation units from accepted solutions to programming contest problems~\citep{rosetta-code, codeforces2020, puri2021codenet, Caballero_Description2Code_Dataset_2016}, which typically do not have cross-file dependencies. We then compile these source files into two IR flavours: size-optimized (\texttt{-Oz} opt-level equivalent) IR and performance-optimized (\texttt{-O3} opt-level equivalent) IR. We further filter only samples with IR shorter than 2500 code lines. The size-optimized IR  allows for larger context windows in LLM inference and is also more uniform across languages; 
being used for deployment, the performance-optimized IR is more prevalent in open-domain code corpora. 
Collecting both enables fine-grained trade-offs between the two IR forms during language modelling. Finally, given the abundance of near-duplicates in programming contest solutions, we perform MinHash-based~\cite{broder1997resemblance} de-duplication. The final dataset, dubbed \texttt{SLTrans}, consists of ca.\,4M samples across 12 programming languages, totalling 26.2B tokens.\footnote{Based on the StarCoderBase tokenizer.} A breakdown of \texttt{SLTrans} is given in \cref{table:slpairs}.


\vspace{-0.3em}
\section{Experimental Setup}
\vspace{-0.5em}
\label{sec:setup}
\noindent\textbf{Data Preparation.}
\label{subsec:datacp}
We leverage LLVM IR to ground matching constructs across heterogeneous languages and facilitate cross-lingual transfer: as header data (along with superfluous platform, vendor, and memory layout information) does not contribute to this goal, we remove it from IR before pairing with source code. We choose the size-optimized IR 80\% of the time and performance-optimized IR for 20\% of the training samples.

Given our computational budget, we could afford to perform continued pretraining on IR-grounded code on approximately 1.5B tokens. Given that (i) this is substantially smaller than the overall size of \texttt{SLTrans} and (ii) acknowledging the skewed language distribution of the dataset, we sub-sample the training corpora using token-level UniMax-1 sampling~\cite{chung2023unimax}, based on the StarCoderBase tokenizer \cite{li2023starcoder}. 
We select a token budget of 600M tokens this way. We next source 200M tokens of unpaired open-domain IR code from TheStack~\cite{kocetkov2023the}, to allow Code-LMs to better learn the IR itself. Finally, to avoid catastrophic forgetting of pre-trained source-code knowledge, we spend the remaining 700M tokens of our budget on high-quality code and text data: these include math articles from the OpenWebMath dataset~\cite{paster2023openwebmath}, research articles from the PeS2o dataset~\cite{peS2o}, source code sampled from language splits in TheStack present in \texttt{SLTrans} and single-file changing GitHub commits\footnote{We source contents before and after the change, along with the commit message}. 
The breakdown of the final dataset, to which we refer with Paired, is given in \Cref{table:data-abl}.
\vspace{-0.2em}

\begin{table}[h]
    \centering
    \scalebox{0.66}{
    \begin{tabular}{lccc}
        \toprule
        \scalerel*{\includegraphics{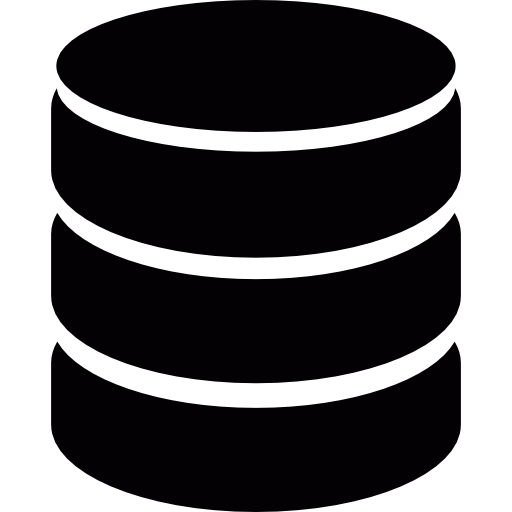}}{B} \texttt{Data Mix} & \texttt{CodeText} & \texttt{Unpaired} & \texttt{Paired} \\
        \midrule

        \scalerel*{\includegraphics{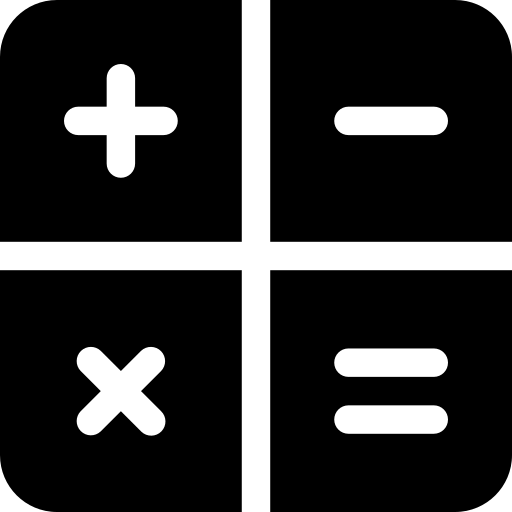}}{B} \texttt{OpenWebMath} & \texttt{300M} & \texttt{200M} & \texttt{100M} \\
        \scalerel*{\includegraphics{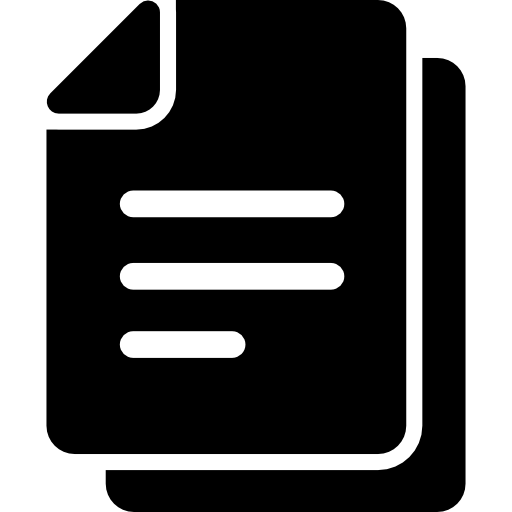}}{B} \texttt{PeS2o} & \texttt{500M} & \texttt{300M} & \texttt{200M} \\
        \scalerel*{\includegraphics{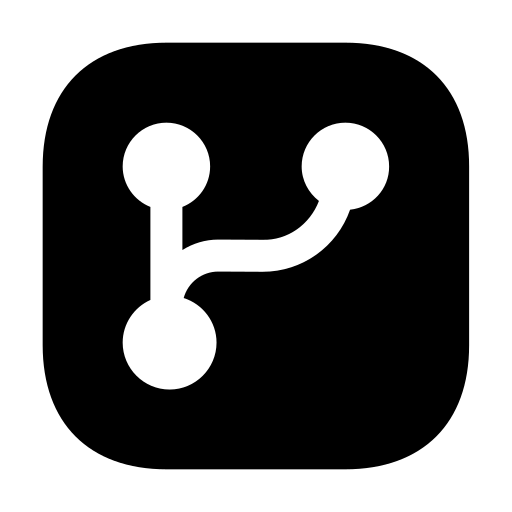}}{B} \texttt{Git Commits} & \texttt{200M} & \texttt{150M} & \texttt{100M} \\
        \scalerel*{\includegraphics{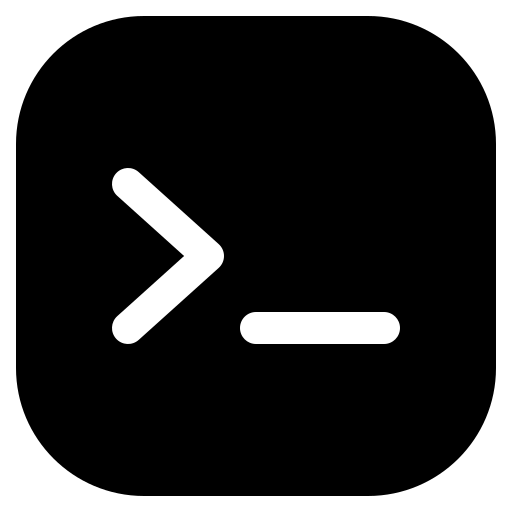}}{B} \texttt{TheStack} & \texttt{500M} & \texttt{400M} & \texttt{300M} \\
        \scalerel*{\includegraphics{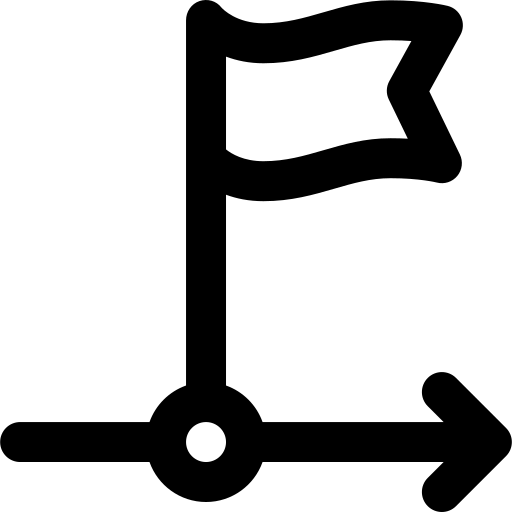}}{B} \texttt{Unpaired IR} & \texttt{-} & \texttt{450M} & \texttt{200M} \\
        \scalerel*{\includegraphics{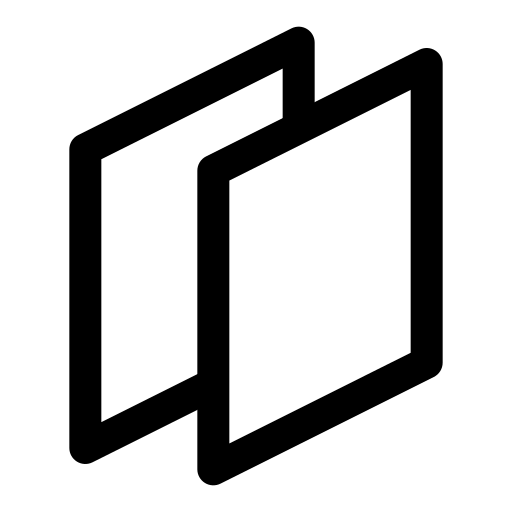}}{B} \texttt{Source-IR Pairs (SLTrans)} & \texttt{-} & \texttt{-} & \texttt{600M} \\
        \midrule
        \texttt{Total: } & \texttt{1.5B} & \texttt{1.5B} & \texttt{1.5B} \\
        \bottomrule
    \end{tabular}
    }
    \vspace{-.3em}
    \caption{Token counts for the paired, unpaired, and codetext pre-training dataset (StarCoderBase tokenizer).}
        \label{table:data-abl}
\end{table}

\vspace{-0.5em}
\noindent\textbf{Model Training.}
\label{subsec:cont-lm} Aiming for robust findings, we test the effects of IR grounding on six different Code-LMs from three different providers, ranging in size from 1.1B to 7.3B parameters: StarCoderBase~\cite{li2023starcoder} 1.1B, 3.1B, and 7.3B; DeepSeekCoder~\cite{guo2024deepseek} 1.3B and 5.7B; and CodeLlama~\cite{roziere2023code} 6.7B. 

We perform continued LM training for each of these models on the Paired dataset built from \texttt{SLTrans}.   
We introduce two new sentinel tokens--- \texttt{<s2l>} and \texttt{<l2s>}---into the models' vocabulary and use them, respectively, for two possible directions of grounding (each sampled for 50\% of training instances):

\begin{center}    
\begin{BVerbatim}[commandchars=!\{\}]
!textcolor{Navy}{source_code} <s2l> !textcolor{Saffron}{llvm_ir} <|EOS|>
\end{BVerbatim}    
\end{center}

\vspace{-1em}
\begin{center}    
\begin{BVerbatim}[commandchars=!\{\}]
!textcolor{Saffron}{llvm_ir} <l2s> !textcolor{Navy}{source_code} <|EOS|>
\end{BVerbatim}
\end{center}

\noindent We randomly initialize the sentinel tokens' embeddings from a Gaussian distribution with the mean set to the average of all pre-trained vocabulary embeddings and retain the variance from the models' initializer configurations. 

We rely on LoRA~\cite{hu2022lora} for parameter-efficient continued pre-training (we set \texttt{r} to 256 and an $\alpha$ to 128), while keeping the embedding layers trainable. We resort to DeepSpeed~\cite{rasley2020deepspeed} Zero Stage-2 to accelerate our training jobs. We train with a maximum sequence length of 4096 tokens using the Adam~\cite{KingBa15} optimizer ($\beta$ values of (0.95, 0.99)) with a base learning rate of $1e-4$ for the LoRA modules and $4e-5$ for the embedding layers, employing a cosine schedule (culminates at 10\% of the base).

\vspace{-0.5em}
\section{Results and Discussion}
\label{sec:results}
\vspace{-0.5em}
\noindent\textbf{\textbf{RQ1}: Pairing source code and IR matters.}
\label{subsec:rq1}
We first test whether the grounding of source code in IR, i.e., language modelling on paired source-IR instances, actually matters.  
To this end, we compare the performance of models trained on the Paired data against counterparts trained on (1) the unpaired concatenation of code and IR data (dubbed Unpaired) and (2) just more source-code data (referred to as CodeText). CodeText is derived from the same sources as Paired (see~\Cref{subsec:datacp}) but it does not contain any (paired or unpaired) LLVM IR data: instead, we simply upsample other sources to reach 1.5B tokens. 
The Unpaired is upsampled from all sources except the source-IR pairs from our \texttt{SLTrans}, i.e., compared to CodeText, it additionally samples training examples from the 450M token corpus of (unpaired) LLVM IR code from TheStack. 
\Cref{table:data-abl} details the compositions of Unpaired, and CodeText corpora against our primary Paired corpus. 

\begin{table}[t]
    \centering
    \scalebox{0.66}{
    \begin{tabular}{rcccc}
        \toprule
        \multirow{3}{*}{\scalerel*{\includegraphics{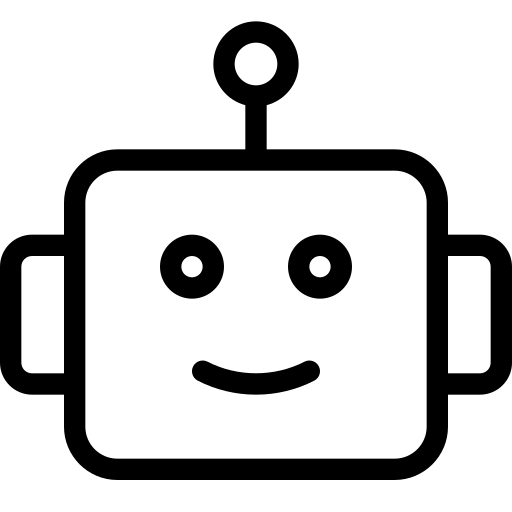}}{B} \texttt{Model}} & \multirow{3}{*}{\texttt{Base}} & \multicolumn{3}{c}{\texttt{Continued Pre-Training}}\\
        \cmidrule(lr){3-5} 
       &  & \multirow{2}{*}{\texttt{CodeText}} & \multirow{2}{*}{\texttt{Unpaired}} & \texttt{Paired} \\
       &  &  &  & \texttt{(IRCoder)} \\
        \midrule

        \multirow{2}{*}{\texttt{StarCoderBase 1.1B}} & \texttt{8.35} & \texttt{8.39} & \texttt{8.59} & \texttt{8.76} \vspace{-0.2em}\\
         & \textbf{\texttt{-}} & \diffup{0.04} & \diffup{0.24} & \diffup{0.41} \vspace{0.2em}\\
        \multirow{2}{*}{\texttt{DeepSeekCoder 1.3B}} & \texttt{18.34} & \texttt{18.22} & \texttt{18.77} & \texttt{20.51} \vspace{-0.2em}\\
         & \textbf{\texttt{-}} & \diffdo{0.12} & \diffup{0.43} & \diffup{2.17} \vspace{0.2em}\\
        \multirow{2}{*}{\texttt{StarCoderBase 3.1B}} & \texttt{12.78} & \texttt{12.75} & \texttt{12.97} & \texttt{14.36} \vspace{-0.2em}\\
         & \textbf{\texttt{-}} & \diffdo{0.03} & \diffup{0.19} & \diffup{1.58} \vspace{0.2em}\\
        \multirow{2}{*}{\texttt{DeepSeekCoder 5.7B}} & \texttt{30.47} & \texttt{30.22} & \texttt{30.44} & \texttt{31.14} \vspace{-0.2em}\\
        & \textbf{\texttt{-}} & \diffdo{0.25} & \diffdo{0.03} & \diffup{0.67} \vspace{0.2em}\\
        \multirow{2}{*}{\texttt{CodeLlama 6.7B}} & \texttt{21.83} & \texttt{21.94} & \texttt{22.14} & \texttt{24.06} \vspace{-0.2em}\\
        & \textbf{\texttt{-}} & \diffup{0.11} & \diffup{0.31} & \diffup{2.23} \vspace{0.2em}\\
        \multirow{2}{*}{\texttt{StarCoderBase 7.3B}} & \texttt{17.94} & \texttt{17.73} & \texttt{18.03} & \texttt{18.46} \vspace{-0.2em}\\
        & \textbf{\texttt{-}} & \diffdo{0.21} & \diffup{0.09} & \diffup{0.52} \vspace{0.2em}\\
        \bottomrule
    \end{tabular}
    }
    \caption{\textbf{RQ1}: Multipl-E \texttt{pass@1} all language average performance comparison between different continued pre-training settings.}
        \label{table:avg-abl}
        \vspace{0.5em}
\end{table}

We benchmark Code-LLMs additionally trained on these three corpora against their base variants on the Multipl-E~\cite{cassano2022multipl} code completion benchmark (with the \texttt{pass@1} metric), a transpiled expansion of the popular HumanEval~\cite{chen2021evaluating} benchmark to 18 programming languages. We limit our evaluation to the subset of languages present in \texttt{SLTrans}: \texttt{C++}, \texttt{D}, \texttt{Go}, \texttt{Python}, \texttt{Ruby}, \texttt{Rust} and \texttt{Swift}. 
Comparison of models' performance, displayed in   \Cref{table:avg-abl}, brings the key insights: while adding unpaired IR data can bring some performance gains (compare Unpaired vs. Base and CodeText), these gains are much less pronounced than the gains we obtain by adding paired source-IR data to the training mix (Paired vs. Unpaired). These results strongly suggest that grounding of heterogeneous source code languages in the same IR accounts for the majority of performance gains, and not the mere exposure to IR.    
Comparison between CodeText and Base reveals that continued training on the data distribution similar to that used in the original pre-training can hurt performance: most models additionally trained on CodeText exhibit small drops in performance compared to their Base variants.  
This observation is in line with prior findings \cite{cassano2023knowledge} and is likely the result of degradations caused by repeating data in language modeling~\cite{allamanis2019adverse}.\footnote{While the exact pre-training corpora of DeepSeekCoder and CodeLlama is unknown, their data collection pipelines promise large overlaps with our CodeText corpus.}


\begin{table}[t!]
    \centering
    \scalebox{0.66}{
    \begin{tabular}{rccc}
        \toprule
        \multirow{2}{*}{\scalerel*{\includegraphics{Graphics/robot.png}}{B} \texttt{Model}} & \multicolumn{3}{c}{\texttt{ReCode}} \\
        \cmidrule(lr){2-4} 
        & \texttt{Format} & \texttt{Syntax} & \texttt{Function}\\
        \midrule
         
        \texttt{StarCoderBase 1.1B} & \texttt{28.08} & \texttt{26.39} & \texttt{11.31}\\
        \texttt{DeepSeekCoder 1.3B} & \texttt{49.61} & \texttt{44.88} & \texttt{25.13}\\
        \texttt{StarCoderBase 3.1B} & \texttt{38.70} & \texttt{33.29} & \texttt{19.04}\\
        \texttt{DeepSeekCoder 5.7B} & \texttt{62.37} & \texttt{55.43} & \texttt{36.73}\\
        \texttt{CodeLlama 6.7B} & \texttt{54.50} & \texttt{45.23} & \texttt{24.49}\\
        \texttt{StarCoderBase 7.3B} & \texttt{46.30} & \texttt{41.50} & \texttt{23.53}\\
        \midrule
        \multirow{2}{*}{\texttt{IRCoder 1.1B}} & \texttt{30.18} & \texttt{27.50} & \texttt{12.01} \vspace{-0.2em}\\
        & \diffup{2.10} & \diffup{1.11} & \diffup{0.70}\vspace{0.2em}\\
        \multirow{2}{*}{\texttt{IRCoder 1.3B}} & \texttt{49.85} & \texttt{45.43} & \texttt{25.75}\vspace{-0.2em}\\
         & \diffup{0.24} & \diffup{0.55} & \diffup{0.63}\vspace{0.2em}\\
        \multirow{2}{*}{\texttt{IRCoder 3.1B}} & \texttt{39.78} & \texttt{34.42} & \texttt{18.80}\vspace{-0.2em}\\
         & \diffup{1.08} & \diffup{1.13} & \diffdo{0.24}\vspace{0.2em}\\
        \multirow{2}{*}{\texttt{IRCoder 5.7B}} & \texttt{65.76} & \texttt{59.24} & \texttt{38.66} \vspace{-0.2em}\\
         & \diffup{3.39} & \diffup{3.82} & \diffup{1.93}\vspace{0.2em}\\
        \multirow{2}{*}{\texttt{IRCoder 6.7B}} & \texttt{56.41} & \texttt{48.11} & \texttt{25.47}\vspace{-0.2em}\\
         & \diffup{1.91} & \diffup{2.88} & \diffup{0.98}\vspace{0.2em}\\
        \multirow{2}{*}{\texttt{IRCoder 7.3B}} & \texttt{46.62} & \texttt{41.82} &  \texttt{23.76}\vspace{-0.2em}\\
         & \diffup{0.32} & \diffup{0.32} & \diffup{0.23}\vspace{0.2em}\\
        \bottomrule
    \end{tabular}
    }
    \vspace{-0.3em}
    \caption{\textbf{RQ2}: ReCode split average \texttt{pass@1} comparison between \texttt{IRCoder} and the corresponding base models. For detailed perturbation-level breakdowns in the Format, Syntax, and Function splits refer to \Cref{table:recode-format,table:recode-syntax,table:recode-function} in the Appendix respectively.}
    \label{table:avg-rob}
\end{table}


\begin{table*}[t]
    \centering
    \scalebox{0.66}{
    \begin{tabular}{rcccccccc}
        \toprule
        \multirow{2}{*}{\scalerel*{\includegraphics{Graphics/robot.png}}{B} \texttt{Model}} & \multicolumn{3}{c}{\texttt{Multipl-E}} & \texttt{CodeXGLUE Code-Text} & \multicolumn{2}{c}{\texttt{Commit Chronicle}} & \multicolumn{2}{c}{\texttt{HumanEvalFixDocs}}\\
        \cmidrule(lr){2-4} \cmidrule(lr){5-5} \cmidrule(lr){6-7} \cmidrule(lr){8-9}
        & \texttt{pass@1} & \texttt{pass@10} & \texttt{pass@25} & \texttt{BLEU-4} & \texttt{ROUGE-2} & \texttt{ROUGE-L} & \texttt{pass@1} & \texttt{pass@10}\\
        \midrule
         
        \texttt{StarCoderBase 1.1B} & \texttt{8.35} & \texttt{13.43} & \texttt{16.43} & \texttt{10.05} & \texttt{12.41} & \texttt{33.67} & \texttt{12.23} & \texttt{17.70}\\
        \texttt{DeepSeekCoder 1.3B} & \texttt{18.34} & \texttt{26.12} & \texttt{31.36} & \texttt{9.63} & \texttt{12.33} & \texttt{33.16} & \texttt{25.48} & \texttt{36.22}\\
        \texttt{StarCoderBase 3.1B} & \texttt{12.78} & \texttt{19.09} & \texttt{22.62} & \texttt{10.61} & \texttt{14.35} & \texttt{33.70} & \texttt{28.44} & \texttt{42.91}\\
        \texttt{DeepSeekCoder 5.7B} & \texttt{30.47} & \texttt{41.38} & \texttt{48.04} & \texttt{11.80} & \texttt{13.77} & \texttt{35.18} & \texttt{48.21} & \texttt{61.05}\\
        \texttt{CodeLlama 6.7B} & \texttt{21.83} & \texttt{34.78} & \texttt{42.50} & \texttt{9.74} & \texttt{14.46} & \texttt{35.91} & \texttt{44.50} & \texttt{56.79}\\
        \texttt{StarCoderBase 7.3B} & \texttt{17.94} & \texttt{27.38} & \texttt{34.12} & \texttt{10.74} & \texttt{15.22} & \texttt{37.74} & \texttt{40.74} & \texttt{55.27}\\
        \midrule
        \multirow{2}{*}{\texttt{IRCoder 1.1B}} & \texttt{8.76} & \texttt{14.51} & \texttt{19.32}
 & \texttt{11.41} & \texttt{13.15} & \texttt{35.04} & \texttt{13.01} & \texttt{18.00} \vspace{-0.2em}\\
         & \diffup{0.41} & \diffup{1.08} & \diffup{2.89} & \diffup{1.36} & \diffup{0.73} & \diffup{1.38} & \diffup{0.78} & \diffup{0.30}\vspace{0.2em}\\
        \multirow{2}{*}{\texttt{IRCoder 1.3B}} & \texttt{20.51} & \texttt{31.14} & \texttt{37.90} & \texttt{10.79}
        & \texttt{13.12} & \texttt{34.57} & \texttt{27.07} & \texttt{37.95} \vspace{-0.2em}\\
         & \diffup{2.17} & \diffup{5.02} & \diffup{6.54} & \diffup{1.16} & \diffup{0.79} & \diffup{1.41} & \diffup{1.59} & \diffup{1.73}\vspace{0.2em}\\
        \multirow{2}{*}{\texttt{IRCoder 3.1B}} & \texttt{14.36} & \texttt{22.58} & \texttt{28.02} & \texttt{11.74} & \texttt{14.29} & \texttt{36.81} & \texttt{28.99} & \texttt{42.76} \vspace{-0.2em}\\
         & \diffup{1.58} & \diffup{3.49} & \diffup{5.39} & \diffup{1.13} & \diffdo{0.06} & \diffup{1.11} & \diffup{0.55} & \diffdo{0.15}\vspace{0.2em}\\
        \multirow{2}{*}{\texttt{IRCoder 5.7B}} & \texttt{31.14} & \texttt{45.00} & \texttt{51.29} & \texttt{13.21} & \texttt{14.71} & \texttt{37.15} & \texttt{49.79} & \texttt{66.35} \vspace{-0.2em}\\
         & \diffup{0.67} & \diffup{3.62} & \diffup{3.25} & \diffup{1.41} & \diffup{0.93} & \diffup{1.97} & \diffup{1.57} & \diffup{4.29}\vspace{0.2em}\\
        \multirow{2}{*}{\texttt{IRCoder 6.7B}} & \texttt{24.06} & \texttt{39.38} & \texttt{47.03} & \texttt{11.15} & \texttt{14.95} & \texttt{36.82} &  \texttt{46.59} & \texttt{58.74} \vspace{-0.2em}\\
         & \diffup{2.23} & \diffup{4.60} & \diffup{4.53} & \diffup{1.41} & \diffup{0.49} & \diffup{0.91} & \diffup{2.09} & \diffup{1.95} \vspace{0.2em}\\
        \multirow{2}{*}{\texttt{IRCoder 7.3B}} & \texttt{18.46} & \texttt{30.43} & \texttt{38.04} & \texttt{11.16} & \texttt{15.88} & \texttt{38.96} & \texttt{44.07} & \texttt{57.38} \vspace{-0.2em}\\
         & \diffup{0.52} & \diffup{3.05} & \diffup{3.92} & \diffup{0.42} & \diffup{0.67} & \diffup{1.22} & \diffup{3.33} & \diffup{2.11} \vspace{0.2em}\\
        \bottomrule
    \end{tabular}
    }
    \vspace{-0.4em}
    \caption{\textbf{RQ3} and \textbf{RQ4}: All language average performance comparison between \texttt{IRCoder} and the corresponding base models on multilingual tasks. For detailed language-wise breakdowns in Multipl-E results refer to \Cref{table:multiple-1,table:multiple-10,table:multiple-25}, CodeXGLUE code to text results refer to \Cref{table:code2text}, Commit Chronicle results refer to \Cref{table:cc-2,table:cc-L}, and HumanEvalFixDocs results refer to \Cref{table:hefixdocs-1,table:hefixdocs-10} in the Appendix.}
        \label{table:avg-mult}
\end{table*}

\vspace{.9mm}
\noindent\textbf{\textbf{RQ2}: Grounding in IR improves robustness to prompt perturbations.}
\label{subsec:rq2}
We next investigate how our source-IR grounding affects the perturbation robustness of Code-LMs. Such robustness is critical, as malformed and adversarial prompts have been shown to successfully lead to the generation of incorrect~\cite{zhou2022adversarial} and insecure code~\cite{dinh2023large, wu2023deceptprompt}. Our intuition is that grounding on IR should reduce the vulnerability of Code-LMs to such perturbations, as IR is the result of several transformations that tend to remove the effects of minor semantic variances or even mistakes in the source code. 
We test our hypothesis using 5 differently seeded ReCode~\cite{wang-etal-2023-recode} transformations of HumanEval to measure robustness to three classes of perturbations in \texttt{Python}: code formatting, syntactic variation, and function name mangling. 

As evidenced by the detailed results in \Cref{table:avg-rob}, \texttt{IRCoder} displays gains across the board, with particularly significant gains in robustness against syntactic variations that are typical for human-written prompts. 
Interestingly, the gains for robustness to function header mangling are substantially smaller. We believe that this is the artefact of the benchmark, abundant with prompts that include headers and docstrings, which may underestimate the functional robustness of IR grounding ``in the wild''.

\vspace{.9mm}\noindent\textbf{{RQ3}: IR grounding improves multilingual code understanding.}
We next test the multilingual code completion and understanding capabilities of the models after IR grounding, 
both in zero-shot and fine-tuning setups. For completion, we report performance on Multipl-E in terms of \texttt{pass@1}, \texttt{pass@10}, and \texttt{pass@25}. 
We test zero-shot code understanding on CodeXGLUE~\cite{lu2021codexglue} docstring generation task, which requires models to generate a docstring description given the function code as the prompt. We measure the performance with Smoothed \texttt{BLEU-4}~\cite{lin-och-2004-orange} scores w.r.t. the reference docstrings for the languages present in \texttt{SLTrans}: \texttt{Python}, \texttt{Ruby}, and \texttt{Go}.

Regarding fine-tuning, we benchmark on the Commit Chronicle~\cite{eliseeva2023commit} commit message generation task\footnote{The task also indirectly tests the commonsense knowledge in various languages as it requires the commit message to be generated purely from the \textit{diff} in the absence of the original code}. For the 8 languages present in \texttt{SLTrans}--- \texttt{C}, \texttt{C++}, \texttt{Go}, \texttt{Objective-C}, \texttt{Python}, \texttt{Ruby}, \texttt{Rust} and \texttt{Swift}---we fine-tune the IR-grounded Code-LLMs and report the performance in terms of \texttt{ROUGE-2} and \texttt{ROUGE-L} against the reference commit messages.

Results in \Cref{table:avg-mult} show that  \texttt{IRCoder} significantly and consistently outperforms the base LLMs on all multilingual benchmarks. The language-level breakdown of results (in \Cref{app:det}), suggests that grounding in IR facilitates cross-lingual transfer since we observe substantial improvements for low-resource language. 

The results demand one further point of discussion. Our findings are in contrast with the findings of \newcite{pmlr-v202-orlanski23a} who show a trade-off between the performance on high and low-resource languages: we, instead, observe gains across the board with no evidence of interference even between typologically disparate programming languages. 
We find that the IR-grounding also substantially boosts performance on high-resource languages like \texttt{C++} and \texttt{Python} for which the Code-LLMs have seen hundreds of billions of tokens in pre-training. This contributes to the hypothesis that, despite their large-scale pre-training, Code-LMs gain a limited understanding of higher-level concepts such as control and data flow~\cite{hooda2024large}, instead resorting to superficial attributes such as identifier names for anchoring representations across languages~\cite{ahmed2022multilingual}. IR, instead, quite intuitively, does have the potential to align code representations over such concepts. For example, the single-static assignment (SSA) form used by LLVM alongside transformations such as loop vectorization and register allocation specifies the data flow explicitly; other modifications captured by IR, such as loop simplification also aid in simplifying the control flow of source code thus aiding code understanding.

\vspace{.9mm}
\noindent\textbf{{RQ4}: Grounding in IR improves multilingual instruction following.}
Finally, we test if the improvements from IR grounding extend to instruction following. To this end, we perform 3 epochs of instruction tuning on 23.5k instruction-output pairs and evaluate instruction following on the HumanEvalFixDocs~\cite{muennighoff2023octopack} task. The task instructs the model to fix buggy code snippets given the docstring of the correct sub-routine and tests the models' ability to correct faults such as identifier and operator misuse as well as missing or excess logic. We evaluate for \texttt{SLTrans} languages: \texttt{C++}, \texttt{Go}, \texttt{Python}, and \texttt{Rust}. \Cref{table:avg-mult} shows again that IR grounding brings performance gains, with the largest improvements observed for the strongest Code-LMs.  
This is consistent with existing work which shows that the benefits of instruction tuning are most apparent for strong base models~\cite{muennighoff2023octopack, longpre2023flan}.
\vspace{-.2em}

\section{Conclusion}
\label{sec:conclusion}
\vspace{-.6em}
In this work, we investigate the effects of grounding heterogeneous source-code to a shared intermediate representation (IR) on code understanding and generation abilities of Code-LLMs. To this end, we first create \texttt{SLTrans}, a 26.2B token source code-IR parallel dataset containing nearly 4M training examples. We then perform continued pretraining on the corpus that includes parallel source code-IR data from \texttt{SLTrans} for 6 established Code-LMs, demonstrating the IR grounding brings substantial performance gains in prompt robustness, multilingual code completion, code understanding, and instruction following, all while training on data orders of magnitude smaller than Code-LLM pre-training corpus. We hope that our encouraging results catalyze broader research efforts on the inclusion of intermediate code representations both in Code-LLM pre-training as well as in the post-hoc adaptation of pre-trained models.



\section*{Limitations}
We show that compiler IR is a powerful source of cross-lingual alignment that allows for the structures in various languages to be anchored in common IR constructs. However, this is by no means a perfect process. Different frontends make disparate choices regarding how source code must be transformed to IR leading to several `dialects' of IR that are all valid but may slightly differ. While this does not seem to get in the way of our gains, it might have an effect when our approach is extended to newer languages with less mature toolchains.

Additionally, while the middle-end LLVM IR is intended to be a target-platform agnostic representation, this constraint can sometimes be violated due to the presence of platform-specific constants, application binary interface code, and linker logic. For our purposes, this was worked around by some data cleaning and by sourcing the IR consistently from the same platform.

Thirdly, there is a risk that the IR may not be able to anchor all the constructs of a language. While in some languages like \texttt{C} and \texttt{C++} there is a strong mapping between the language constructs and LLVM ones, in others the association might be less tight. For instance, in \texttt{Rust}, the source code is first transformed to the language's own IR\footnote{\url{https://rustc-dev-guide.rust-lang.org/mir/index.html}} before the LLVM framework is used. Our results indicate that this hasn't gotten in the way so far.

Finally, due to the IR being on average several times longer than the source code, there arise constraints on the types of models to which our approach can be applied. Most competitive Code-LMs have a context window of at least 4096 tokens making this largely a non-issue. However, it might pose problems in applying this method to older Code-LMs.

\section*{Ethical Risks}
Our work does not directly interface with any human annotators, with our data collection and evaluation being completely automated. However, there still exists the risk of our improved model being more competent at generating malicious code. This is a prospect we haven't explicitly evaluated for. We take mitigating steps by releasing the Docker containers used in our training and evaluation jobs, to minimize the risks to downstream users employing our models and methods. As a further guardrail, we plan to release our artefacts under a non-commercial license \faCreativeCommons\ \faCreativeCommonsBy\ \faCreativeCommonsNc\ \faCreativeCommonsSa.

\section*{Acknowledgements}

This work has been funded by Huawei Technologies (Ireland) Co., Ltd.

\vspace{-1em}
\bibliography{anthology, custom}

\begin{thebibliography}{88}
\expandafter\ifx\csname natexlab\endcsname\relax\def\natexlab#1{#1}\fi

\bibitem[{Ahmed and Devanbu(2022)}]{ahmed2022multilingual}
Toufique Ahmed and Premkumar Devanbu. 2022.
\newblock \href {https://dl.acm.org/doi/abs/10.1145/3510003.3510049} {Multilingual training for software engineering}.
\newblock In \emph{Proceedings of the 44th International Conference on Software Engineering}, pages 1443--1455.

\bibitem[{Allamanis(2019)}]{allamanis2019adverse}
Miltiadis Allamanis. 2019.
\newblock \href {https://dl.acm.org/doi/10.1145/3359591.3359735} {The adverse effects of code duplication in machine learning models of code}.
\newblock In \emph{Proceedings of the 2019 ACM SIGPLAN International Symposium on New Ideas, New Paradigms, and Reflections on Programming and Software}, pages 143--153.

\bibitem[{Ansell et~al.(2022)Ansell, Ponti, Korhonen, and Vuli{\'c}}]{ansell-etal-2022-composable}
Alan Ansell, Edoardo Ponti, Anna Korhonen, and Ivan Vuli{\'c}. 2022.
\newblock \href {https://doi.org/10.18653/v1/2022.acl-long.125} {Composable sparse fine-tuning for cross-lingual transfer}.
\newblock In \emph{Proceedings of the 60th Annual Meeting of the Association for Computational Linguistics (Volume 1: Long Papers)}, pages 1778--1796, Dublin, Ireland. Association for Computational Linguistics.

\bibitem[{Arivazhagan et~al.(2019)Arivazhagan, Bapna, Firat, Lepikhin, Johnson, Krikun, Chen, Cao, Foster, Cherry et~al.}]{arivazhagan2019massively}
Naveen Arivazhagan, Ankur Bapna, Orhan Firat, Dmitry Lepikhin, Melvin Johnson, Maxim Krikun, Mia~Xu Chen, Yuan Cao, George Foster, Colin Cherry, et~al. 2019.
\newblock \href {https://arxiv.org/abs/1907.05019} {Massively multilingual neural machine translation in the wild: Findings and challenges}.
\newblock \emph{arXiv preprint 1907.05019}.

\bibitem[{Artetxe and Schwenk(2019)}]{artetxe-schwenk-2019-massively}
Mikel Artetxe and Holger Schwenk. 2019.
\newblock \href {https://doi.org/10.1162/tacl_a_00288} {Massively multilingual sentence embeddings for zero-shot cross-lingual transfer and beyond}.
\newblock \emph{Transactions of the Association for Computational Linguistics}, 7:597--610.

\bibitem[{Athiwaratkun et~al.(2023)Athiwaratkun, Gouda, Wang, Li, Tian, Tan, Ahmad, Wang, Sun, Shang, Gonugondla, Ding, Kumar, Fulton, Farahani, Jain, Giaquinto, Qian, Ramanathan, and Nallapati}]{athiwaratkun2023multilingual}
Ben Athiwaratkun, Sanjay~Krishna Gouda, Zijian Wang, Xiaopeng Li, Yuchen Tian, Ming Tan, Wasi~Uddin Ahmad, Shiqi Wang, Qing Sun, Mingyue Shang, Sujan~Kumar Gonugondla, Hantian Ding, Varun Kumar, Nathan Fulton, Arash Farahani, Siddhartha Jain, Robert Giaquinto, Haifeng Qian, Murali~Krishna Ramanathan, and Ramesh Nallapati. 2023.
\newblock \href {https://openreview.net/pdf?id=Bo7eeXm6An8} {Multi-lingual evaluation of code generation models}.
\newblock In \emph{The Eleventh International Conference on Learning Representations, {ICLR} 2023, Kigali, Rwanda, May 1-5, 2023}. OpenReview.net.

\bibitem[{Austin et~al.(2021)Austin, Odena, Nye, Bosma, Michalewski, Dohan, Jiang, Cai, Terry, Le et~al.}]{austin2021program}
Jacob Austin, Augustus Odena, Maxwell Nye, Maarten Bosma, Henryk Michalewski, David Dohan, Ellen Jiang, Carrie Cai, Michael Terry, Quoc Le, et~al. 2021.
\newblock \href {https://arxiv.org/abs/2108.07732} {Program synthesis with large language models}.
\newblock \emph{arXiv preprint 2108.07732}.

\bibitem[{Bahrami et~al.(2021)Bahrami, Shrikanth, Mizobuchi, Liu, Fukuyori, Chen, and Munakata}]{bahrami2021augmentedcode}
Mehdi Bahrami, NC~Shrikanth, Yuji Mizobuchi, Lei Liu, Masahiro Fukuyori, Wei-Peng Chen, and Kazuki Munakata. 2021.
\newblock \href {https://arxiv.org/abs/2110.08512} {Augmentedcode: Examining the effects of natural language resources in code retrieval models}.
\newblock \emph{arXiv preprint 2110.08512}.

\bibitem[{Blevins et~al.(2024)Blevins, Limisiewicz, Gururangan, Li, Gonen, Smith, and Zettlemoyer}]{blevins2024breaking}
Terra Blevins, Tomasz Limisiewicz, Suchin Gururangan, Margaret Li, Hila Gonen, Noah~A Smith, and Luke Zettlemoyer. 2024.
\newblock \href {https://arxiv.org/abs/2401.10440} {Breaking the curse of multilinguality with cross-lingual expert language models}.
\newblock \emph{arXiv preprint 2401.10440}.

\bibitem[{Brauckmann et~al.(2020)Brauckmann, Goens, Ertel, and Castrillon}]{brauckmann2020compiler}
Alexander Brauckmann, Andr{\'e}s Goens, Sebastian Ertel, and Jeronimo Castrillon. 2020.
\newblock \href {https://dl.acm.org/doi/abs/10.1145/3377555.3377894} {Compiler-based graph representations for deep learning models of code}.
\newblock In \emph{Proceedings of the 29th International Conference on Compiler Construction}, pages 201--211.

\bibitem[{Brockschmidt et~al.(2019)Brockschmidt, Allamanis, Gaunt, and Polozov}]{brockschmidt2018generative}
Marc Brockschmidt, Miltiadis Allamanis, Alexander~L. Gaunt, and Oleksandr Polozov. 2019.
\newblock \href {https://openreview.net/forum?id=Bke4KsA5FX} {Generative code modeling with graphs}.
\newblock In \emph{7th International Conference on Learning Representations, {ICLR} 2019, New Orleans, LA, USA, May 6-9, 2019}. OpenReview.net.

\bibitem[{Broder(1997)}]{broder1997resemblance}
Andrei~Z Broder. 1997.
\newblock \href {https://www.computer.org/csdl/proceedings-article/sequences/1997/81320021/12OmNwDACjh} {On the resemblance and containment of documents}.
\newblock In \emph{Proceedings. Compression and Complexity of SEQUENCES 1997 (Cat. No. 97TB100171)}, pages 21--29. IEEE.

\bibitem[{Caballero and Sutskever(2021)}]{Caballero_Description2Code_Dataset_2016}
Ethan Caballero and Ilya Sutskever. 2021.
\newblock \href {https://doi.org/10.5281/zenodo.5665051} {Description2code dataset}.

\bibitem[{Cassano et~al.(2023{\natexlab{a}})Cassano, Gouwar, Lucchetti, Schlesinger, Anderson, Greenberg, Jangda, and Guha}]{cassano2023knowledge}
Federico Cassano, John Gouwar, Francesca Lucchetti, Claire Schlesinger, Carolyn~Jane Anderson, Michael Greenberg, Abhinav Jangda, and Arjun Guha. 2023{\natexlab{a}}.
\newblock \href {http://arxiv.org/abs/2308.09895} {Knowledge transfer from high-resource to low-resource programming languages for code llms}.

\bibitem[{Cassano et~al.(2022)Cassano, Gouwar, Nguyen, Nguyen, Phipps-Costin, Pinckney, Yee, Zi, Anderson, Feldman et~al.}]{cassano2022multipl}
Federico Cassano, John Gouwar, Daniel Nguyen, Sydney Nguyen, Luna Phipps-Costin, Donald Pinckney, Ming-Ho Yee, Yangtian Zi, Carolyn~Jane Anderson, Molly~Q Feldman, et~al. 2022.
\newblock \href {https://arxiv.org/abs/2208.08227} {Multipl-e: A scalable and extensible approach to benchmarking neural code generation}.
\newblock \emph{arXiv preprint 2208.08227}.

\bibitem[{Cassano et~al.(2023{\natexlab{b}})Cassano, Li, Sethi, Shinn, Brennan-Jones, Lozhkov, Anderson, and Guha}]{cassano2023edit}
Federico Cassano, Luisa Li, Akul Sethi, Noah Shinn, Abby Brennan-Jones, Anton Lozhkov, Carolyn Anderson, and Arjun Guha. 2023{\natexlab{b}}.
\newblock \href {https://arxiv.org/abs/2312.12450} {Can it edit? evaluating the ability of large language models to follow code editing instructions}.
\newblock \emph{arXiv preprint 2312.12450}.

\bibitem[{Chai et~al.(2023)Chai, Wang, Pang, Sun, Tian, and Wu}]{chai-etal-2023-ernie}
Yekun Chai, Shuohuan Wang, Chao Pang, Yu~Sun, Hao Tian, and Hua Wu. 2023.
\newblock \href {https://doi.org/10.18653/v1/2023.findings-acl.676} {{ERNIE}-code: Beyond {E}nglish-centric cross-lingual pretraining for programming languages}.
\newblock In \emph{Findings of the Association for Computational Linguistics: ACL 2023}, pages 10628--10650, Toronto, Canada. Association for Computational Linguistics.

\bibitem[{Chakraborty et~al.(2022)Chakraborty, Ahmed, Ding, Devanbu, and Ray}]{chakraborty2022natgen}
Saikat Chakraborty, Toufique Ahmed, Yangruibo Ding, Premkumar~T Devanbu, and Baishakhi Ray. 2022.
\newblock \href {https://dl.acm.org/doi/abs/10.1145/3540250.3549162} {Natgen: generative pre-training by “naturalizing” source code}.
\newblock In \emph{Proceedings of the 30th ACM Joint European Software Engineering Conference and Symposium on the Foundations of Software Engineering}, pages 18--30.

\bibitem[{Chen et~al.(2022)Chen, Fard, Lo, and Bryksin}]{chen2022transferability}
Fuxiang Chen, Fatemeh~H. Fard, David Lo, and Timofey Bryksin. 2022.
\newblock \href {https://doi.org/10.1145/3524610.3527917} {On the transferability of pre-trained language models for low-resource programming languages}.
\newblock In \emph{Proceedings of the 30th IEEE/ACM International Conference on Program Comprehension}, ICPC '22, page 401–412, New York, NY, USA. Association for Computing Machinery.

\bibitem[{Chen et~al.(2021)Chen, Tworek, Jun, Yuan, Pinto, Kaplan, Edwards, Burda, Joseph, Brockman et~al.}]{chen2021evaluating}
Mark Chen, Jerry Tworek, Heewoo Jun, Qiming Yuan, Henrique Ponde de~Oliveira Pinto, Jared Kaplan, Harri Edwards, Yuri Burda, Nicholas Joseph, Greg Brockman, et~al. 2021.
\newblock \href {https://arxiv.org/abs/2107.03374} {Evaluating large language models trained on code}.
\newblock \emph{arXiv preprint 2107.03374}.

\bibitem[{Chen et~al.(2023{\natexlab{a}})Chen, Sun, Wang, Li, and Gao}]{chen-etal-2023-pass}
Nuo Chen, Qiushi Sun, Jianing Wang, Xiang Li, and Ming Gao. 2023{\natexlab{a}}.
\newblock \href {https://doi.org/10.18653/v1/2023.findings-emnlp.42} {Pass-tuning: Towards structure-aware parameter-efficient tuning for code representation learning}.
\newblock In \emph{Findings of the Association for Computational Linguistics: EMNLP 2023}, pages 577--591, Singapore. Association for Computational Linguistics.

\bibitem[{Chen et~al.(2023{\natexlab{b}})Chen, Lin, Sch{\"a}rli, and Zhou}]{chen2023teaching}
Xinyun Chen, Maxwell Lin, Nathanael Sch{\"a}rli, and Denny Zhou. 2023{\natexlab{b}}.
\newblock \href {https://arxiv.org/abs/2304.05128} {Teaching large language models to self-debug}.
\newblock \emph{arXiv preprint 2304.05128}.

\bibitem[{Chung et~al.(2023)Chung, Garcia, Roberts, Tay, Firat, Narang, and Constant}]{chung2023unimax}
Hyung~Won Chung, Xavier Garcia, Adam Roberts, Yi~Tay, Orhan Firat, Sharan Narang, and Noah Constant. 2023.
\newblock \href {https://openreview.net/pdf?id=kXwdL1cWOAi} {Unimax: Fairer and more effective language sampling for large-scale multilingual pretraining}.
\newblock In \emph{The Eleventh International Conference on Learning Representations, {ICLR} 2023, Kigali, Rwanda, May 1-5, 2023}. OpenReview.net.

\bibitem[{Conneau et~al.(2020)Conneau, Khandelwal, Goyal, Chaudhary, Wenzek, Guzm{\'a}n, Grave, Ott, Zettlemoyer, and Stoyanov}]{conneau-etal-2020-unsupervised}
Alexis Conneau, Kartikay Khandelwal, Naman Goyal, Vishrav Chaudhary, Guillaume Wenzek, Francisco Guzm{\'a}n, Edouard Grave, Myle Ott, Luke Zettlemoyer, and Veselin Stoyanov. 2020.
\newblock \href {https://doi.org/10.18653/v1/2020.acl-main.747} {Unsupervised cross-lingual representation learning at scale}.
\newblock In \emph{Proceedings of the 58th Annual Meeting of the Association for Computational Linguistics}, pages 8440--8451, Online. Association for Computational Linguistics.

\bibitem[{Dhole et~al.(2021)Dhole, Gangal, Gehrmann, Gupta, Li, Mahamood, Mahendiran, Mille, Shrivastava, Tan et~al.}]{dhole2022nlaugmenter}
Kaustubh~D Dhole, Varun Gangal, Sebastian Gehrmann, Aadesh Gupta, Zhenhao Li, Saad Mahamood, Abinaya Mahendiran, Simon Mille, Ashish Shrivastava, Samson Tan, et~al. 2021.
\newblock \href {https://arxiv.org/abs/2112.02721} {Nl-augmenter: A framework for task-sensitive natural language augmentation}.
\newblock \emph{arXiv preprint 2112.02721}.

\bibitem[{Dinh et~al.(2023)Dinh, Zhao, Tan, Negrinho, Lausen, Zha, and Karypis}]{dinh2023large}
Tuan Dinh, Jinman Zhao, Samson Tan, Renato Negrinho, Leonard Lausen, Sheng Zha, and George Karypis. 2023.
\newblock \href {http://papers.nips.cc/paper\_files/paper/2023/hash/819cebb05f993840e8a52d7564c5c282-Abstract-Conference.html} {Large language models of code fail at completing code with potential bugs}.
\newblock In \emph{Advances in Neural Information Processing Systems 36: Annual Conference on Neural Information Processing Systems 2023, NeurIPS 2023, New Orleans, LA, USA, December 10 - 16, 2023}.

\bibitem[{Dunay et~al.(2024)Dunay, Cheng, Tait, Thakkar, Rigby, Chiu, Ahmad, Ganesan, Maddila, Murali et~al.}]{dunay2024multiline}
Omer Dunay, Daniel Cheng, Adam Tait, Parth Thakkar, Peter~C Rigby, Andy Chiu, Imad Ahmad, Arun Ganesan, Chandra Maddila, Vijayaraghavan Murali, et~al. 2024.
\newblock \href {https://arxiv.org/abs/2402.04141} {Multi-line ai-assisted code authoring}.
\newblock \emph{arXiv preprint 2402.04141}.

\bibitem[{Eliseeva et~al.(2023)Eliseeva, Sokolov, Bogomolov, Golubev, Dig, and Bryksin}]{eliseeva2023commit}
Aleksandra Eliseeva, Yaroslav Sokolov, Egor Bogomolov, Yaroslav Golubev, Danny Dig, and Timofey Bryksin. 2023.
\newblock \href {https://doi.org/10.1109/ASE56229.2023.00078} {From commit message generation to history-aware commit message completion}.
\newblock In \emph{38th {IEEE/ACM} International Conference on Automated Software Engineering, {ASE} 2023, Luxembourg, September 11-15, 2023}, pages 723--735. {IEEE}.

\bibitem[{Frömmgen et~al.(2024)Frömmgen, Austin, Choy, Ghelani, Kharatyan, Surita, Khrapko, Lamblin, Manzagol, Revaj, Tabachnyk, Tarlow, Villela, Zheng, Chandra, and Maniatis}]{didact2024google}
Alexander Frömmgen, Jacob Austin, Peter Choy, Nimesh Ghelani, Lera Kharatyan, Gabriela Surita, Elena Khrapko, Pascal Lamblin, Pierre-Antoine Manzagol, Marcus Revaj, Maxim Tabachnyk, Daniel Tarlow, Kevin Villela, Dan Zheng, Satish Chandra, and Petros Maniatis. 2024.
\newblock \href {https://blog.research.google/2023/05/resolving-code-review-comments-with-ml.html} {Resolving code review comments with machine learning}.
\newblock In \emph{2024 IEEE/ACM 46th International Conference on Software Engineering: Software Engineering in Practice (ICSE-SEIP)}.

\bibitem[{Gao et~al.(2023)Gao, Zhang, He, Wu, and Wang}]{gao-etal-2023-learning-multilingual}
Pengzhi Gao, Liwen Zhang, Zhongjun He, Hua Wu, and Haifeng Wang. 2023.
\newblock \href {https://doi.org/10.18653/v1/2023.emnlp-industry.25} {Learning multilingual sentence representations with cross-lingual consistency regularization}.
\newblock In \emph{Proceedings of the 2023 Conference on Empirical Methods in Natural Language Processing: Industry Track}, pages 243--262, Singapore. Association for Computational Linguistics.

\bibitem[{Gong et~al.(2024)Gong, Elhoushi, and Cheung}]{gong2024ast}
Linyuan Gong, Mostafa Elhoushi, and Alvin Cheung. 2024.
\newblock \href {https://arxiv.org/abs/2401.03003} {Ast-t5: Structure-aware pretraining for code generation and understanding}.
\newblock \emph{arXiv preprint 2401.03003}.

\bibitem[{Gou et~al.(2023)Gou, Shao, Gong, Shen, Yang, Duan, and Chen}]{gou2023critic}
Zhibin Gou, Zhihong Shao, Yeyun Gong, Yelong Shen, Yujiu Yang, Nan Duan, and Weizhu Chen. 2023.
\newblock \href {https://arxiv.org/abs/2305.11738} {Critic: Large language models can self-correct with tool-interactive critiquing}.
\newblock \emph{arXiv preprint 2305.11738}.

\bibitem[{Grossman et~al.(2023)Grossman, Paehler, Parasyris, Ben-Nun, Hegna, Moses, Diaz, Trofin, and Doerfert}]{grossman2023compile}
Aiden Grossman, Ludger Paehler, Konstantinos Parasyris, Tal Ben-Nun, Jacob Hegna, William Moses, Jose M~Monsalve Diaz, Mircea Trofin, and Johannes Doerfert. 2023.
\newblock \href {https://arxiv.org/abs/2309.15432} {Compile: A large ir dataset from production sources}.
\newblock \emph{arXiv preprint 2309.15432}.

\bibitem[{Gunasekar et~al.(2023)Gunasekar, Zhang, Aneja, Mendes, Del~Giorno, Gopi, Javaheripi, Kauffmann, de~Rosa, Saarikivi et~al.}]{gunasekar2023textbooks}
Suriya Gunasekar, Yi~Zhang, Jyoti Aneja, Caio C{\'e}sar~Teodoro Mendes, Allie Del~Giorno, Sivakanth Gopi, Mojan Javaheripi, Piero Kauffmann, Gustavo de~Rosa, Olli Saarikivi, et~al. 2023.
\newblock \href {https://arxiv.org/abs/2306.11644} {Textbooks are all you need}.
\newblock \emph{arXiv preprint 2306.11644}.

\bibitem[{Guo et~al.(2022)Guo, Lu, Duan, Wang, Zhou, and Yin}]{guo-etal-2022-unixcoder}
Daya Guo, Shuai Lu, Nan Duan, Yanlin Wang, Ming Zhou, and Jian Yin. 2022.
\newblock \href {https://doi.org/10.18653/v1/2022.acl-long.499} {{U}ni{X}coder: Unified cross-modal pre-training for code representation}.
\newblock In \emph{Proceedings of the 60th Annual Meeting of the Association for Computational Linguistics (Volume 1: Long Papers)}, pages 7212--7225, Dublin, Ireland. Association for Computational Linguistics.

\bibitem[{Guo et~al.(2024)Guo, Zhu, Yang, Xie, Dong, Zhang, Chen, Bi, Wu, Li et~al.}]{guo2024deepseek}
Daya Guo, Qihao Zhu, Dejian Yang, Zhenda Xie, Kai Dong, Wentao Zhang, Guanting Chen, Xiao Bi, Y~Wu, YK~Li, et~al. 2024.
\newblock \href {https://arxiv.org/abs/2401.14196} {Deepseek-coder: When the large language model meets programming--the rise of code intelligence}.
\newblock \emph{arXiv preprint 2401.14196}.

\bibitem[{Hendrycks et~al.(2021)Hendrycks, Basart, Kadavath, Mazeika, Arora, Guo, Burns, Puranik, He, Song, and Steinhardt}]{hendrycks2021measuring}
Dan Hendrycks, Steven Basart, Saurav Kadavath, Mantas Mazeika, Akul Arora, Ethan Guo, Collin Burns, Samir Puranik, Horace He, Dawn Song, and Jacob Steinhardt. 2021.
\newblock \href {https://datasets-benchmarks-proceedings.neurips.cc/paper/2021/hash/c24cd76e1ce41366a4bbe8a49b02a028-Abstract-round2.html} {Measuring coding challenge competence with {APPS}}.
\newblock In \emph{Proceedings of the Neural Information Processing Systems Track on Datasets and Benchmarks 1, NeurIPS Datasets and Benchmarks 2021, December 2021, virtual}.

\bibitem[{Hooda et~al.(2024)Hooda, Christodorescu, Allamanis, Wilson, Fawaz, and Jha}]{hooda2024large}
Ashish Hooda, Mihai Christodorescu, Miltos Allamanis, Aaron Wilson, Kassem Fawaz, and Somesh Jha. 2024.
\newblock \href {https://arxiv.org/abs/2402.05980} {Do large code models understand programming concepts? a black-box approach}.
\newblock \emph{arXiv preprint 2402.05980}.

\bibitem[{Hu et~al.(2022)Hu, Shen, Wallis, Allen{-}Zhu, Li, Wang, Wang, and Chen}]{hu2022lora}
Edward~J. Hu, Yelong Shen, Phillip Wallis, Zeyuan Allen{-}Zhu, Yuanzhi Li, Shean Wang, Lu~Wang, and Weizhu Chen. 2022.
\newblock \href {https://openreview.net/forum?id=nZeVKeeFYf9} {Lora: Low-rank adaptation of large language models}.
\newblock In \emph{The Tenth International Conference on Learning Representations, {ICLR} 2022, Virtual Event, April 25-29, 2022}. OpenReview.net.

\bibitem[{Jain et~al.(2021)Jain, Jain, Zhang, Abbeel, Gonzalez, and Stoica}]{jain-etal-2021-contrastive}
Paras Jain, Ajay Jain, Tianjun Zhang, Pieter Abbeel, Joseph Gonzalez, and Ion Stoica. 2021.
\newblock \href {https://doi.org/10.18653/v1/2021.emnlp-main.482} {Contrastive code representation learning}.
\newblock In \emph{Proceedings of the 2021 Conference on Empirical Methods in Natural Language Processing}, pages 5954--5971, Online and Punta Cana, Dominican Republic. Association for Computational Linguistics.

\bibitem[{Jiang et~al.(2022)Jiang, Song, Ge, Meng, Yao, and Su}]{jiang2022ast}
Hui Jiang, Linfeng Song, Yubin Ge, Fandong Meng, Junfeng Yao, and Jinsong Su. 2022.
\newblock \href {https://doi.org/10.1109/TASLP.2021.3138717} {An {AST} structure enhanced decoder for code generation}.
\newblock \emph{{IEEE} {ACM} Trans. Audio Speech Lang. Process.}, 30:468--476.

\bibitem[{Jiang et~al.(2023)Jiang, Wang, and Wang}]{jiang2023selfevolve}
Shuyang Jiang, Yuhao Wang, and Yu~Wang. 2023.
\newblock \href {https://arxiv.org/abs/2306.02907} {Selfevolve: A code evolution framework via large language models}.
\newblock \emph{arXiv preprint 2306.02907}.

\bibitem[{Kingma and Ba(2015)}]{KingBa15}
Diederik~P. Kingma and Jimmy Ba. 2015.
\newblock \href {http://arxiv.org/abs/1412.6980} {Adam: {A} method for stochastic optimization}.
\newblock In \emph{3rd International Conference on Learning Representations, {ICLR} 2015, San Diego, CA, USA, May 7-9, 2015, Conference Track Proceedings}.

\bibitem[{Kocetkov et~al.(2023)Kocetkov, Li, allal, LI, Mou, Jernite, Mitchell, Ferrandis, Hughes, Wolf, Bahdanau, Werra, and de~Vries}]{kocetkov2023the}
Denis Kocetkov, Raymond Li, Loubna~Ben allal, Jia LI, Chenghao Mou, Yacine Jernite, Margaret Mitchell, Carlos~Mu{\~n}oz Ferrandis, Sean Hughes, Thomas Wolf, Dzmitry Bahdanau, Leandro~Von Werra, and Harm de~Vries. 2023.
\newblock \href {https://openreview.net/forum?id=pxpbTdUEpD} {The stack: 3 {TB} of permissively licensed source code}.
\newblock \emph{Transactions on Machine Learning Research}.

\bibitem[{K{\"{o}}pf et~al.(2023)K{\"{o}}pf, Kilcher, von R{\"{u}}tte, Anagnostidis, Tam, Stevens, Barhoum, Nguyen, Stanley, Nagyfi, ES, Suri, Glushkov, Dantuluri, Maguire, Schuhmann, Nguyen, and Mattick}]{köpf2023openassistant}
Andreas K{\"{o}}pf, Yannic Kilcher, Dimitri von R{\"{u}}tte, Sotiris Anagnostidis, Zhi~Rui Tam, Keith Stevens, Abdullah Barhoum, Duc Nguyen, Oliver Stanley, Rich{\'{a}}rd Nagyfi, Shahul ES, Sameer Suri, David Glushkov, Arnav Dantuluri, Andrew Maguire, Christoph Schuhmann, Huu Nguyen, and Alexander Mattick. 2023.
\newblock \href {http://papers.nips.cc/paper\_files/paper/2023/hash/949f0f8f32267d297c2d4e3ee10a2e7e-Abstract-Datasets\_and\_Benchmarks.html} {Openassistant conversations - democratizing large language model alignment}.
\newblock In \emph{Advances in Neural Information Processing Systems 36: Annual Conference on Neural Information Processing Systems 2023, NeurIPS 2023, New Orleans, LA, USA, December 10 - 16, 2023}.

\bibitem[{Kwon et~al.(2023)Kwon, Li, Zhuang, Sheng, Zheng, Yu, Gonzalez, Zhang, and Stoica}]{kwon2023efficient}
Woosuk Kwon, Zhuohan Li, Siyuan Zhuang, Ying Sheng, Lianmin Zheng, Cody~Hao Yu, Joseph Gonzalez, Hao Zhang, and Ion Stoica. 2023.
\newblock \href {https://doi.org/10.1145/3600006.3613165} {Efficient memory management for large language model serving with pagedattention}.
\newblock In \emph{Proceedings of the 29th Symposium on Operating Systems Principles, {SOSP} 2023, Koblenz, Germany, October 23-26, 2023}, pages 611--626. {ACM}.

\bibitem[{Lattner and Adve(2004)}]{lattner2004llvm}
Chris Lattner and Vikram~S. Adve. 2004.
\newblock \href {https://doi.org/10.1109/CGO.2004.1281665} {{LLVM:} {A} compilation framework for lifelong program analysis {\&} transformation}.
\newblock In \emph{2nd {IEEE} / {ACM} International Symposium on Code Generation and Optimization {(CGO} 2004), 20-24 March 2004, San Jose, CA, {USA}}, pages 75--88. {IEEE} Computer Society.

\bibitem[{Lauscher et~al.(2020)Lauscher, Ravishankar, Vuli{\'c}, and Glava{\v{s}}}]{lauscher-etal-2020-zero}
Anne Lauscher, Vinit Ravishankar, Ivan Vuli{\'c}, and Goran Glava{\v{s}}. 2020.
\newblock \href {https://doi.org/10.18653/v1/2020.emnlp-main.363} {From zero to hero: {O}n the limitations of zero-shot language transfer with multilingual {T}ransformers}.
\newblock In \emph{Proceedings of the 2020 Conference on Empirical Methods in Natural Language Processing (EMNLP)}, pages 4483--4499, Online. Association for Computational Linguistics.

\bibitem[{Le et~al.(2022)Le, Wang, Gotmare, Savarese, and Hoi}]{le2022coderl}
Hung Le, Yue Wang, Akhilesh~Deepak Gotmare, Silvio Savarese, and Steven~Chu{-}Hong Hoi. 2022.
\newblock \href {http://papers.nips.cc/paper\_files/paper/2022/hash/8636419dea1aa9fbd25fc4248e702da4-Abstract-Conference.html} {Coderl: Mastering code generation through pretrained models and deep reinforcement learning}.
\newblock In \emph{Advances in Neural Information Processing Systems 35: Annual Conference on Neural Information Processing Systems 2022, NeurIPS 2022, New Orleans, LA, USA, November 28 - December 9, 2022}.

\bibitem[{Lee and Hwang(2023)}]{lee-hwang-2023-multilingual}
Jaeseong Lee and Seung-won Hwang. 2023.
\newblock \href {https://doi.org/10.18653/v1/2023.findings-emnlp.629} {Multilingual lottery tickets to pretrain language models}.
\newblock In \emph{Findings of the Association for Computational Linguistics: EMNLP 2023}, pages 9387--9398, Singapore. Association for Computational Linguistics.

\bibitem[{Lee et~al.(2022)Lee, Ippolito, Nystrom, Zhang, Eck, Callison-Burch, and Carlini}]{lee-etal-2022-deduplicating}
Katherine Lee, Daphne Ippolito, Andrew Nystrom, Chiyuan Zhang, Douglas Eck, Chris Callison-Burch, and Nicholas Carlini. 2022.
\newblock \href {https://doi.org/10.18653/v1/2022.acl-long.577} {Deduplicating training data makes language models better}.
\newblock In \emph{Proceedings of the 60th Annual Meeting of the Association for Computational Linguistics (Volume 1: Long Papers)}, pages 8424--8445, Dublin, Ireland. Association for Computational Linguistics.

\bibitem[{Li et~al.(2023)Li, allal, Zi, Muennighoff, Kocetkov, Mou, Marone, Akiki, LI, Chim, Liu, Zheltonozhskii, Zhuo, Wang, Dehaene, Lamy-Poirier, Monteiro, Gontier, Yee, Umapathi, Zhu, Lipkin, Oblokulov, Wang, Murthy, Stillerman, Patel, Abulkhanov, Zocca, Dey, Zhang, Bhattacharyya, Yu, Luccioni, Villegas, Zhdanov, Lee, Timor, Ding, Schlesinger, Schoelkopf, Ebert, Dao, Mishra, Gu, Anderson, Dolan-Gavitt, Contractor, Reddy, Fried, Bahdanau, Jernite, Ferrandis, Hughes, Wolf, Guha, Werra, and de~Vries}]{li2023starcoder}
Raymond Li, Loubna~Ben allal, Yangtian Zi, Niklas Muennighoff, Denis Kocetkov, Chenghao Mou, Marc Marone, Christopher Akiki, Jia LI, Jenny Chim, Qian Liu, Evgenii Zheltonozhskii, Terry~Yue Zhuo, Thomas Wang, Olivier Dehaene, Joel Lamy-Poirier, Joao Monteiro, Nicolas Gontier, Ming-Ho Yee, Logesh~Kumar Umapathi, Jian Zhu, Ben Lipkin, Muhtasham Oblokulov, Zhiruo Wang, Rudra Murthy, Jason~T Stillerman, Siva~Sankalp Patel, Dmitry Abulkhanov, Marco Zocca, Manan Dey, Zhihan Zhang, Urvashi Bhattacharyya, Wenhao Yu, Sasha Luccioni, Paulo Villegas, Fedor Zhdanov, Tony Lee, Nadav Timor, Jennifer Ding, Claire~S Schlesinger, Hailey Schoelkopf, Jan Ebert, Tri Dao, Mayank Mishra, Alex Gu, Carolyn~Jane Anderson, Brendan Dolan-Gavitt, Danish Contractor, Siva Reddy, Daniel Fried, Dzmitry Bahdanau, Yacine Jernite, Carlos~Mu{\~n}oz Ferrandis, Sean Hughes, Thomas Wolf, Arjun Guha, Leandro~Von Werra, and Harm de~Vries. 2023.
\newblock \href {https://openreview.net/forum?id=KoFOg41haE} {Starcoder: may the source be with you!}
\newblock \emph{Transactions on Machine Learning Research}.
\newblock Reproducibility Certification.

\bibitem[{Li et~al.(2022)Li, Ma, Wang, Wang, Tang, Nie, and Wu}]{li2022unleashing}
Zongjie Li, Pingchuan Ma, Huaijin Wang, Shuai Wang, Qiyi Tang, Sen Nie, and Shi Wu. 2022.
\newblock \href {https://doi.org/10.1145/3510003.3510217} {Unleashing the power of compiler intermediate representation to enhance neural program embeddings}.
\newblock In \emph{44th {IEEE/ACM} 44th International Conference on Software Engineering, {ICSE} 2022, Pittsburgh, PA, USA, May 25-27, 2022}, pages 2253--2265. {ACM}.

\bibitem[{Lian et~al.(2023)Lian, Goodson, Pentland, Cook, Vong, and "Teknium"}]{OpenOrca}
Wing Lian, Bleys Goodson, Eugene Pentland, Austin Cook, Chanvichet Vong, and "Teknium". 2023.
\newblock Openorca: An open dataset of gpt augmented flan reasoning traces.
\newblock \url{https://https://huggingface.co/Open-Orca/OpenOrca}.

\bibitem[{Lin and Och(2004)}]{lin-och-2004-orange}
Chin-Yew Lin and Franz~Josef Och. 2004.
\newblock \href {https://aclanthology.org/C04-1072} {{ORANGE}: a method for evaluating automatic evaluation metrics for machine translation}.
\newblock In \emph{{COLING} 2004: Proceedings of the 20th International Conference on Computational Linguistics}, pages 501--507, Geneva, Switzerland. COLING.

\bibitem[{Liu et~al.(2023)Liu, Zhu, Xiao, Fu, Han, Yang, and Ye}]{liu2023rltf}
Jiate Liu, Yiqin Zhu, Kaiwen Xiao, Qiang Fu, Xiao Han, Wei Yang, and Deheng Ye. 2023.
\newblock \href {https://arxiv.org/abs/2307.04349} {Rltf: Reinforcement learning from unit test feedback}.
\newblock \emph{arXiv preprint 2307.04349}.

\bibitem[{Longpre et~al.(2023)Longpre, Hou, Vu, Webson, Chung, Tay, Zhou, Le, Zoph, Wei, and Roberts}]{longpre2023flan}
Shayne Longpre, Le~Hou, Tu~Vu, Albert Webson, Hyung~Won Chung, Yi~Tay, Denny Zhou, Quoc~V. Le, Barret Zoph, Jason Wei, and Adam Roberts. 2023.
\newblock \href {https://proceedings.mlr.press/v202/longpre23a.html} {The flan collection: Designing data and methods for effective instruction tuning}.
\newblock In \emph{International Conference on Machine Learning, {ICML} 2023, 23-29 July 2023, Honolulu, Hawaii, {USA}}, volume 202 of \emph{Proceedings of Machine Learning Research}, pages 22631--22648. {PMLR}.

\bibitem[{Lu et~al.(2021)Lu, Guo, Ren, Huang, Svyatkovskiy, Blanco, Clement, Drain, Jiang, Tang, Li, Zhou, Shou, Zhou, Tufano, Gong, Zhou, Duan, Sundaresan, Deng, Fu, and Liu}]{lu2021codexglue}
Shuai Lu, Daya Guo, Shuo Ren, Junjie Huang, Alexey Svyatkovskiy, Ambrosio Blanco, Colin~B. Clement, Dawn Drain, Daxin Jiang, Duyu Tang, Ge~Li, Lidong Zhou, Linjun Shou, Long Zhou, Michele Tufano, Ming Gong, Ming Zhou, Nan Duan, Neel Sundaresan, Shao~Kun Deng, Shengyu Fu, and Shujie Liu. 2021.
\newblock \href {https://datasets-benchmarks-proceedings.neurips.cc/paper/2021/hash/c16a5320fa475530d9583c34fd356ef5-Abstract-round1.html} {Codexglue: {A} machine learning benchmark dataset for code understanding and generation}.
\newblock In \emph{Proceedings of the Neural Information Processing Systems Track on Datasets and Benchmarks 1, NeurIPS Datasets and Benchmarks 2021, December 2021, virtual}.

\bibitem[{Ma et~al.(2023)Ma, Yu, Li, Jia, Ma, Xu, Dong, and Liao}]{ma2023mulcs}
Yingwei Ma, Yue Yu, Shanshan Li, Zhouyang Jia, Jun Ma, Rulin Xu, Wei Dong, and Xiangke Liao. 2023.
\newblock \href {https://doi.org/10.1109/SANER56733.2023.00021} {Mulcs: Towards a unified deep representation for multilingual code search}.
\newblock In \emph{{IEEE} International Conference on Software Analysis, Evolution and Reengineering, {SANER} 2023, Taipa, Macao, March 21-24, 2023}, pages 120--131. {IEEE}.

\bibitem[{Mirzayanov(2020)}]{codeforces2020}
Mike Mirzayanov. 2020.
\newblock Codeforces: Results of 2020 [annual report].
\newblock \url{https://codeforces.com/blog/entry/89502}.

\bibitem[{Mou et~al.(2016)Mou, Li, Zhang, Wang, and Jin}]{mou2016convolutional}
Lili Mou, Ge~Li, Lu~Zhang, Tao Wang, and Zhi Jin. 2016.
\newblock \href {https://doi.org/10.1609/AAAI.V30I1.10139} {Convolutional neural networks over tree structures for programming language processing}.
\newblock In \emph{Proceedings of the Thirtieth {AAAI} Conference on Artificial Intelligence, February 12-17, 2016, Phoenix, Arizona, {USA}}, pages 1287--1293. {AAAI} Press.

\bibitem[{Muennighoff et~al.(2023)Muennighoff, Liu, Zebaze, Zheng, Hui, Zhuo, Singh, Tang, Werra, and Longpre}]{muennighoff2023octopack}
Niklas Muennighoff, Qian Liu, Armel Zebaze, Qinkai Zheng, Binyuan Hui, Terry~Yue Zhuo, Swayam Singh, Xiangru Tang, Leandro~Von Werra, and Shayne Longpre. 2023.
\newblock \href {https://openreview.net/forum?id=CjrPqvvUXL} {Octopack: Instruction tuning code large language models}.
\newblock In \emph{NeurIPS 2023 Workshop on Instruction Tuning and Instruction Following}.

\bibitem[{Nair et~al.(2020)Nair, Roy, and Meinke}]{nair2020funcgnn}
Aravind~Ashok Nair, Avijit Roy, and Karl Meinke. 2020.
\newblock \href {https://doi.org/10.1145/3382494.3410675} {funcgnn: {A} graph neural network approach to program similarity}.
\newblock In \emph{{ESEM} '20: {ACM} / {IEEE} International Symposium on Empirical Software Engineering and Measurement, Bari, Italy, October 5-7, 2020}, pages 10:1--10:11. {ACM}.

\bibitem[{Nijkamp et~al.(2023)Nijkamp, Pang, Hayashi, Tu, Wang, Zhou, Savarese, and Xiong}]{nijkamp2023codegen}
Erik Nijkamp, Bo~Pang, Hiroaki Hayashi, Lifu Tu, Huan Wang, Yingbo Zhou, Silvio Savarese, and Caiming Xiong. 2023.
\newblock \href {https://openreview.net/pdf?id=iaYcJKpY2B\_} {Codegen: An open large language model for code with multi-turn program synthesis}.
\newblock In \emph{The Eleventh International Conference on Learning Representations, {ICLR} 2023, Kigali, Rwanda, May 1-5, 2023}. OpenReview.net.

\bibitem[{Orlanski et~al.(2023)Orlanski, Xiao, Garcia, Hui, Howland, Malmaud, Austin, Singh, and Catasta}]{pmlr-v202-orlanski23a}
Gabriel Orlanski, Kefan Xiao, Xavier Garcia, Jeffrey Hui, Joshua Howland, Jonathan Malmaud, Jacob Austin, Rishabh Singh, and Michele Catasta. 2023.
\newblock \href {https://proceedings.mlr.press/v202/orlanski23a.html} {Measuring the impact of programming language distribution}.
\newblock In \emph{International Conference on Machine Learning, {ICML} 2023, 23-29 July 2023, Honolulu, Hawaii, {USA}}, volume 202 of \emph{Proceedings of Machine Learning Research}, pages 26619--26645. {PMLR}.

\bibitem[{Paster et~al.(2023)Paster, Santos, Azerbayev, and Ba}]{paster2023openwebmath}
Keiran Paster, Marco~Dos Santos, Zhangir Azerbayev, and Jimmy Ba. 2023.
\newblock \href {https://openreview.net/forum?id=5hZTBUtkeh} {Openwebmath: An open dataset of high-quality mathematical web text}.
\newblock In \emph{The 3rd Workshop on Mathematical Reasoning and AI at NeurIPS'23}.

\bibitem[{Pfeiffer et~al.(2022)Pfeiffer, Goyal, Lin, Li, Cross, Riedel, and Artetxe}]{pfeiffer-etal-2022-lifting}
Jonas Pfeiffer, Naman Goyal, Xi~Lin, Xian Li, James Cross, Sebastian Riedel, and Mikel Artetxe. 2022.
\newblock \href {https://doi.org/10.18653/v1/2022.naacl-main.255} {Lifting the curse of multilinguality by pre-training modular transformers}.
\newblock In \emph{Proceedings of the 2022 Conference of the North American Chapter of the Association for Computational Linguistics: Human Language Technologies}, pages 3479--3495, Seattle, United States. Association for Computational Linguistics.

\bibitem[{Pian et~al.(2023)Pian, Peng, Tang, Sun, Tian, Habib, Klein, and Bissyand{\'{e}}}]{pian2023metatptrans}
Weiguo Pian, Hanyu Peng, Xunzhu Tang, Tiezhu Sun, Haoye Tian, Andrew Habib, Jacques Klein, and Tegawend{\'{e}}~F. Bissyand{\'{e}}. 2023.
\newblock \href {https://doi.org/10.1609/AAAI.V37I4.25654} {Metatptrans: {A} meta learning approach for multilingual code representation learning}.
\newblock In \emph{Thirty-Seventh {AAAI} Conference on Artificial Intelligence, {AAAI} 2023, Thirty-Fifth Conference on Innovative Applications of Artificial Intelligence, {IAAI} 2023, Thirteenth Symposium on Educational Advances in Artificial Intelligence, {EAAI} 2023, Washington, DC, USA, February 7-14, 2023}, pages 5239--5247. {AAAI} Press.

\bibitem[{Puri et~al.(2021)Puri, Kung, Janssen, Zhang, Domeniconi, Zolotov, Dolby, Chen, Choudhury, Decker, Thost, Buratti, Pujar, Ramji, Finkler, Malaika, and Reiss}]{puri2021codenet}
Ruchir Puri, David~S. Kung, Geert Janssen, Wei Zhang, Giacomo Domeniconi, Vladimir Zolotov, Julian Dolby, Jie Chen, Mihir~R. Choudhury, Lindsey Decker, Veronika Thost, Luca Buratti, Saurabh Pujar, Shyam Ramji, Ulrich Finkler, Susan Malaika, and Frederick Reiss. 2021.
\newblock \href {https://datasets-benchmarks-proceedings.neurips.cc/paper/2021/hash/a5bfc9e07964f8dddeb95fc584cd965d-Abstract-round2.html} {Codenet: {A} large-scale {AI} for code dataset for learning a diversity of coding tasks}.
\newblock In \emph{Proceedings of the Neural Information Processing Systems Track on Datasets and Benchmarks 1, NeurIPS Datasets and Benchmarks 2021, December 2021, virtual}.

\bibitem[{Quiring et~al.(2019)Quiring, Maier, and Rieck}]{quiring2019misleading}
Erwin Quiring, Alwin Maier, and Konrad Rieck. 2019.
\newblock \href {https://www.usenix.org/conference/usenixsecurity19/presentation/quiring} {Misleading authorship attribution of source code using adversarial learning}.
\newblock In \emph{28th {USENIX} Security Symposium, {USENIX} Security 2019, Santa Clara, CA, USA, August 14-16, 2019}, pages 479--496. {USENIX} Association.

\bibitem[{Rasley et~al.(2020)Rasley, Rajbhandari, Ruwase, and He}]{rasley2020deepspeed}
Jeff Rasley, Samyam Rajbhandari, Olatunji Ruwase, and Yuxiong He. 2020.
\newblock \href {https://doi.org/10.1145/3394486.3406703} {Deepspeed: System optimizations enable training deep learning models with over 100 billion parameters}.
\newblock In \emph{{KDD} '20: The 26th {ACM} {SIGKDD} Conference on Knowledge Discovery and Data Mining, Virtual Event, CA, USA, August 23-27, 2020}, pages 3505--3506. {ACM}.

\bibitem[{{Rosetta Code}(2023)}]{rosetta-code}
{Rosetta Code}. 2023.
\newblock \href {https://rosettacode.org/} {Rosetta code}.

\bibitem[{Roziere et~al.(2023)Roziere, Gehring, Gloeckle, Sootla, Gat, Tan, Adi, Liu, Remez, Rapin et~al.}]{roziere2023code}
Baptiste Roziere, Jonas Gehring, Fabian Gloeckle, Sten Sootla, Itai Gat, Xiaoqing~Ellen Tan, Yossi Adi, Jingyu Liu, Tal Remez, J{\'e}r{\'e}my Rapin, et~al. 2023.
\newblock \href {https://arxiv.org/abs/2308.12950} {Code llama: Open foundation models for code}.
\newblock \emph{arXiv preprint 2308.12950}.

\bibitem[{Rozi{\`{e}}re et~al.(2022)Rozi{\`{e}}re, Zhang, Charton, Harman, Synnaeve, and Lample}]{roziere2022leveraging}
Baptiste Rozi{\`{e}}re, Jie Zhang, Fran{\c{c}}ois Charton, Mark Harman, Gabriel Synnaeve, and Guillaume Lample. 2022.
\newblock \href {https://openreview.net/forum?id=cmt-6KtR4c4} {Leveraging automated unit tests for unsupervised code translation}.
\newblock In \emph{The Tenth International Conference on Learning Representations, {ICLR} 2022, Virtual Event, April 25-29, 2022}. OpenReview.net.

\bibitem[{Shojaee et~al.(2023)Shojaee, Jain, Tipirneni, and Reddy}]{shojaee2023executionbased}
Parshin Shojaee, Aneesh Jain, Sindhu Tipirneni, and Chandan~K. Reddy. 2023.
\newblock \href {https://openreview.net/forum?id=0XBuaxqEcG} {Execution-based code generation using deep reinforcement learning}.
\newblock \emph{Transactions on Machine Learning Research}.

\bibitem[{Soldaini and Lo(2023)}]{peS2o}
Luca Soldaini and Kyle Lo. 2023.
\newblock {peS2o (Pretraining Efficiently on S2ORC) Dataset}.
\newblock Technical report, {Allen Institute for AI}.
\newblock ODC-By, \url{https://github.com/allenai/pes2o}.

\bibitem[{Sun et~al.(2020)Sun, Zhu, Xiong, Sun, Mou, and Zhang}]{sun2020treegen}
Zeyu Sun, Qihao Zhu, Yingfei Xiong, Yican Sun, Lili Mou, and Lu~Zhang. 2020.
\newblock \href {https://aaai.org/ojs/index.php/AAAI/article/view/6430} {Treegen: {A} tree-based transformer architecture for code generation}.
\newblock In \emph{The Thirty-Fourth {AAAI} Conference on Artificial Intelligence, {AAAI} 2020, The Thirty-Second Innovative Applications of Artificial Intelligence Conference, {IAAI} 2020, The Tenth {AAAI} Symposium on Educational Advances in Artificial Intelligence, {EAAI} 2020, New York, NY, USA, February 7-12, 2020}, pages 8984--8991. {AAAI} Press.

\bibitem[{Szafraniec et~al.(2023)Szafraniec, Rozi{\`{e}}re, Leather, Labatut, Charton, and Synnaeve}]{szafraniec2023code}
Marc Szafraniec, Baptiste Rozi{\`{e}}re, Hugh Leather, Patrick Labatut, Fran{\c{c}}ois Charton, and Gabriel Synnaeve. 2023.
\newblock \href {https://openreview.net/pdf?id=XomEU3eNeSQ} {Code translation with compiler representations}.
\newblock In \emph{The Eleventh International Conference on Learning Representations, {ICLR} 2023, Kigali, Rwanda, May 1-5, 2023}. OpenReview.net.

\bibitem[{Tipirneni et~al.(2024)Tipirneni, Zhu, and Reddy}]{tipirneni2024structcoder}
Sindhu Tipirneni, Ming Zhu, and Chandan~K. Reddy. 2024.
\newblock \href {https://doi.org/10.1145/3636430} {Structcoder: Structure-aware transformer for code generation}.
\newblock \emph{{ACM} Trans. Knowl. Discov. Data}, 18(3):70:1--70:20.

\bibitem[{Tracey et~al.(2019)Tracey, Strassel, Bies, Song, Arrigo, Griffitt, Delgado, Graff, Kulick, Mott, and Kuster}]{tracey-etal-2019-corpus}
Jennifer Tracey, Stephanie Strassel, Ann Bies, Zhiyi Song, Michael Arrigo, Kira Griffitt, Dana Delgado, Dave Graff, Seth Kulick, Justin Mott, and Neil Kuster. 2019.
\newblock \href {https://aclanthology.org/W19-6808} {Corpus building for low resource languages in the {DARPA} {LORELEI} program}.
\newblock In \emph{Proceedings of the 2nd Workshop on Technologies for MT of Low Resource Languages}, pages 48--55, Dublin, Ireland. European Association for Machine Translation.

\bibitem[{Wang et~al.(2023)Wang, Li, Qian, Yang, Wang, Shang, Kumar, Tan, Ray, Bhatia, Nallapati, Ramanathan, Roth, and Xiang}]{wang-etal-2023-recode}
Shiqi Wang, Zheng Li, Haifeng Qian, Chenghao Yang, Zijian Wang, Mingyue Shang, Varun Kumar, Samson Tan, Baishakhi Ray, Parminder Bhatia, Ramesh Nallapati, Murali~Krishna Ramanathan, Dan Roth, and Bing Xiang. 2023.
\newblock \href {https://doi.org/10.18653/v1/2023.acl-long.773} {{R}e{C}ode: Robustness evaluation of code generation models}.
\newblock In \emph{Proceedings of the 61st Annual Meeting of the Association for Computational Linguistics (Volume 1: Long Papers)}, pages 13818--13843, Toronto, Canada. Association for Computational Linguistics.

\bibitem[{Wang et~al.(2020)Wang, Lipton, and Tsvetkov}]{wang-etal-2020-negative}
Zirui Wang, Zachary~C. Lipton, and Yulia Tsvetkov. 2020.
\newblock \href {https://doi.org/10.18653/v1/2020.emnlp-main.359} {On negative interference in multilingual models: Findings and a meta-learning treatment}.
\newblock In \emph{Proceedings of the 2020 Conference on Empirical Methods in Natural Language Processing (EMNLP)}, pages 4438--4450, Online. Association for Computational Linguistics.

\bibitem[{Wu et~al.(2023)Wu, Liu, and Xiao}]{wu2023deceptprompt}
Fangzhou Wu, Xiaogeng Liu, and Chaowei Xiao. 2023.
\newblock \href {https://arxiv.org/abs/2312.04730} {Deceptprompt: Exploiting llm-driven code generation via adversarial natural language instructions}.
\newblock \emph{arXiv preprint 2312.04730}.

\bibitem[{Wu and Dredze(2020)}]{wu-dredze-2020-languages}
Shijie Wu and Mark Dredze. 2020.
\newblock \href {https://doi.org/10.18653/v1/2020.repl4nlp-1.16} {Are all languages created equal in multilingual {BERT}?}
\newblock In \emph{Proceedings of the 5th Workshop on Representation Learning for NLP}, pages 120--130, Online. Association for Computational Linguistics.

\bibitem[{Xue et~al.(2021)Xue, Constant, Roberts, Kale, Al-Rfou, Siddhant, Barua, and Raffel}]{xue-etal-2021-mt5}
Linting Xue, Noah Constant, Adam Roberts, Mihir Kale, Rami Al-Rfou, Aditya Siddhant, Aditya Barua, and Colin Raffel. 2021.
\newblock \href {https://doi.org/10.18653/v1/2021.naacl-main.41} {m{T}5: A massively multilingual pre-trained text-to-text transformer}.
\newblock In \emph{Proceedings of the 2021 Conference of the North American Chapter of the Association for Computational Linguistics: Human Language Technologies}, pages 483--498, Online. Association for Computational Linguistics.

\bibitem[{Yuan et~al.(2022)Yuan, Zhang, He, Fang, Hung, Hao, and Yin}]{yuan2022circle}
Wei Yuan, Quanjun Zhang, Tieke He, Chunrong Fang, Nguyen Quoc~Viet Hung, Xiaodong Hao, and Hongzhi Yin. 2022.
\newblock \href {https://doi.org/10.1145/3533767.3534219} {{CIRCLE:} continual repair across programming languages}.
\newblock In \emph{{ISSTA} '22: 31st {ACM} {SIGSOFT} International Symposium on Software Testing and Analysis, Virtual Event, South Korea, July 18 - 22, 2022}, pages 678--690. {ACM}.

\bibitem[{Zhang et~al.(2022)Zhang, Wang, Zhang, Li, and Jin}]{zhang2022learning}
Kechi Zhang, Wenhan Wang, Huangzhao Zhang, Ge~Li, and Zhi Jin. 2022.
\newblock \href {https://doi.org/10.1145/3524610.3527905} {Learning to represent programs with heterogeneous graphs}.
\newblock In \emph{Proceedings of the 30th {IEEE/ACM} International Conference on Program Comprehension, {ICPC} 2022, Virtual Event, May 16-17, 2022}, pages 378--389. {ACM}.

\bibitem[{Zhou et~al.(2022)Zhou, Zhang, Shen, Han, Chen, and Gall}]{zhou2022adversarial}
Yu~Zhou, Xiaoqing Zhang, Juanjuan Shen, Tingting Han, Taolue Chen, and Harald~C. Gall. 2022.
\newblock \href {https://doi.org/10.1145/3501256} {Adversarial robustness of deep code comment generation}.
\newblock \emph{{ACM} Trans. Softw. Eng. Methodol.}, 31(4):60:1--60:30.

\end{thebibliography}

\appendix

\section{Experimental Details}

We employ Paged Attention~\cite{kwon2023efficient} via vLLM on model checkpoints loaded in half-precision for efficient inference while evaluating our models on zero-shot inference benchmarks. All our inference runs are conducted on Nvidia A100 80GB GPUs with 95\% of the GPU VRAM explicitly reserved for vLLMs GPU pages. We further set aside 64GB of RAM as a CPU swap, allowing for offloading pages to the CPU during bursts of long sequences. We limit the continuous batching parameter to 32 to minimize incidents of running out of swap space.

\subsection{Multipl-E}

We sample $N=$ 50 continuations of at most 1024 tokens for all Multipl-E runs. While the more common standard is to choose $N=$ 200, in the interest of efficiency, we follow existing work~\cite{li2023starcoder} that shows that one can obtain reliable \texttt{pass@k} estimates in as few as 20 generations. We always use nucleus sampling with \texttt{p} of 0.9.

Our estimates of \texttt{pass@1} emulate usage scenarios where correctness is paramount. Hence, we utilize a low temperature of 0.2. In contrast, for our \texttt{pass@10} and \texttt{pass@25} we mimic scenarios where creativity and diversity of generations are more important and hence use a higher temperature of 0.8. This practice keeps us in line with prior work~\cite{roziere2023code}.

\subsection{ReCode}

We run three of the four ReCode evaluations using HumanEval as the base dataset. The Format sub-task scrutinizes how robust these models are to source formatting variations such as turning docstrings into comments and randomly inserting newlines. The Syntax sub-task tests models' susceptibility to syntactic variation patterns common in human-written code~\cite{chakraborty2022natgen} such as dead-code blocks and renamed variables. Finally, the Function sub-task tests models' robustness to conventional variations seen in function names such as inflectional variations or synonym substitutions. We follow the benchmark authors' guidance and estimate \texttt{pass@1} from one greedily sampled output per prompt of at most 1024 tokens.

We ignore the Docstring sub-task as our pilot runs found the NLAugmenter~\cite{dhole2022nlaugmenter} transformations it uses to be an unrepresentative of the deviations found in developer prompts.

\subsection{CodeXGLUE Code-to-Text}

We greedily sample continuations capped at 512 tokens and measure their smoothed BLEU-4 score against the reference docstrings. The prompts per language are detailed below:

\begin{Verbatim}[commandchars=!\{\}]
!textcolor{CommentGrey}{# Python}
[!textcolor{Navy}{source_code}]
!textcolor{DeepForest}{""" The goal of this function is to:}

!textcolor{CommentGrey}{# Ruby}
[!textcolor{Navy}{source_code}]
!textcolor{DeepForest}{=begin The goal of this function is to:}

!textcolor{CommentGrey}{# Go}
[!textcolor{Navy}{source_code}]
!textcolor{DeepForest}{/* The goal of this function is to:}

\end{Verbatim}

\subsection{Commit Chronicle}

We randomly partition 80\%, 10\% and 10\% of the data into train, validation, and test splits for the 8 languages present in \texttt{SLTrans} --- \texttt{C}, \texttt{C++}, \texttt{Go}, \texttt{Objective-C}, \texttt{Python}, \texttt{Ruby}, \texttt{Rust} and \texttt{Swift}. For languages with a lot of diff samples, we cap the train split at 25,000 samples We train for 3 epochs with a maximum sequence length of 2048 tokens, using LoRA tuning with an \texttt{r} of 32, $\alpha$ of 16, and a batch size of 16. We use the ADAM optimizer with $\beta$ of (0.95, 0.99) and a base learning rate of 3e-4. We employ a cosine scheduler that finishes at 10\% of the base learning rate. Unlike continued pre-training, in this phase, losses are only backpropagated for the continuations.

\subsection{Instruction Tuning}

We collate 18k instruction-output pairs from EditPackFT~\cite{cassano2023edit}, which are derived by re-formatting the file contents and commit messages of single-file edit GitHub commits. In the interest of preserving natural language ability, we also source a further 5.5k code-adjacent natural language instruction-output pairs from the OASST~\cite{köpf2023openassistant} and OpenOrca~\cite{OpenOrca} collections. We perform 3 epochs of instruction tuning on all the base and \texttt{IRCoder} models with a maximum sequence length of 2048 and backpropagate losses on only the continuations. We leverage LoRA tuning with an \texttt{r} of 32, $\alpha$ of 16, and a batch size of 16. We use the ADAM optimizer with $\beta$ of (0.95, 0.99) and a base learning rate of 3e-4. We employ a cosine scheduler that finishes at 10\% of the base learning rate. Our instruction tuning template is outlined below, with losses calculated on only the completions:

\begin{Verbatim}[commandchars=!\{\}]
!textcolor{DeepForest}{### Instruction:}

!textcolor{Navy}{Text Instruction}
[!textcolor{Navy}{optional_code}]

!textcolor{DeepForest}{### Response:}
!textcolor{Saffron}{Completion} <|EOS|>

\end{Verbatim}

\subsection{HumanEvalFixDocs}

We benchmark the instruction following ability of our models using \texttt{pass@1} at temperature 0.2 and \texttt{pass@10} at temperature 0.8 by sampling 20 continuations of at most 1024 tokens. Here again, the first setting is designed to mimic factual generations, and the second is to recreate more creative settings. The task consists of the buggy code followed by the correct docstring and an accompanying instruction to fix the code snippet. This information is usually cast to the models' instruction tuning template and input as a prompt as outlined below:

\begin{Verbatim}[commandchars=!\{\}]
!textcolor{DeepForest}{### Instruction:}
!textcolor{DeepForest}{Fix bugs in [function_name]}

[!textcolor{Navy}{buggy_code}]

[!textcolor{Navy}{Correct Code Docstring}]

!textcolor{DeepForest}{### Response:}

\end{Verbatim}

\section{Detailed Results}
\label{app:det}

For completeness, we detail the split and language-wise performance of the models on all tasks discussed in \Cref{sec:results}.

\captionsetup{justification=centering,singlelinecheck=false}


\begin{table*}[t]
    \centering
    \scalebox{0.66}{
    \begin{tabular}{rccccccc}
        \toprule
        \multirow{2}{*}{\scalerel*{\includegraphics{Graphics/robot.png}}{B} \texttt{Model}} & \scalerel*{\includegraphics{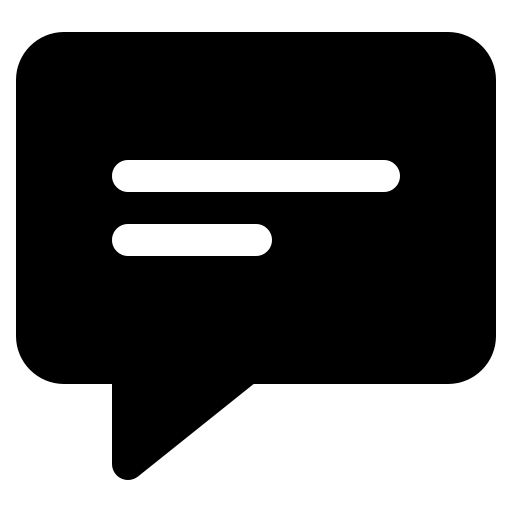}}{B} \texttt{Doc to} & \scalerel*{\includegraphics{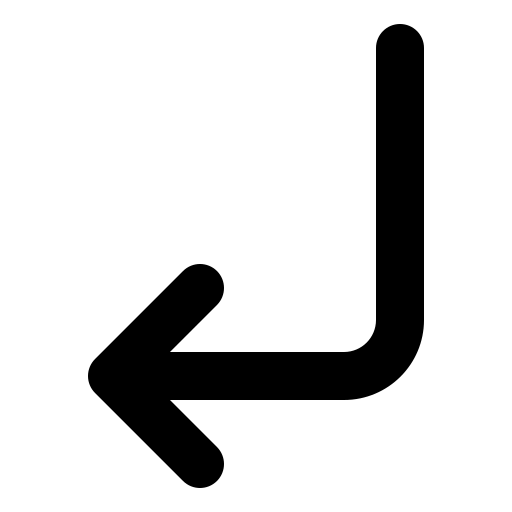}}{B} \texttt{Newline} & \scalerel*{\includegraphics{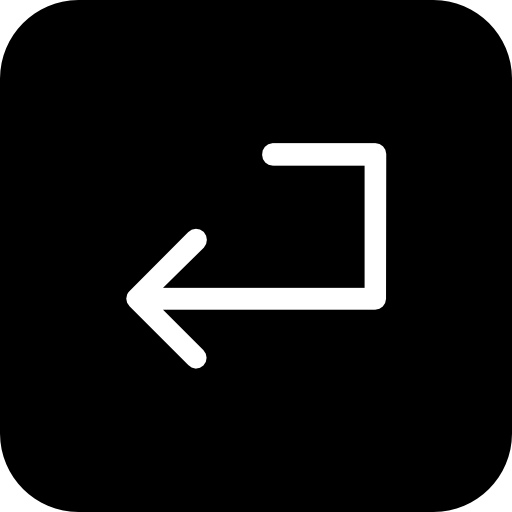}}{B} \texttt{Newline} & \scalerel*{\includegraphics{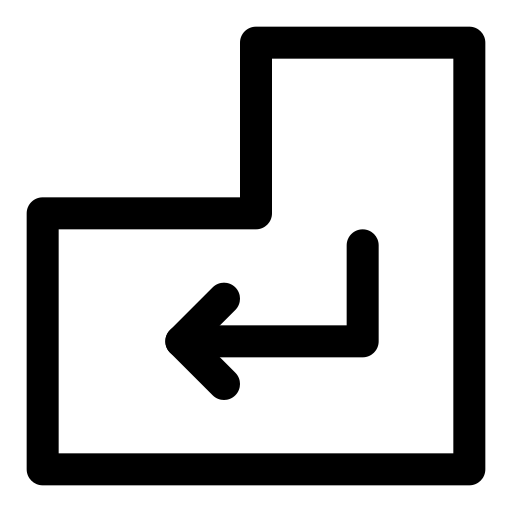}}{B} \texttt{Newline} & \scalerel*{\includegraphics{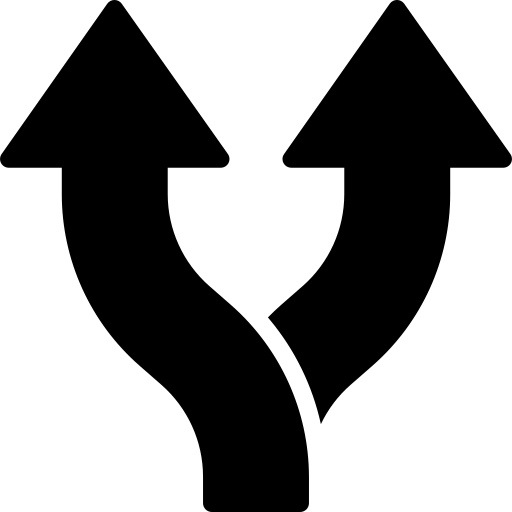}}{B} \texttt{Line} & \scalerel*{\includegraphics{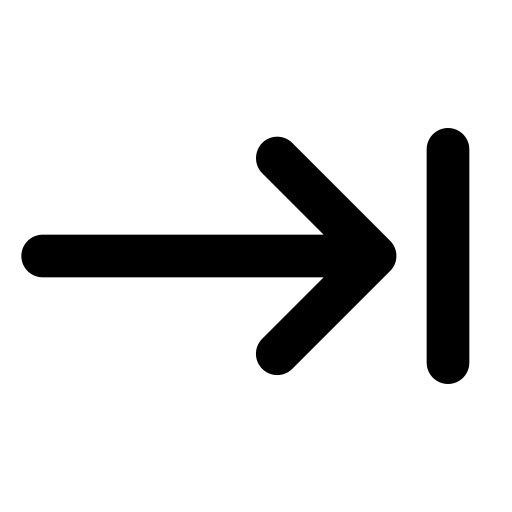}}{B} \texttt{Tab} \\
         & \texttt{Comments} & \texttt{After Code} & \texttt{After Doc} & \texttt{Random} & \texttt{Split} & \texttt{Indent} \\
        \midrule
         
        \texttt{StarCoderBase 1.1B} & \texttt{23.40} & \texttt{29.87} & \texttt{31.70} & \texttt{27.42} & \texttt{27.43} & \texttt{28.65}\\
        \texttt{DeepSeekCoder 1.3B} & \texttt{45.43} & \texttt{53.44} & \texttt{50.98} & \texttt{50.60} & \texttt{44.18} & \texttt{53.04}\\
        \texttt{StarCoderBase 3.1B} & \texttt{34.78} & \texttt{39.63} & \texttt{40.85} & \texttt{37.80} & \texttt{40.11} & \texttt{39.02} \\
        \texttt{DeepSeekCoder 5.7B} & \texttt{50.12} & \texttt{61.42} & \texttt{66.38} & \texttt{64.63} & \texttt{64.67} & \texttt{66.98}\\
        \texttt{CodeLlama 6.7B} & \texttt{50.11} & \texttt{56.09} & \texttt{55.87} & \texttt{54.43} & \texttt{52.58} & \texttt{57.92}\\
        \texttt{StarCoderBase 7.3B} & \texttt{40.11} & \texttt{48.59} & \texttt{47.88} & \texttt{44.89} &\texttt{47.53}  & \texttt{48.78}\\
        \midrule
        \multirow{2}{*}{\texttt{IRCoder 1.1B}} & \texttt{26.14} & \texttt{30.48} & \texttt{34.48} & \texttt{29.21} & \texttt{30.70} & \texttt{30.04} \vspace{-0.2em}\\ 
         & \diffup{2.74} & \diffup{0.61} & \diffup{2.78} & \diffup{1.78} & \diffup{3.27} & \diffup{1.39}\vspace{0.2em}\\
        \multirow{2}{*}{\texttt{IRCoder 1.3B}} & \texttt{42.46} & \texttt{53.97} & \texttt{51.44} & \texttt{49.44} & \texttt{46.34} & \texttt{55.23}\vspace{-0.2em}\\
         & \diffdo{2.75} & \diffup{0.53} & \diffup{0.46} & \diffdo{1.16} & \diffup{2.16} & \diffup{2.19}\vspace{0.2em}\\
        \multirow{2}{*}{\texttt{IRCoder 3.1B}} & \texttt{37.14} & \texttt{40.24} & \texttt{39.63} & \texttt{40.41} & \texttt{40.41} & \texttt{40.85}\vspace{-0.2em}\\
         & \diffup{2.36} & \diffup{0.61} & \diffdo{1.22} & \diffup{2.61} & \diffup{0.30} & \diffup{1.83}\vspace{0.2em}\\
        \multirow{2}{*}{\texttt{IRCoder 5.7B}} & \texttt{57.31} & \texttt{68.28} & \texttt{68.90} & \texttt{66.74} & \texttt{64.98} & \texttt{68.36}\vspace{-0.2em}\\
         & \diffup{7.19} & \diffup{6.86} & \diffup{2.52} & \diffup{2.11} & \diffup{0.31} & \diffup{1.38}\vspace{0.2em}\\
        \multirow{2}{*}{\texttt{IRCoder 6.7B}} & \texttt{54.18} & \texttt{55.85} & \texttt{57.92} & \texttt{55.48} & \texttt{55.71} & \texttt{59.33}\vspace{-0.2em}\\
         & \diffup{4.07} & \diffdo{0.24} & \diffup{2.05} &\diffup{1.05}  & \diffup{3.13} & \diffup{1.41}\vspace{0.2em}\\
        \multirow{2}{*}{\texttt{IRCoder 7.3B}} & \texttt{40.85} & \texttt{43.98} & \texttt{48.06} & \texttt{48.06} & \texttt{49.31} & \texttt{49.37}\vspace{-0.2em}\\
         & \diffup{0.74} & \diffdo{4.61} & \diffup{0.18} & \diffup{3.17} & \diffup{1.88} & \diffup{0.59}\\
        \bottomrule
    \end{tabular}
    }
    \caption{ReCode Format \texttt{pass@1} comparison between \texttt{IRCoder} and its corresponding base models.}
        \label{table:recode-format}
\end{table*}

\begin{table*}[t]
    \centering
    \scalebox{0.66}{
    \begin{tabular}{rccccccc}
        \toprule
        \multirow{3}{*}{\scalerel*{\includegraphics{Graphics/robot.png}}{B} \texttt{Model}} & \scalerel*{\includegraphics{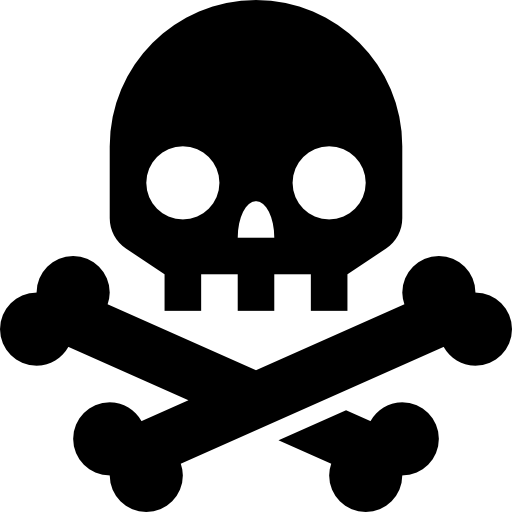}}{B} \texttt{Dead} & \scalerel*{\includegraphics{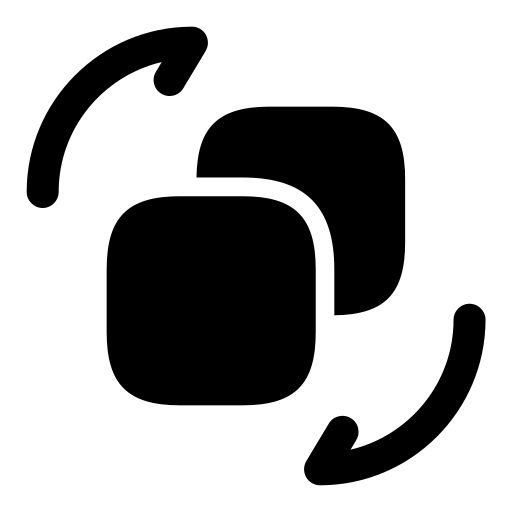}}{B} \texttt{For} & \scalerel*{\includegraphics{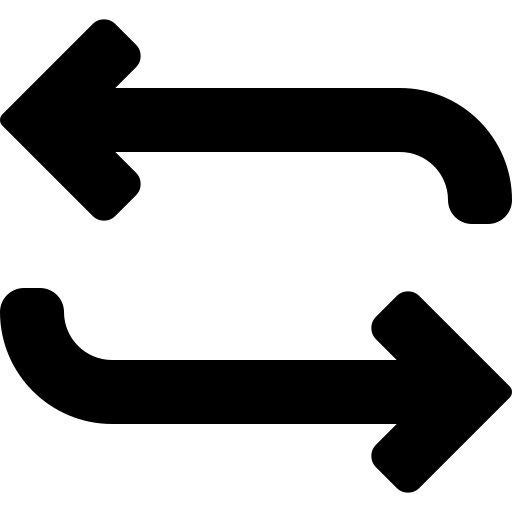}}{B} \texttt{Operand} & \scalerel*{\includegraphics{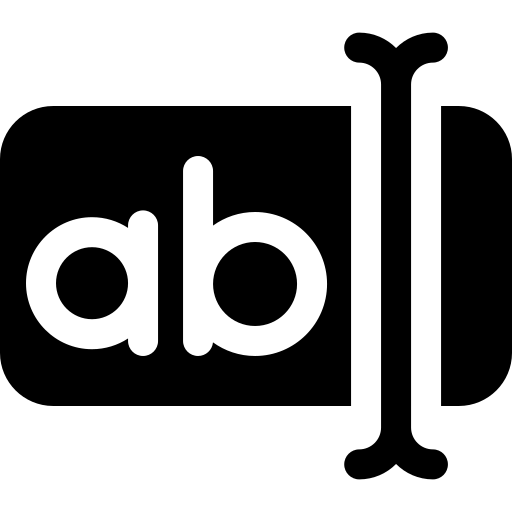}}{B} \texttt{Var} & \scalerel*{\includegraphics{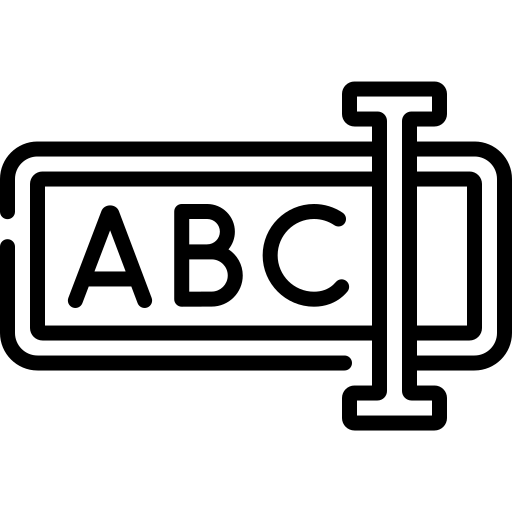}}{B} \texttt{Var} & \scalerel*{\includegraphics{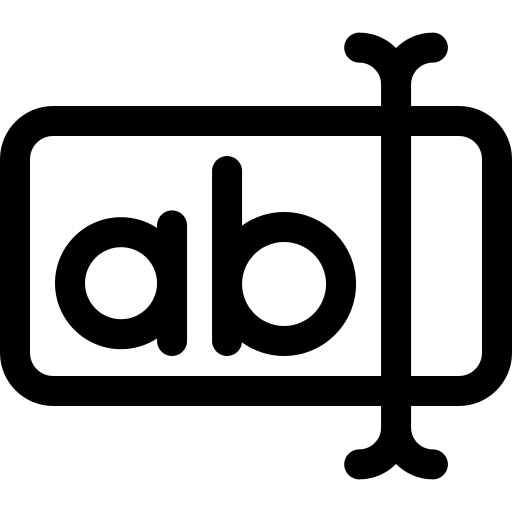}}{B} \texttt{Var} \\
        & \texttt{Code} & \texttt{While} & \texttt{Swap} & \texttt{Renaming} & \texttt{Renaming} & \texttt{Renaming} \\
        & \texttt{Insert} & \texttt{Transform} & & \texttt{CB} & \texttt{Naive} & \texttt{RN} \\
        \midrule
         
        \texttt{StarCoderBase 1.1B} & \texttt{8.53} & \texttt{32.93} & \texttt{29.14} & \texttt{31.70} & \texttt{29.87} & \texttt{26.18}\\
        \texttt{DeepSeekCoder 1.3B} & \texttt{17.64} & \texttt{52.43} & \texttt{50.52} & \texttt{50.52} & \texttt{50.60} & \texttt{47.56}\\
        \texttt{StarCoderBase 3.1B} & \texttt{14.02} & \texttt{38.41} & \texttt{39.63} & \texttt{38.11} & \texttt{37.81} & \texttt{31.78}\\
        \texttt{DeepSeekCoder 5.7B} & \texttt{23.48} & \texttt{64.35} & \texttt{61.63} & \texttt{64.02} & \texttt{60.52} & \texttt{58.56}\\
        \texttt{CodeLlama 6.7B} & \texttt{16.94} & \texttt{52.95} & \texttt{51.97} & \texttt{54.26} & \texttt{51.78} & \texttt{43.47}\\
        \texttt{StarCoderBase 7.3B} & \texttt{14.96} & \texttt{50.31} & \texttt{50.31} & \texttt{45.63} & \texttt{45.12} & \texttt{42.68}\\
        \midrule
        \multirow{2}{*}{\texttt{IRCoder 1.1B}} & \texttt{10.36} & \texttt{33.94} & \texttt{31.70} & \texttt{32.31} & \texttt{30.60} & \texttt{26.11}\vspace{-0.2em}\\
         & \diffup{1.83} & \diffup{1.01} & \diffup{2.56} & \diffup{0.61} & \diffup{0.73} & \diffdo{0.07}\vspace{0.2em}\\
        \multirow{2}{*}{\texttt{IRCoder 1.3B}} & \texttt{18.29} & \texttt{49.96} & \texttt{50.60} & \texttt{49.86} & \texttt{54.36} & \texttt{49.51}\vspace{-0.2em}\\
         & \diffup{0.65} & \diffdo{2.47} & \diffup{0.08} & \diffdo{0.66} & \diffup{3.76} & \diffup{1.95}\vspace{0.2em}\\
        \multirow{2}{*}{\texttt{IRCoder 3.1B}} & \texttt{12.19} & \texttt{40.11} & \texttt{41.87} & \texttt{43.11} & \texttt{36.93} & \texttt{32.31}\vspace{-0.2em}\\
         & \diffdo{1.83} & \diffup{1.70} & \diffup{2.24} & \diffup{5.00} & \diffdo{0.88} & \diffup{0.53}\vspace{0.2em}\\
        \multirow{2}{*}{\texttt{IRCoder 5.7B}} & \texttt{26.92} & \texttt{66.47} & \texttt{65.84} & \texttt{67.68} & \texttt{66.46} & \texttt{62.19}\vspace{-0.2em}\\
         & \diffup{3.34} & \diffup{2.12} & \diffup{4.21} & \diffup{3.66} & \diffup{5.94} & \diffup{3.63}\vspace{0.2em}\\
        \multirow{2}{*}{\texttt{IRCoder 6.7B}} & \texttt{18.90} & \texttt{56.09} & \texttt{55.65} & \texttt{54.44} & \texttt{54.41} & \texttt{49.17}\vspace{-0.2em}\\
         & \diffup{1.96} & \diffup{3.14} & \diffup{3.68} & \diffup{0.18} & \diffup{2.36} & \diffup{5.70}\vspace{0.2em}\\
        \multirow{2}{*}{\texttt{IRCoder 7.3B}} & \texttt{14.85} & \texttt{49.77} & \texttt{50.46} & \texttt{49.63} & \texttt{45.85} & \texttt{40.36}\vspace{-0.2em}\\
         & \diffdo{0.11} & \diffdo{0.54} & \diffup{0.15} & \diffup{4.00} & \diffup{0.73} & \diffdo{2.32}\vspace{0.2em}\\
        \bottomrule
    \end{tabular}
    }
    \caption{ReCode Syntax \texttt{pass@1} comparison between \texttt{IRCoder} and its corresponding base models.}
        \label{table:recode-syntax}
\end{table*}

\begin{table*}[t]
    \centering
    \scalebox{0.66}{
    \begin{tabular}{rccccccc}
        \toprule
        \multirow{3}{*}{\scalerel*{\includegraphics{Graphics/robot.png}}{B} \texttt{Model}} & \scalerel*{\includegraphics{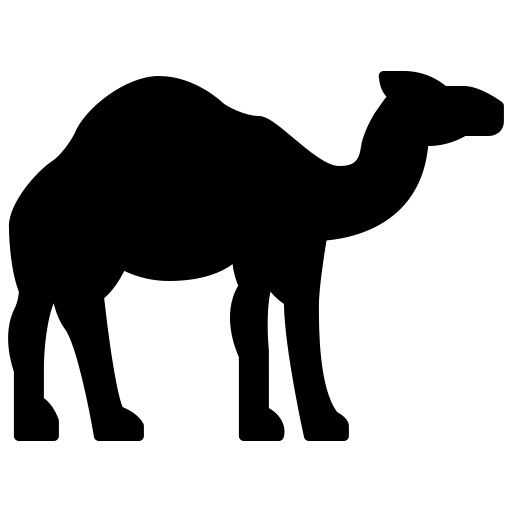}}{B} \texttt{Camel} & \scalerel*{\includegraphics{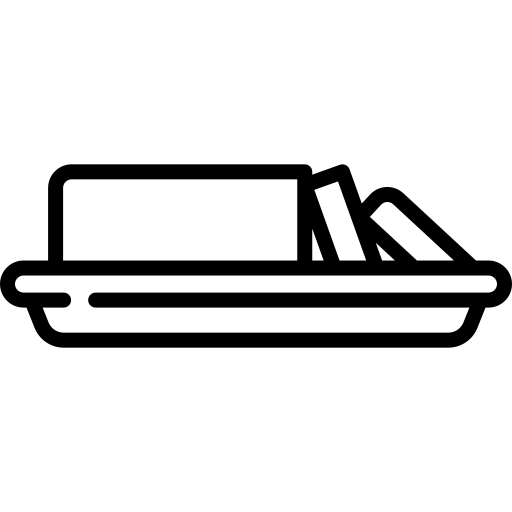}}{B} \texttt{Butter} & \scalerel*{\includegraphics{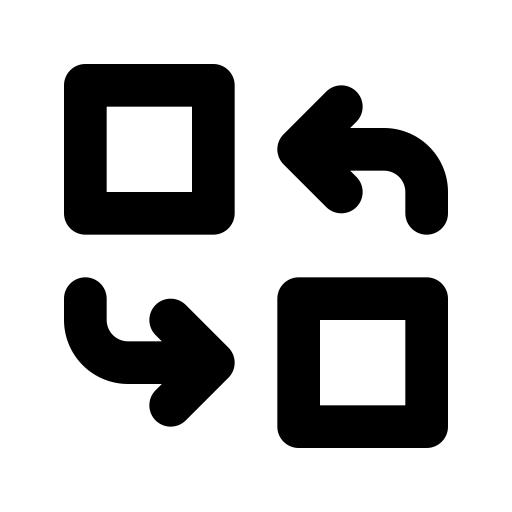}}{B} \texttt{Swap} & \scalerel*{\includegraphics{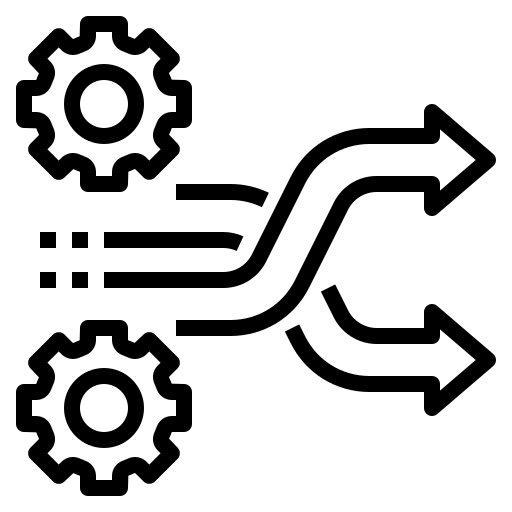}}{B} \texttt{Change} & \scalerel*{\includegraphics{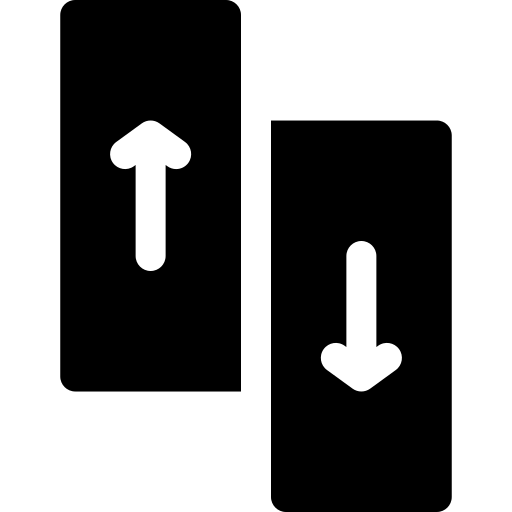}}{B} \texttt{Inflectional} & \scalerel*{\includegraphics{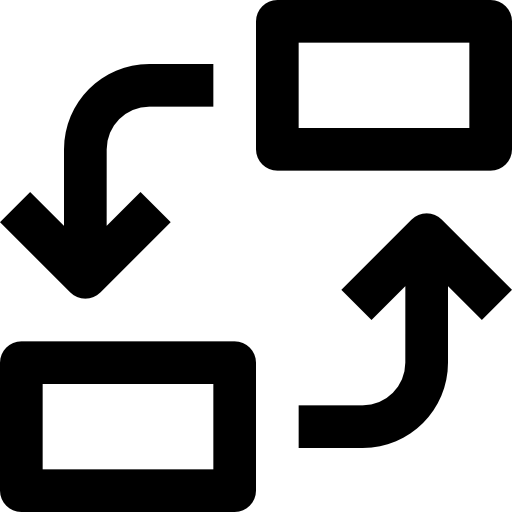}}{B} \texttt{Synonym} \\
        & \texttt{Case} & \texttt{Fingers} & \texttt{Characters} & \texttt{Character} & \texttt{Variation} & \texttt{Substitution} \\
        &  &  & & \texttt{Case} &  &  \\
        \midrule
        \texttt{StarCoderBase 1.1B} & \texttt{10.24} & \texttt{11.44} & \texttt{13.14} & \texttt{10.33} & \texttt{12.44} & \texttt{10.26}\\
        \texttt{DeepSeekCoder 1.3B} & \texttt{26.99} & \texttt{23.93} & \texttt{26.12} & \texttt{19.65} & \texttt{25.67} & \texttt{28.41}\\
        \texttt{StarCoderBase 3.1B} & \texttt{20.87} & \texttt{18.02} & \texttt{19.42} & \texttt{17.07} & \texttt{20.32} & \texttt{18.56}\\
        \texttt{DeepSeekCoder 5.7B} & \texttt{40.24} & \texttt{34.32} & \texttt{38.47} & \texttt{30.26} & \texttt{40.33} & \texttt{36.76} \\
        \texttt{CodeLlama 6.7B} & \texttt{27.44} & \texttt{25.11} & \texttt{24.41} & \texttt{21.08} & \texttt{23.79} & \texttt{25.09}\\
        \texttt{StarCoderBase 7.3B} & \texttt{26.33} & \texttt{25.02} & \texttt{22.17} & \texttt{19.68} & \texttt{25.39} & \texttt{22.56} \\
        \midrule
        \multirow{2}{*}{\texttt{IRCoder 1.1B}} & \texttt{11.82} & \texttt{11.36} & \texttt{12.49} & \texttt{13.02} & \texttt{12.63} & \texttt{10.76}\vspace{-0.2em}\\
         & \diffup{1.58} & \diffdo{0.08} & \diffdo{0.65} & \diffup{2.69} & \diffup{0.19} & \diffup{0.50}\vspace{0.2em}\\
        \multirow{2}{*}{\texttt{IRCoder 1.3B}} & \texttt{29.94} & \texttt{23.17} & \texttt{25.67} & \texttt{22.06}
        & \texttt{26.76} & \texttt{26.88}\vspace{-0.2em}\\
         & \diffup{2.95} & \diffdo{0.76} & \diffdo{0.45} & \diffup{2.41} & \diffup{1.09} & \diffdo{1.53}\vspace{0.2em}\\
        \multirow{2}{*}{\texttt{IRCoder 3.1B}} & \texttt{21.88} & \texttt{17.67} & \texttt{18.39} & \texttt{15.12} & \texttt{20.68} & \texttt{19.07}\vspace{-0.2em}\\
         & \diffup{1.01} & \diffdo{0.35} & \diffdo{1.03} & \diffdo{1.95}& \diffup{0.36} & \diffup{0.51}\vspace{0.2em}\\
        \multirow{2}{*}{\texttt{IRCoder 5.7B}} & \texttt{43.86} & \texttt{37.84} & \texttt{37.19} & \texttt{34.57} & \texttt{39.91} & \texttt{38.58} \vspace{-0.2em}\\
         & \diffup{3.62} & \diffup{3.52} & \diffdo{1.28} & \diffup{4.31} & \diffdo{0.42} & \diffup{1.82} \vspace{0.2em}\\
        \multirow{2}{*}{\texttt{IRCoder 6.7B}} & \texttt{27.11} & \texttt{25.17} & \texttt{24.56} & \texttt{25.78} & \texttt{25.44} & \texttt{24.77} \vspace{-0.2em}\\
         & \diffdo{0.33} & \diffup{0.06} & \diffup{0.15} & \diffup{4.70} & \diffup{1.65} & \diffdo{0.32}\vspace{0.2em}\\
        \multirow{2}{*}{\texttt{IRCoder 7.3B}} & \texttt{26.94} & \texttt{24.07} & \texttt{22.51} & \texttt{20.84} & \texttt{25.92} & \texttt{22.30} \vspace{-0.2em}\\
         & \diffup{0.61} & \diffdo{0.95} & \diffup{0.34} & \diffup{1.16} & \diffup{0.53} & \diffdo{0.26}\vspace{0.2em}\\
        \bottomrule
    \end{tabular}
    }
    \caption{ReCode Function \texttt{pass@1} comparison between \texttt{IRCoder} and its corresponding base models.}
        \label{table:recode-function}
\end{table*}


\begin{table*}[t]
    \centering
    \scalebox{0.66}{
    \begin{tabular}{rccccccc}
        \toprule
        \scalerel*{\includegraphics{Graphics/robot.png}}{B} \texttt{Model} & \scalerel*{\includegraphics{Graphics/cpp.png}}{B} \texttt{C++} & \scalerel*{\includegraphics{Graphics/d.png}}{B} \texttt{D} & \scalerel*{\includegraphics{Graphics/go.png}}{B} \texttt{Go} & \scalerel*{\includegraphics{Graphics/python.png}}{B} \texttt{Python} & \scalerel*{\includegraphics{Graphics/ruby.png}}{B} \texttt{Ruby} & \scalerel*{\includegraphics{Graphics/rust.png}}{B} \texttt{Rust} & \scalerel*{\includegraphics{Graphics/swift.png}}{B} \texttt{Swift} \\
        \midrule
         
        \texttt{StarCoderBase 1.1B} & \texttt{10.22} & \texttt{3.87} & \texttt{12.79} & \texttt{14.26} & \texttt{4.46} & \texttt{9.21} & \texttt{3.64}\\
        \texttt{DeepSeekCoder 1.3B} & \texttt{28.21} & \texttt{9.77} & \texttt{15.87} & \texttt{27.91} & \texttt{21.21} & \texttt{16.46} & \texttt{8.94}\\
        \texttt{StarCoderBase 3.1B} & \texttt{16.64} & \texttt{4.89} & \texttt{15.63} & \texttt{21.51} & \texttt{4.52} & \texttt{16.31} & \texttt{9.98}\\
        \texttt{DeepSeekCoder 5.7B} & \texttt{43.44} & \texttt{13.65} & \texttt{24.64} & \texttt{42.67} & \texttt{33.43} & \texttt{31.79} & \texttt{23.79}\\
        \texttt{CodeLlama 6.7B} & \texttt{26.72} & \texttt{9.67} & \texttt{18.69} & \texttt{31.13} & \texttt{25.28} & \texttt{21.43} & \texttt{19.87}\\
        \texttt{StarCoderBase 7.3B} & \texttt{23.19} & \texttt{7.62} & \texttt{16.76} & \texttt{27.88} & \texttt{16.96} & \texttt{18.81} & \texttt{14.38}\\
        \midrule
        \multirow{2}{*}{\texttt{IRCoder 1.1B}} & \texttt{11.10} & \texttt{4.65} & \texttt{11.78} & \texttt{14.29} & \texttt{6.34} & \texttt{9.62} & \texttt{3.76}\vspace{-0.2em}\\
         & \diffup{0.88} & \diffup{0.78} & \diffdo{1.01} & \diffup{0.03} & \diffup{1.88} & \diffup{0.21} & \diffup{0.12}\vspace{0.2em}\\
        \multirow{2}{*}{\texttt{IRCoder 1.3B}} & \texttt{31.79} & \texttt{10.57} & \texttt{16.17} & \texttt{30.61} & \texttt{24.35} & \texttt{20.91} & \texttt{9.14}\vspace{-0.2em}\\
         & \diffup{3.58} & \diffup{0.80} & \diffup{0.30} & \diffup{2.70} & \diffup{3.14} & \diffup{4.45} & \diffup{0.20}\vspace{0.2em}\\
        \multirow{2}{*}{\texttt{IRCoder 3.1B}} & \texttt{16.87} & \texttt{5.67} & \texttt{17.78} & \texttt{21.98} & \texttt{11.46} & \texttt{16.78} & \texttt{9.96}\vspace{-0.2em}\\
         & \diffup{0.23} & \diffup{0.78} & \diffup{2.15} & \diffup{0.47} & \diffup{6.94} & \diffup{0.47} & \diffdo{0.02}\vspace{0.2em}\\
        \multirow{2}{*}{\texttt{IRCoder 5.7B}} & \texttt{45.61} & \texttt{15.96} & \texttt{23.77} & \texttt{42.92} & \texttt{34.60} & \texttt{33.94} & \texttt{21.17}\vspace{-0.2em}\\
         & \diffup{2.17} & \diffup{2.41} & \diffdo{0.87} & \diffup{0.25} & \diffup{1.17} & \diffup{2.15} & \diffdo{2.62}\vspace{0.2em}\\
        \multirow{2}{*}{\texttt{IRCoder 6.7B}} & \texttt{29.12} & \texttt{13.02} & \texttt{19.10} & \texttt{31.11} & \texttt{26.28} & \texttt{24.37} & \texttt{25.45}\vspace{-0.2em}\\
         & \diffup{2.40} &\diffup{3.35}  & \diffup{0.41} & \diffdo{0.02} & \diffup{1.00} & \diffup{2.94} & \diffup{5.58}\vspace{0.2em}\\
        \multirow{2}{*}{\texttt{IRCoder 7.3B}} & \texttt{23.06} & \texttt{11.97} & \texttt{16.81} & \texttt{25.24} & \texttt{19.52} & \texttt{19.63} & \texttt{12.99}\vspace{-0.2em}\\
         & \diffdo{0.13} & \diffup{4.35} & \diffup{0.05} & \diffdo{2.64} & \diffup{2.56} & \diffup{0.82} & \diffdo{1.39}\\
        \bottomrule
    \end{tabular}
    }
    \caption{Multipl-E \texttt{pass@1} comparison between \texttt{IRCoder} and its corresponding base models.}
        \label{table:multiple-1}
\end{table*}


\begin{table*}[t]
    \centering
    \scalebox{0.66}{
    \begin{tabular}{rccccccc}
        \toprule
        \scalerel*{\includegraphics{Graphics/robot.png}}{B} \texttt{Model} & \scalerel*{\includegraphics{Graphics/cpp.png}}{B} \texttt{C++} & \scalerel*{\includegraphics{Graphics/d.png}}{B} \texttt{D} & \scalerel*{\includegraphics{Graphics/go.png}}{B} \texttt{Go} & \scalerel*{\includegraphics{Graphics/python.png}}{B} \texttt{Python} & \scalerel*{\includegraphics{Graphics/ruby.png}}{B} \texttt{Ruby} & \scalerel*{\includegraphics{Graphics/rust.png}}{B} \texttt{Rust} & \scalerel*{\includegraphics{Graphics/swift.png}}{B} \texttt{Swift} \\
        \midrule
         
        \texttt{StarCoderBase 1.1B} & \texttt{20.81} & \texttt{7.81} & \texttt{17.33} & \texttt{19.97} & \texttt{6.92} & \texttt{12.19} & \texttt{8.87}\\
        \texttt{DeepSeekCoder 1.3B} & \texttt{34.36} & \texttt{16.76} & \texttt{18.98} & \texttt{41.30} & \texttt{37.86} & \texttt{18.94} & \texttt{14.65}\\
        \texttt{StarCoderBase 3.1B} & \texttt{24.16} & \texttt{11.06} & \texttt{18.11} & \texttt{30.86} & \texttt{10.39} & \texttt{22.61} & \texttt{16.43} \\
        \texttt{DeepSeekCoder 5.7B} & \texttt{54.67} & \texttt{22.74} & \texttt{28.78} & \texttt{61.53} & \texttt{46.24} & \texttt{42.96} & \texttt{32.76}\\
        \texttt{CodeLlama 6.7B} & \texttt{45.86} & \texttt{15.56} & \texttt{23.11} & \texttt{52.75} & \texttt{45.87} & \texttt{28.67} & \texttt{31.65}\\
        \texttt{StarCoderBase 7.3B} & \texttt{36.97} & \texttt{16.14} & \texttt{20.21} & \texttt{40.58} & \texttt{33.42} & \texttt{25.63} & \texttt{18.69}\\
        \midrule
        \multirow{2}{*}{\texttt{IRCoder 1.1B}} & \texttt{21.24} & \texttt{9.07} & \texttt{17.84} & \texttt{22.61} & \texttt{9.76} & \texttt{11.86} & \texttt{9.17}\vspace{-0.2em}\\
         & \diffup{0.43} & \diffup{1.18} & \diffup{0.51} & \diffup{2.64} & \diffup{2.84} & \diffdo{0.33} & \diffup{0.30}\vspace{0.2em}\\
        \multirow{2}{*}{\texttt{IRCoder 1.3B}} & \texttt{36.42} & \texttt{24.27} & \texttt{18.95} & \texttt{48.67} & \texttt{42.39} & \texttt{25.48} & \texttt{21.82}\vspace{-0.2em}\\
         & \diffup{5.17} & \diffdo{0.13} & \diffup{1.12} & \diffup{4.36} & \diffup{12.54} & \diffdo{0.04} & \diffup{1.41}\vspace{0.2em}\\
        \multirow{2}{*}{\texttt{IRCoder 3.1B}} & \texttt{29.33} & \texttt{10.93} & \texttt{19.23} & \texttt{35.22} & \texttt{22.93} & \texttt{22.57} & \texttt{17.84}\vspace{-0.2em}\\
         & \diffup{5.17} & \diffdo{0.13} & \diffup{1.12} & \diffup{4.36} & \diffup{12.54} & \diffdo{0.04} & \diffup{1.41}\vspace{0.2em}\\
        \multirow{2}{*}{\texttt{IRCoder 5.7B}} & \texttt{58.51} & \texttt{28.64} & \texttt{28.59} & \texttt{68.11} & \texttt{49.58} & \texttt{44.09} & \texttt{37.45}\vspace{-0.2em}\\
         & \diffup{3.84} & \diffup{5.90} & \diffdo{0.19} & \diffup{6.58} & \diffup{3.34} & \diffup{1.13} & \diffup{4.69}\vspace{0.2em}\\
        \multirow{2}{*}{\texttt{IRCoder 6.7B}} & \texttt{51.76} & \texttt{26.14} & \texttt{25.48} & \texttt{56.71} & \texttt{49.40} & \texttt{31.28} & \texttt{34.89}\vspace{-0.2em}\\
         & \diffup{5.90} & \diffup{10.58} & \diffup{2.37} & \diffup{3.96} & \diffup{3.53} & \diffup{2.61} & \diffup{3.24}\vspace{0.2em}\\
        \multirow{2}{*}{\texttt{IRCoder 7.3B}} & \texttt{39.24} & \texttt{20.23} & \texttt{19.68} & \texttt{45.69} & \texttt{36.88} & \texttt{27.65} & \texttt{23.45}\vspace{-0.2em}\\
         & \diffup{2.27} & \diffup{4.09} & \diffdo{0.53} & \diffup{5.31} & \diffup{3.46} & \diffup{2.02} & \diffup{4.76}\\
        \bottomrule
    \end{tabular}
    }
    \caption{Multipl-E \texttt{pass@10} comparison between \texttt{IRCoder} and its corresponding base models.}
        \label{table:multiple-10}
\end{table*}


\begin{table*}[t]
    \centering
    \scalebox{0.66}{
    \begin{tabular}{rccccccc}
        \toprule
        \scalerel*{\includegraphics{Graphics/robot.png}}{B} \texttt{Model} & \scalerel*{\includegraphics{Graphics/cpp.png}}{B} \texttt{C++} & \scalerel*{\includegraphics{Graphics/d.png}}{B} \texttt{D} & \scalerel*{\includegraphics{Graphics/go.png}}{B} \texttt{Go} & \scalerel*{\includegraphics{Graphics/python.png}}{B} \texttt{Python} & \scalerel*{\includegraphics{Graphics/ruby.png}}{B} \texttt{Ruby} & \scalerel*{\includegraphics{Graphics/rust.png}}{B} \texttt{Rust} & \scalerel*{\includegraphics{Graphics/swift.png}}{B} \texttt{Swift} \\
        \midrule
         
        \texttt{StarCoderBase 1.1B} & \texttt{28.19} & \texttt{12.78} & \texttt{19.22} & \texttt{23.04} & \texttt{7.65} & \texttt{13.45} & \texttt{10.69}\\
        \texttt{DeepSeekCoder 1.3B} & \texttt{38.44} & \texttt{20.71} & \texttt{22.79} & \texttt{50.68} & \texttt{43.26} & \texttt{20.13} & \texttt{23.49}\\
        \texttt{StarCoderBase 3.1B} & \texttt{27.38} & \texttt{13.98} & \texttt{20.87} & \texttt{33.89} & \texttt{14.59} & \texttt{24.97} & \texttt{22.68}\\
        \texttt{DeepSeekCoder 5.7B} & \texttt{58.59} & \texttt{28.32} & \texttt{31.11} & \texttt{69.14} & \texttt{53.76} & \texttt{50.84} & \texttt{44.55}\\
        \texttt{CodeLlama 6.7B} & \texttt{57.65} & \texttt{22.86} & \texttt{25.73} & \texttt{62.43} & \texttt{55.96} & \texttt{31.95} & \texttt{40.93}\\
        \texttt{StarCoderBase 7.3B} & \texttt{40.28} & \texttt{20.93} & \texttt{22.14} & \texttt{53.44} & \texttt{44.85} & \texttt{30.80} & \texttt{26.41}\\
        \midrule
        \multirow{2}{*}{\texttt{IRCoder 1.1B}} & \texttt{29.79} & \texttt{13.45} & \texttt{20.02} & \texttt{28.43} & \texttt{18.79} & \texttt{14.31} & \texttt{10.43}\vspace{-0.2em}\\
         & \diffup{1.60} & \diffup{0.67} & \diffup{0.80} & \diffup{5.39} & \diffup{11.14} & \diffup{0.86} & \diffdo{0.26}\vspace{0.2em}\\
        \multirow{2}{*}{\texttt{IRCoder 1.3B}} & \texttt{40.55} & \texttt{31.89} & \texttt{23.47} & \texttt{57.84} & \texttt{52.46} & \texttt{28.63} & \texttt{30.44}\vspace{-0.2em}\\
         & \diffup{2.11} & \diffup{11.18} & \diffup{0.68} & \diffup{7.16} & \diffup{9.20} & \diffup{8.50} & \diffup{6.95} \vspace{0.2em}\\
        \multirow{2}{*}{\texttt{IRCoder 3.1B}} & \texttt{35.18} & \texttt{14.77} & \texttt{23.59} & \texttt{46.19} & \texttt{28.72} & \texttt{24.12} & \texttt{23.55}\vspace{-0.2em}\\
         & \diffup{7.80} & \diffup{0.79} & \diffup{2.72} & \diffup{12.30} & \diffup{14.13} & \diffdo{0.85} & \diffup{0.87}\vspace{0.2em}\\
        \multirow{2}{*}{\texttt{IRCoder 5.7B}} & \texttt{62.16} & \texttt{33.79} & \texttt{33.86} & \texttt{73.41} & \texttt{55.08} & \texttt{51.69} & \texttt{49.04}\vspace{-0.2em}\\
         & \diffup{3.57} & \diffup{5.47} & \diffup{2.75} & \diffup{4.27} & \diffup{1.32} & \diffup{0.85} & \diffup{4.49} \vspace{0.2em}\\
        \multirow{2}{*}{\texttt{IRCoder 6.7B}} & \texttt{60.08} & \texttt{35.27} & \texttt{26.31} & \texttt{67.49} & \texttt{59.88} & \texttt{35.39} & \texttt{44.77}\vspace{-0.2em}\\
         & \diffup{2.43} & \diffup{12.41} & \diffup{0.58} & \diffup{5.06} & \diffup{3.92} & \diffup{3.44} & \diffup{3.84}\vspace{0.2em}\\
        \multirow{2}{*}{\texttt{IRCoder 7.3B}} & \texttt{47.87} & \texttt{27.67} & \texttt{21.87} & \texttt{56.47} & \texttt{49.35} & \texttt{30.92} & \texttt{32.12}\vspace{-0.2em}\\
         & \diffup{7.59} & \diffup{6.74} & \diffdo{0.27} & \diffup{3.03} & \diffup{4.50} & \diffup{0.12} & \diffup{5.71}\\
        \bottomrule
    \end{tabular}
    }
    \caption{Multipl-E \texttt{pass@25} comparison between \texttt{IRCoder} and its corresponding base models.}
        \label{table:multiple-25}
\end{table*}


\begin{table*}[t]
    \centering
    \scalebox{0.66}{
    \begin{tabular}{rccc}
        \toprule
        \scalerel*{\includegraphics{Graphics/robot.png}}{B} \texttt{Model} & \scalerel*{\includegraphics{Graphics/go.png}}{B} \texttt{Go} & \scalerel*{\includegraphics{Graphics/python.png}}{B} \texttt{Python} & \scalerel*{\includegraphics{Graphics/ruby.png}}{B} \texttt{Ruby} \\
        \midrule
         
        \texttt{StarCoderBase 1.1B} & \texttt{10.23} & \texttt{12.89} & \texttt{7.03}  \\
        \texttt{DeepSeekCoder 1.3B} & \texttt{11.29} & \texttt{14.07} & \texttt{3.52} \\
        \texttt{StarCoderBase 3.1B} & \texttt{10.33} & \texttt{12.78} & \texttt{8.71} \\
        \texttt{DeepSeekCoder 5.7B} & \texttt{10.09} & \texttt{14.15} & \texttt{11.16} \\
        \texttt{CodeLlama 6.7B} & \texttt{9.96} & \texttt{14.33} & \texttt{4.94} \\
        \texttt{StarCoderBase 7.3B} & \texttt{9.54} & \texttt{13.52} & \texttt{9.17} \\
        \midrule
        \multirow{2}{*}{\texttt{IRCoder 1.1B}} & \texttt{12.33} & \texttt{12.77} & \texttt{9.14} \vspace{-0.2em}\\
         & \diffup{2.10} & \diffdo{-0.12} & \diffup{2.11}\vspace{0.2em}\\
        \multirow{2}{*}{\texttt{IRCoder 1.3B}} & \texttt{11.87} & \texttt{16.62} & \texttt{3.88} \vspace{-0.2em}\\
         & \diffup{0.58} & \diffup{2.55} & \diffup{0.36} \vspace{0.2em}\\
        \multirow{2}{*}{\texttt{IRCoder 3.1B}} & \texttt{11.99} & \texttt{13.34} &  \texttt{9.88}\vspace{-0.2em}\\
         & \diffup{1.66} & \diffup{0.56} & \diffup{1.17} \vspace{0.2em}\\
        \multirow{2}{*}{\texttt{IRCoder 5.7B}} & \texttt{11.81} & \texttt{16.28} & \texttt{11.25} \vspace{-0.2em}\\
         & \diffup{1.72} & \diffup{2.13} & \diffup{0.39}\vspace{0.2em}\\
        \multirow{2}{*}{\texttt{IRCoder 6.7B}} & \texttt{9.90} & \texttt{15.61} & \texttt{7.95} \vspace{-0.2em}\\
         & \diffdo{0.06} & \diffup{1.28} & \diffup{3.01} \vspace{0.2em}\\
        \multirow{2}{*}{\texttt{IRCoder 7.3B}} & \texttt{10.71} & \texttt{13.49} & \texttt{9.28} \vspace{-0.2em}\\
         & \diffup{1.17} & \diffdo{0.03} & \diffup{0.11}\\
        \bottomrule
    \end{tabular}
    }
    \caption{CodeXGLUE code-to-text smoothed \texttt{BLEU-4} comparison between \texttt{IRCoder} and its corresponding base models.}
        \label{table:code2text}
\end{table*}


\begin{table*}[t]
    \centering
    \scalebox{0.66}{
    \begin{tabular}{rcccccccc}
        \toprule
        \scalerel*{\includegraphics{Graphics/robot.png}}{B} \texttt{Model} & \scalerel*{\includegraphics{Graphics/c.png}}{B} \texttt{C} & \scalerel*{\includegraphics{Graphics/cpp.png}}{B} \texttt{C++}  & \scalerel*{\includegraphics{Graphics/go.png}}{B} \texttt{Go} & \scalerel*{\includegraphics{Graphics/objc.png}}{B} \texttt{Obj-C} & \scalerel*{\includegraphics{Graphics/python.png}}{B} \texttt{Python} & \scalerel*{\includegraphics{Graphics/ruby.png}}{B} \texttt{Ruby} & \scalerel*{\includegraphics{Graphics/rust.png}}{B} \texttt{Rust} & \scalerel*{\includegraphics{Graphics/swift.png}}{B} \texttt{Swift} \\
        \midrule
         
        \texttt{StarCoderBase 1.1B} & \texttt{11.76} & \texttt{13.14} & \texttt{10.55} & \texttt{10.81} & \texttt{13.73} & \texttt{17.44} & \texttt{11.13} & \texttt{10.74}\\
        \texttt{DeepSeekCoder 1.3B} & \texttt{11.96} & \texttt{12.67} & \texttt{12.89} & \texttt{10.02} & \texttt{12.77} & \texttt{16.16} & \texttt{11.53} & \texttt{10.66}\\
        \texttt{StarCoderBase 3.1B} & \texttt{13.19} & \texttt{15.74} & \texttt{13.19} & \texttt{14.36} & \texttt{14.57} & \texttt{17.35} & \texttt{13.81} & \texttt{12.56}\\
        \texttt{DeepSeekCoder 5.7B} & \texttt{14.01} & \texttt{14.48} & \texttt{12.91} & \texttt{12.95} & \texttt{14.13} & \texttt{17.83} & \texttt{12.15} & \texttt{11.72}\\
        \texttt{CodeLlama 6.7B}& \texttt{14.26} & \texttt{14.89} & \texttt{13.96} & \texttt{14.24} & \texttt{14.44} & \texttt{17.73} & \texttt{13.37} & \texttt{12.76}\\
        \texttt{StarCoderBase 7.3B} & \texttt{15.06} & \texttt{17.01} & \texttt{14.56} & \texttt{14.68} & \texttt{15.01} & \texttt{18.38} & \texttt{14.06} & \texttt{12.97}\\
        \midrule
        \multirow{2}{*}{\texttt{IRCoder 1.1B}} & \texttt{11.91} & \texttt{12.99} & \texttt{11.99} & \texttt{12.25} & \texttt{14.09} & \texttt{18.91} & \texttt{11.98} & \texttt{11.04}\vspace{-0.2em}\\
         & \diffup{0.15} & \diffdo{0.15} & \diffup{1.44} & \diffup{1.44} & \diffup{0.36} & \diffup{1.47} & \diffup{0.85} & \diffup{0.30}\vspace{0.2em}\\
        \multirow{2}{*}{\texttt{IRCoder 1.3B}} & \texttt{12.20} & \texttt{14.68} & \texttt{12.85} & \texttt{11.93} & \texttt{13.29} & \texttt{
16.56} & \texttt{12.45} & \texttt{10.99}\vspace{-0.2em}\\
         & \diffup{0.24} & \diffup{2.01} & \diffup{0.04} & \diffup{1.91} & \diffup{0.52} & \diffup{0.40} & \diffup{0.92} & \diffup{0.33}\vspace{0.2em}\\
        \multirow{2}{*}{\texttt{IRCoder 3.1B}} & \texttt{13.39} & \texttt{14.69} & \texttt{13.88} & \texttt{13.95} & \texttt{14.51} & \texttt{17.41} & \texttt{13.64} & \texttt{12.81}\vspace{-0.2em}\\
         & \diffup{0.20} & \diffdo{1.05} & \diffup{0.69} & \diffdo{0.41} & \diffdo{0.06} & \diffup{0.06} & \diffdo{0.17} & \diffup{0.25}\vspace{0.2em}\\
        \multirow{2}{*}{\texttt{IRCoder 5.7B}} & \texttt{14.56} & \texttt{16.98} & \texttt{14.59} & \texttt{13.22} & \texttt{14.02} & \texttt{18.08} & \texttt{14.03} & \texttt{12.17}\vspace{-0.2em}\\
         & \diffup{0.55} & \diffup{2.50} & \diffup{1.68} & \diffup{0.27} & \diffdo{0.11} & \diffup{0.25} & \diffup{1.88} & \diffup{0.45}\vspace{0.2em}\\
        \multirow{2}{*}{\texttt{IRCoder 6.7B}} & \texttt{14.36} & \texttt{16.06} & \texttt{14.96} & \texttt{13.92} & \texttt{14.64} & \texttt{18.51} & \texttt{14.46} & \texttt{12.66}\vspace{-0.2em}\\
         & \diffup{0.10} & \diffup{1.17} & \diffup{1.00} & \diffdo{0.32} & \diffup{0.20} & \diffup{0.78} & \diffup{1.09} & \diffdo{0.10}\vspace{0.2em}\\
        \multirow{2}{*}{\texttt{IRCoder 7.3B}} & \texttt{15.32} & \texttt{17.75} & \texttt{15.76} & \texttt{14.92} & \texttt{15.98} & \texttt{19.26} & \texttt{14.98} & \texttt{13.09}\vspace{-0.2em}\\
         & \diffup{0.26} & \diffup{0.74} & \diffup{1.20} & \diffup{0.24} & \diffup{0.97} & \diffup{0.88} & \diffup{0.92} & \diffup{0.12}\\
        \bottomrule
    \end{tabular}
    }
    \caption{CommitChronicle \texttt{ROUGE-2} comparison between \texttt{IRCoder} and its corresponding base models.}
        \label{table:cc-2}
\end{table*}


\begin{table*}[t]
    \centering
    \scalebox{0.66}{
    \begin{tabular}{rcccccccc}
        \toprule
        \scalerel*{\includegraphics{Graphics/robot.png}}{B} \texttt{Model} & \scalerel*{\includegraphics{Graphics/c.png}}{B} \texttt{C} & \scalerel*{\includegraphics{Graphics/cpp.png}}{B} \texttt{C++}  & \scalerel*{\includegraphics{Graphics/go.png}}{B} \texttt{Go} & \scalerel*{\includegraphics{Graphics/objc.png}}{B} \texttt{Obj-C} & \scalerel*{\includegraphics{Graphics/python.png}}{B} \texttt{Python} & \scalerel*{\includegraphics{Graphics/ruby.png}}{B} \texttt{Ruby} & \scalerel*{\includegraphics{Graphics/rust.png}}{B} \texttt{Rust} & \scalerel*{\includegraphics{Graphics/swift.png}}{B} \texttt{Swift} \\
        \midrule
         
        \texttt{StarCoderBase 1.1B} & \texttt{29.78} & \texttt{35.63} & \texttt{32.56} & \texttt{30.03} & \texttt{35.47} & \texttt{39.56} & \texttt{33.97} & \texttt{32.32}\\
        \texttt{DeepSeekCoder 1.3B} & \texttt{31.87} & \texttt{35.91} & \texttt{33.98} & \texttt{30.59} & \texttt{34.71} & \texttt{34.62} & \texttt{32.86} & \texttt{30.77}\\
        \texttt{StarCoderBase 3.1B} & \texttt{32.76} & \texttt{37.84} & \texttt{36.73} & \texttt{37.72} & \texttt{35.76} & \texttt{37.48} & \texttt{34.23} & \texttt{33.06}\\
        \texttt{DeepSeekCoder 5.7B} & \texttt{32.61} & \texttt{36.25} & \texttt{35.48} & \texttt{32.63} & \texttt{36.56} & \texttt{40.07} & \texttt{34.69} & \texttt{33.16}\\
        \texttt{CodeLlama 6.7B} & \texttt{35.02} & \texttt{37.24} & \texttt{36.82} & \texttt{33.43} & \texttt{33.73} & \texttt{39.75} & \texttt{34.46} & \texttt{33.83}\\
        \texttt{StarCoderBase 7.3B} & \texttt{35.93} & \texttt{38.67} & \texttt{37.24} & \texttt{38.53} & \texttt{38.17} & \texttt{40.97} & \texttt{37.28} & \texttt{35.09}\\
        \midrule
        \multirow{2}{*}{\texttt{IRCoder 1.1B}} & \texttt{32.67} & \texttt{34.84} & \texttt{34.79} & \texttt{33.43} & \texttt{35.56} & \texttt{40.19} & \texttt{35.21} & \texttt{33.63}\vspace{-0.2em}\\
         & \diffup{2.89} & \diffdo{0.79} & \diffup{2.23} & \diffup{3.40} & \diffup{0.09} & \diffup{0.63} & \diffup{1.24} & \diffup{1.31}\vspace{0.2em}\\
        \multirow{2}{*}{\texttt{IRCoder 1.3B}} & \texttt{33.30} & \texttt{36.14} & \texttt{35.11} & \texttt{31.58} & \texttt{34.25} & \texttt{39.34} & \texttt{34.69} & \texttt{32.14}\vspace{-0.2em}\\
         & \diffup{1.43} & \diffup{0.23} & \diffup{1.13} & \diffup{0.99} & \diffdo{0.46} & \diffup{4.72} & \diffup{1.83} & \diffup{1.37}\vspace{0.2em}\\
        \multirow{2}{*}{\texttt{IRCoder 3.1B}} & \texttt{35.29} & \texttt{38.21} & \texttt{36.90} & \texttt{36.83} & \texttt{37.01} & \texttt{40.40} & \texttt{36.07} & \texttt{33.76}\vspace{-0.2em}\\
         & \diffup{2.53} & \diffup{0.37} & \diffup{0.17} & \diffdo{0.89} & \diffup{1.25} & \diffup{2.92} & \diffup{1.84} & \diffup{0.70}\vspace{0.2em}\\
        \multirow{2}{*}{\texttt{IRCoder 5.7B}} & \texttt{36.15} & \texttt{39.39} & \texttt{37.44} & \texttt{34.73} & \texttt{36.78} & \texttt{41.38} & \texttt{36.97} & \texttt{34.33}\vspace{-0.2em}\\
         & \diffup{3.54} & \diffup{3.14} & \diffup{1.96} & \diffup{2.10} & \diffup{0.22} & \diffup{1.31} & \diffup{2.28} & \diffup{1.17}\vspace{0.2em}\\
        \multirow{2}{*}{\texttt{IRCoder 6.7B}} & \texttt{35.94} & \texttt{37.81} & \texttt{38.11} & \texttt{34.08} & \texttt{37.01} & \texttt{41.02} & \texttt{36.41} & \texttt{41.17}\vspace{-0.2em}\\
         & \diffup{0.92} & \diffup{0.57} & \diffup{1.29} & \diffup{0.65} & \diffup{0.28} & \diffup{1.27} & \diffup{1.95} & \diffup{0.34}\vspace{0.2em}\\
        \multirow{2}{*}{\texttt{IRCoder 7.3B}} & \texttt{37.52} & \texttt{40.48} & \texttt{38.96} & \texttt{38.59} & \texttt{38.69} & \texttt{43.17} & \texttt{38.41} & \texttt{35.82}\vspace{-0.2em}\\
         & \diffup{1.59} & \diffup{1.81} & \diffup{1.72} & \diffup{0.06} & \diffup{0.52} & \diffup{2.20} & \diffup{1.13} & \diffup{0.73}\\
        \bottomrule
    \end{tabular}
    }
    \caption{CommitChronicle \texttt{ROUGE-L} comparison between \texttt{IRCoder} and its corresponding base models.}
        \label{table:cc-L}
\end{table*}

\makeatletter
\setlength{\@dblfptop}{0pt}
\setlength{\@dblfpbot}{290pt}
\makeatother


\begin{table*}[t]
    \centering
    \scalebox{0.66}{
    \begin{tabular}{rcccc}
        \toprule
        \scalerel*{\includegraphics{Graphics/robot.png}}{B} \texttt{Model} & \scalerel*{\includegraphics{Graphics/cpp.png}}{B} \texttt{C++} & \scalerel*{\includegraphics{Graphics/go.png}}{B} \texttt{Go} & \scalerel*{\includegraphics{Graphics/python.png}}{B} \texttt{Python} & \scalerel*{\includegraphics{Graphics/rust.png}}{B} \texttt{Rust} \\
        \midrule
         
        \texttt{StarCoderBase 1.1B} & \texttt{11.26} & \texttt{10.07} & \texttt{18.87} & \texttt{8.70} \\
        \texttt{DeepSeekCoder 1.3B} & \texttt{24.01} & \texttt{25.89} & \texttt{35.36} & \texttt{16.67} \\
        \texttt{StarCoderBase 3.1B} & \texttt{33.87} & \texttt{27.18} & \texttt{34.92} & \texttt{18.89} \\
        \texttt{DeepSeekCoder 5.7B} & \texttt{52.32} & \texttt{52.89} & \texttt{57.88} & \texttt{29.76}\\
        \texttt{CodeLlama 6.7B} & \texttt{49.39} & \texttt{49.76} & \texttt{51.68} & \texttt{27.17}\\
        \texttt{StarCoderBase 7.3B} & \texttt{44.37} & \texttt{43.77} & \texttt{48.33} & \texttt{26.48}\\
        \midrule
        \multirow{2}{*}{\texttt{IRCoder 1.1B}} & \texttt{11.71} & \texttt{10.76} & \texttt{19.93} & \texttt{9.63}\vspace{-0.2em}\\
         & \diffup{0.45} & \diffup{0.69} & \diffup{1.06} & \diffup{0.93}\vspace{0.2em}\\
        \multirow{2}{*}{\texttt{IRCoder 1.3B}} & \texttt{26.97} & \texttt{27.12} & \texttt{35.21} & \texttt{18.98}\vspace{-0.2em}\\
         & \diffup{2.96} & \diffup{1.23} & \diffdo{0.15} & \diffup{2.31}\vspace{0.2em}\\
        \multirow{2}{*}{\texttt{IRCoder 3.1B}} & \texttt{32.99} & \texttt{25.48} & \texttt{35.67} & \texttt{21.82}\vspace{-0.2em}\\
         & \diffdo{0.88} & \diffdo{1.70} & \diffup{0.75} & \diffup{4.03}\vspace{0.2em}\\
        \multirow{2}{*}{\texttt{IRCoder 5.7B}} & \texttt{53.94} & \texttt{52.64} & \texttt{58.77} & \texttt{33.79}\vspace{-0.2em}\\
         & \diffup{1.62} & \diffdo{0.25} & \diffup{0.89} & \diffup{4.03}\vspace{0.2em}\\
        \multirow{2}{*}{\texttt{IRCoder 6.7B}} & \texttt{50.96} & \texttt{51.33} & \texttt{55.69} & \texttt{28.38}\vspace{-0.2em}\\
         & \diffup{1.57} & \diffup{1.57} & \diffup{4.01} & \diffup{1.21}\vspace{0.2em}\\
        \multirow{2}{*}{\texttt{IRCoder 7.3B}} & \texttt{48.62} & \texttt{46.04} & \texttt{49.52} & \texttt{32.11}\vspace{-0.2em}\\
         & \diffup{4.25} & \diffup{2.27} & \diffup{1.19} & \diffup{5.63}\\
        \bottomrule
    \end{tabular}
    }
    \caption{HumanEvalFixDocs \texttt{pass@1} comparison between \texttt{IRCoder} and its corresponding base models.}
        \label{table:hefixdocs-1}
\end{table*}

\begin{table*}[ht!]
    \centering
    \scalebox{0.66}{
    \begin{tabular}{rcccc}
        \toprule
        \scalerel*{\includegraphics{Graphics/robot.png}}{B} \texttt{Model} & \scalerel*{\includegraphics{Graphics/cpp.png}}{B} \texttt{C++} & \scalerel*{\includegraphics{Graphics/go.png}}{B} \texttt{Go} & \scalerel*{\includegraphics{Graphics/python.png}}{B} \texttt{Python} & \scalerel*{\includegraphics{Graphics/rust.png}}{B} \texttt{Rust} \\
        \midrule
         
        \texttt{StarCoderBase 1.1B} & \texttt{16.91} & \texttt{14.48} & \texttt{26.67} & \texttt{12.74} \\
        \texttt{DeepSeekCoder 1.3B} & \texttt{35.49} & \texttt{40.78} & \texttt{42.94} & \texttt{25.68} \\
        \texttt{StarCoderBase 3.1B} & \texttt{45.39} & \texttt{44.11} & \texttt{54.67} & \texttt{27.16} \\
        \texttt{DeepSeekCoder 5.7B} & \texttt{64.10} & \texttt{60.44} & \texttt{70.59} & \texttt{49.07}\\
        \texttt{CodeLlama 6.7B} & \texttt{61.32} & \texttt{58.73} & \texttt{61.67} & \texttt{45.44}\\
        \texttt{StarCoderBase 7.3B} & \texttt{55.94} & \texttt{58.36} & \texttt{59.56} & \texttt{47.22}\\
        \midrule
        \multirow{2}{*}{\texttt{IRCoder 1.1B}} & \texttt{18.16} & \texttt{13.99} & \texttt{26.67} & \texttt{13,18}\vspace{-0.2em}\\
         & \diffup{1.25} & \diffdo{0.49} & \textbf{\texttt{-}} & \diffup{0.44}\vspace{0.2em}\\
        \multirow{2}{*}{\texttt{IRCoder 1.3B}} & \texttt{38.41} & \texttt{43.11} & \texttt{43.15} & \texttt{27.11}\vspace{-0.2em}\\
         & \diffup{2.92} & \diffup{2.31} & \diffup{0.23} & \diffup{1.43}\vspace{0.2em}\\
        \multirow{2}{*}{\texttt{IRCoder 3.1B}} & \texttt{42.06} & \texttt{45.17} & \texttt{55.06} & \texttt{25.74}\vspace{-0.2em}\\
         & \diffdo{3.33} & \diffup{0.76} & \diffup{0.39} & \diffup{1.58}\vspace{0.2em}\\
        \multirow{2}{*}{\texttt{IRCoder 5.7B}} & \texttt{68.22} & \texttt{63.47} & \texttt{73.42} & \texttt{56.27}\vspace{-0.2em}\\
         & \diffup{4.12} & \diffup{3.03} & \diffup{2.83} & \diffup{7.20}\vspace{0.2em}\\
        \multirow{2}{*}{\texttt{IRCoder 6.7B}} & \texttt{61.87} & \texttt{60.59} & \texttt{62.04} & \texttt{50.44}\vspace{-0.2em}\\
         & \diffup{0.55} & \diffup{1.86} & \diffup{0.37} & \diffup{5.00}\vspace{0.2em}\\
        \multirow{2}{*}{\texttt{IRCoder 7.3B}} & \texttt{57.35} & \texttt{59.33} & \texttt{60.11} & \texttt{52.71}\vspace{-0.2em}\\
         & \diffup{1.41} & \diffup{0.97} & \diffup{0.55} & \diffup{5.49}\\
        \bottomrule
    \end{tabular}
    }
    \caption{HumanEvalFixDocs \texttt{pass@10} comparison between \texttt{IRCoder} and its corresponding base models.}
        \label{table:hefixdocs-10}
\end{table*}

\end{document}